\documentclass{article}

\usepackage{microtype}
\usepackage{graphicx}
\usepackage{booktabs} 

\usepackage{hyperref}
\usepackage{subcaption}
\usepackage{balance}
\usepackage{tocloft}

\usepackage[accepted]{icml2025}

\usepackage{amsmath}
\usepackage{amssymb}
\usepackage{mathtools}
\usepackage{amsthm}

\usepackage[capitalize,noabbrev]{cleveref}

\usepackage{multirow}
\usepackage{footmisc}
\usepackage{enumitem}

\usepackage[normalem]{ulem}
\useunder{\uline}{\ul}{}

\newcommand{\ours}{LeTE}
\newcommand{\oursfull}{\uline{Le}arnable Transformation-based Generalized \uline{T}ime \uline{E}ncoding}

\theoremstyle{plain}
\newtheorem{theorem}{Theorem}[section]
\newtheorem{proposition}[theorem]{Proposition}

\theoremstyle{definition}

\theoremstyle{remark}

\usepackage[textsize=tiny]{todonotes}

\icmltitlerunning{Rethinking Time Encoding via Learnable Transformation Functions}

\begin{document}

\twocolumn[
\icmltitle{Rethinking Time Encoding via Learnable Transformation Functions}



\icmlsetsymbol{equal}{*}

\begin{icmlauthorlist}
\icmlauthor{Xi Chen}{fudan}
\icmlauthor{Yateng Tang}{tencent}
\icmlauthor{Jiarong Xu}{fudan2}
\icmlauthor{Jiawei Zhang}{davis}
\icmlauthor{Siwei Zhang}{fudan}
\icmlauthor{Sijia Peng}{fudan}
\icmlauthor{Xuehao Zheng}{tencent}
\icmlauthor{Yun Xiong}{fudan}
\end{icmlauthorlist}

\icmlaffiliation{fudan}{Shanghai Key Laboratory of Data Science, School of Computer Science, Fudan University, Shanghai, China}
\icmlaffiliation{fudan2}{School of Management, Fudan University, Shanghai, China}
\icmlaffiliation{davis}{IFM Lab, University of California, Davis, CA, USA}
\icmlaffiliation{tencent}{Tencent Weixin Group, Shenzhen, China}

\icmlcorrespondingauthor{Yun Xiong}{yunx@fudan.edu.cn}

\icmlkeywords{Time Encoding, Deep Function Learning}

\vskip 0.3in
]



\printAffiliationsAndNotice{} 

\begin{abstract}
Effectively modeling time information and incorporating it into applications or models involving chronologically occurring events is crucial. 
Real-world scenarios often involve diverse and complex time patterns, which pose significant challenges for time encoding methods. While previous methods focus on capturing time patterns, many rely on specific inductive biases, such as using trigonometric functions to model periodicity. This narrow focus on single-pattern modeling makes them less effective in handling the diversity and complexities of real-world time patterns.
In this paper, we investigate to improve the existing commonly used time encoding methods and introduce \textbf{\oursfull~(\ours)}. We propose using deep function learning techniques to parameterize non-linear transformations in time encoding, making them learnable and capable of modeling generalized time patterns, including diverse and complex temporal dynamics.
By enabling learnable transformations, \ours~encompasses previous methods as specific cases and allows seamless integration into a wide range of tasks. Through extensive experiments across diverse domains, we demonstrate the versatility and effectiveness of \ours.
\end{abstract}

\section{Introduction}
\label{sec-intro}

Time-related data are commonly observed in real-world applications, such as user transaction data in financial institutions \cite{kazemi2020representation, lezmi2023time}, purchase behavior sequences in e-commerce \cite{kang2018self, tgn, skarding2021foundations}, and climate observations in weather forecasting \cite{murat2018forecasting, neumann2024intrinsic}.
Adopting time series models and dynamic graph models to handle time-related data are two common approaches \cite{timesnet, dygformer}. In both cases, effectively incorporating time information is crucial for making accurate predictions. 
To achieve this, existing research works typically employ time encoding methods to capture and represent time information, with the resulting time embedding being treated as an independent feature in time series forecasting and dynamic graph representation learning models.

\begin{figure*}[t]
    \centering
    \includegraphics[width=1\linewidth]{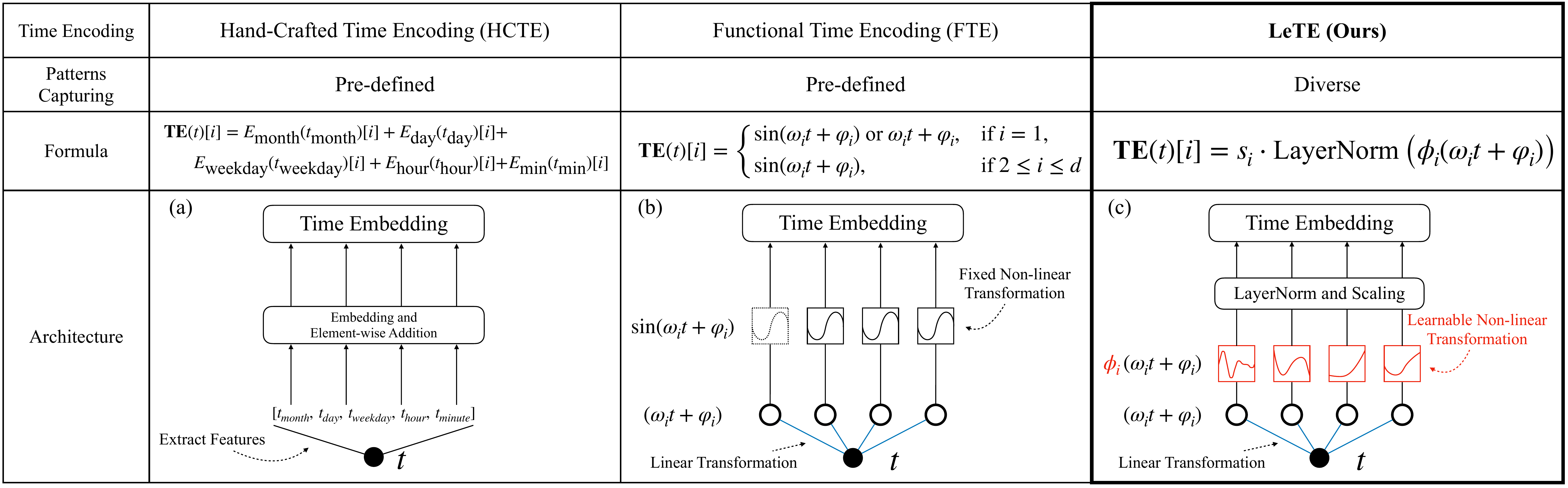}
    \caption{A comparison of previous time encoding methods and proposed \ours.}
    \label{fig_t2v}
\vspace{-12pt}
\end{figure*}

Early studies represent time by using hand-crafted temporal features designed specifically for downstream tasks \cite{choi2016doctor, baytas2017patient, kwon2018retainvis}. A prominent example is the time encoding method illustrated in Figure \ref{fig_t2v} (a), which is widely used in existing time series processing work \cite{MICN, timesnet}. Such methods typically involve manually splitting timestamps into components (e.g., month, day, etc.), assigning a embedding to each component, and adding these embeddings to form the final time embedding. 
However, these methods are resource-intensive and often rely on domain expertise, which may limit their abilities to capture only specific pre-defined time patterns \cite{time2vec}.

With the rapid development of attention mechanisms, which offer advantages such as better handling of long-range dependencies and adaptive weighting of time-related information, subsequent research on time series and dynamic graphs has increasingly leveraged these mechanisms \cite{tgat, dygformer, liu2023itransformer}. To better model time and ensure compatibility between time encoding methods and self-attention, Functional Time Encoding (FTE) methods were proposed, with two representative works: Functional Time Representation \cite{xu2019time} and Time2Vec \cite{time2vec}, as illustrated in Figure \ref{fig_t2v} (b). Nearly all subsequent dynamic graph representation learning research employs these methods to encode time \cite{dygformer}. 
These techniques transform time input into multi-dimensional time embeddings by applying multiple linear transformations followed by pre-defined non-linear transformation functions. 
Due to their reliance on pre-defined non-linear transformations—such as trigonometric functions to capture periodic patterns—these methods are inherently limited to capturing fixed, specific time patterns.
As a result, they often struggle to represent more complex, non-linear temporal dynamics \cite{time2vec, timesnet} and require additional dimensions to account for diverse periodic components \cite{tgat, tgn, zeng2024much}.
Furthermore, these encoding methods frequently lack the capacity to effectively model non-periodic patterns, such as trends, irregularities.

We observe that previous time encoding methods—whether hand-crafted or functional—primarily introduce a strong inductive bias rooted in the periodic nature of human behavior and natural phenomena \cite{li2017time, xu2019time, time2vec}. They mainly focus on capturing pre-defined periodic patterns, often struggling to capture more complex ones, such as non-periodic and mixed patterns. 

However, in real-world scenarios, data often exhibits a complex interplay of mixed patterns, making accurate modeling more challenging.
For instance, in financial risk control, periodic patterns—such as daily transaction peaks, weekly spending habits, and seasonal trends around holidays or salary payments—offer valuable insights into predictable behaviors. On the other hand, non-periodic events, such as sudden spikes in transactions caused by market fluctuations, regulatory changes, or potential fraudulent activities, necessitate flexible and adaptive modeling techniques. 
Moreover, different patterns often coexist. For example, fraudsters may blend regular periodic transactions with abnormal non-periodic activities to evade the detection by regulators.
To illustrate the presence of complex mixed time patterns in real-world data, we conduct extensive investigations in this paper, with partial results in Appendix \ref{app_periodic_and_non}.

This motivates us to rethink the design of time encodings. We argue that an effective time encoding method should adhere to a key principle to enable comprehensive and accurate analysis: Capacity for modeling diverse and complex time patterns, i.e., the method should be capable of capturing a wide range of time patterns, including periodic, non-periodic, and mixed patterns. 


To better encode time information and simultaneously capture diverse time patterns, we propose \textbf{\oursfull}, abbreviated as \textbf{\ours}—a simple yet effective time encoding method. 
Instead of hand-crafting time encoding or relying on pre-defined non-linear transformations, 
we draw inspiration from deep function learning, which is known for its generalizability, interpretability, and reusability \cite{tinybig, kan}, and propose to use learnable non-linear transformations for time encoding. 
Specifically, we parameterize non-linear transformation functions using techniques derived from deep function learning. This parameterization makes the transformations learnable and jointly optimizable with the model’s parameters under supervision from downstream tasks, allowing them to flexibly adapt to both linear and arbitrary non-linear forms.
With learnable transformations, \ours~adaptively models time information, enabling different dimensions of time encoding to capture complex time patterns—such as irregular trends, abrupt changes, and overlapping periodicities—that are commonly encountered in real-world scenarios and beyond the capabilities of previous methods. This generalization also allows our method to encompass previous approaches as specific cases.
An illustration of \ours~is in Figure \ref{fig_t2v} (c).


\ours~also offers following advantages (cf. Section \ref{sec:properties}). Since time can be measured on different scales, its representation should be invariant to time rescaling \cite{time2vec}. We prove that \ours~satisfies this property (cf. Appendix \ref{proof_pro_invariant}). Furthermore, we prove that \ours~is a generalized version of previous methods and can integrate seamlessly with various models
(cf. Section \ref{sec:properties}). 
By employing an interpretable deep function learning approach, \ours~achieves a high degree of interpretability (cf. Appendix \ref{app_visual}). Additionally, experimental results demonstrate that \ours~achieves superior results with fewer dimensions than previous time encodings, as the learnable transformations capture part of the complexity that would otherwise require higher-dimensional embeddings (cf. Section \ref{sec-exp_dim}).

We highlight our contributions as follows:
\begin{itemize}
    \item We reinvestigate the design of the existing time encoding methods, highlighting their limitations in handling real-world data and propose \ours, a generalized time encoding method that allows the entire encoding process, including both linear and non-linear transformations, fully parameterized and learnable.
    \item \ours~has the capacity to model diverse and complex time patterns, and it offers additional benefits, including invariance to time rescaling, plug-and-play functionality, enhanced interpretability and improved dimensional efficiency.
    \item Through extensive experiments across diverse domains—including event-based image classification, time series forecasting, dynamic graph representation learning and real-world applications-we demonstrate the effectiveness and versatility of \ours.
\end{itemize}

\section{Preliminaries}
\label{sec-preliminaries}

\subsection{Functional Time Encodings}
\label{sec-pre-time_encoding}
Functional Time Encoding (FTE) methods can be viewed as feature mappings from 1-dimensional input time to a high-dimensional time embedding: $ \Phi: t \in \mathbb{R}^1 \rightarrow \textbf{TE} \in \mathbb{R}^d$, where $t \in [0, t_\text{max}]$ is from the value range bounded by $t_\text{max}$.
Two representative works of FTE are Functional Time Representation (FTR) \cite{xu2019time} and Time2Vec (T2V) \cite{time2vec}. Although these methods construct time encodings from different perspectives, they are mathematically nearly identical (the only difference is that a separate dimension that undergoes only a linear transformation is used by T2V to capture non-periodic patterns). We state the following proposition, with details of the two methods and the proof provided in Appendix \ref{proof_pro_unify_te}.

\begin{proposition}
\label{pro_unify_te} 
Mathematically, with selected values for $\omega_i$ and $\varphi_i$, the aforementioned FTR and T2V can be unified into the following forms:

Including the first dimension:
\begin{equation}
    \textbf{TE}(t)[i]=
\begin{cases}
\sin{(\omega_i t + \varphi_i)} \text{ or } \omega_i t + \varphi_i, & \text{if } i = 1, \\
\sin{(\omega_i t + \varphi_i)}, & \text{if } 2 \leq i \leq d.
\end{cases}
\end{equation}
Or excluding the first dimension:
\begin{equation}
\label{equ_pro_unify_te_1}
      \textbf{TE}(t)[i]=\sin{(\omega_i t + \varphi_i)}, 
\end{equation}
or 
\begin{equation}
\label{equ_pro_unify_te_2}
   \textbf{TE}(t) = \left[\sin(\omega_1 t + \varphi_1), \cdots, \sin(\omega_d t + \varphi_d)\right]
\end{equation}
\end{proposition}

For simplicity, we use Functional Time Encoding (FTE) to refer to both the FTR and T2V throughout the paper.

\subsection{Deep Function Learning}
Deep Function Learning refers to the approach of learning target functions by optimizing parameterized functions, such as polynomials, sinusoidal functions, or splines. This method leverages the flexibility of parameterized functions and optimizes their parameters using deep learning frameworks to approximate complex functions \cite{tinybig}.

\textbf{Fourier Series Expansion:}
The Fourier series expresses a target function $f(x)$ as a combination of sine and cosine functions:
\begin{equation}
    f(x) = a_0 + \sum_{n=1}^{N} \left( a_n \cos(n \omega x) + b_n \sin(n \omega x) \right)
\end{equation}
Here, $a_n$ and $b_n$ are learnable coefficients, and $\omega$ is the fundamental frequency. By optimizing $a_0$, $a_n$ and $b_n$ through a learning process, the function $f(x)$ can approximate complex patterns (cf. Appendix \ref{app_fourier} for details). Unlike fixed sine functions, parameterized functions adjust their amplitude, frequency, and phase through downstream supervisory signals, enabling them to effectively model a wider range of patterns.

\textbf{Spline Functions:}
Spline functions approximate a target function $f(x)$ using a sum of piecewise polynomial basis functions:
\begin{equation}
    f(x) = \sum_{i=1}^{n} c_i B_i(x)
\end{equation}
Here, $c_i$ are control points, and $B_i(t)$ are the basis functions (e.g., B-splines). By learning and optimizing the control points, knot positions, and weights (cf. Appendix \ref{app_spline} for details), splines provide a smooth and accurate representation of diverse functions. Their piecewise and localized structure makes them highly adaptable for complex function modeling.



\section{Methods}
\label{sec-methods}

\begin{figure*}[t]
    \centering
    \includegraphics[width=0.8\linewidth]{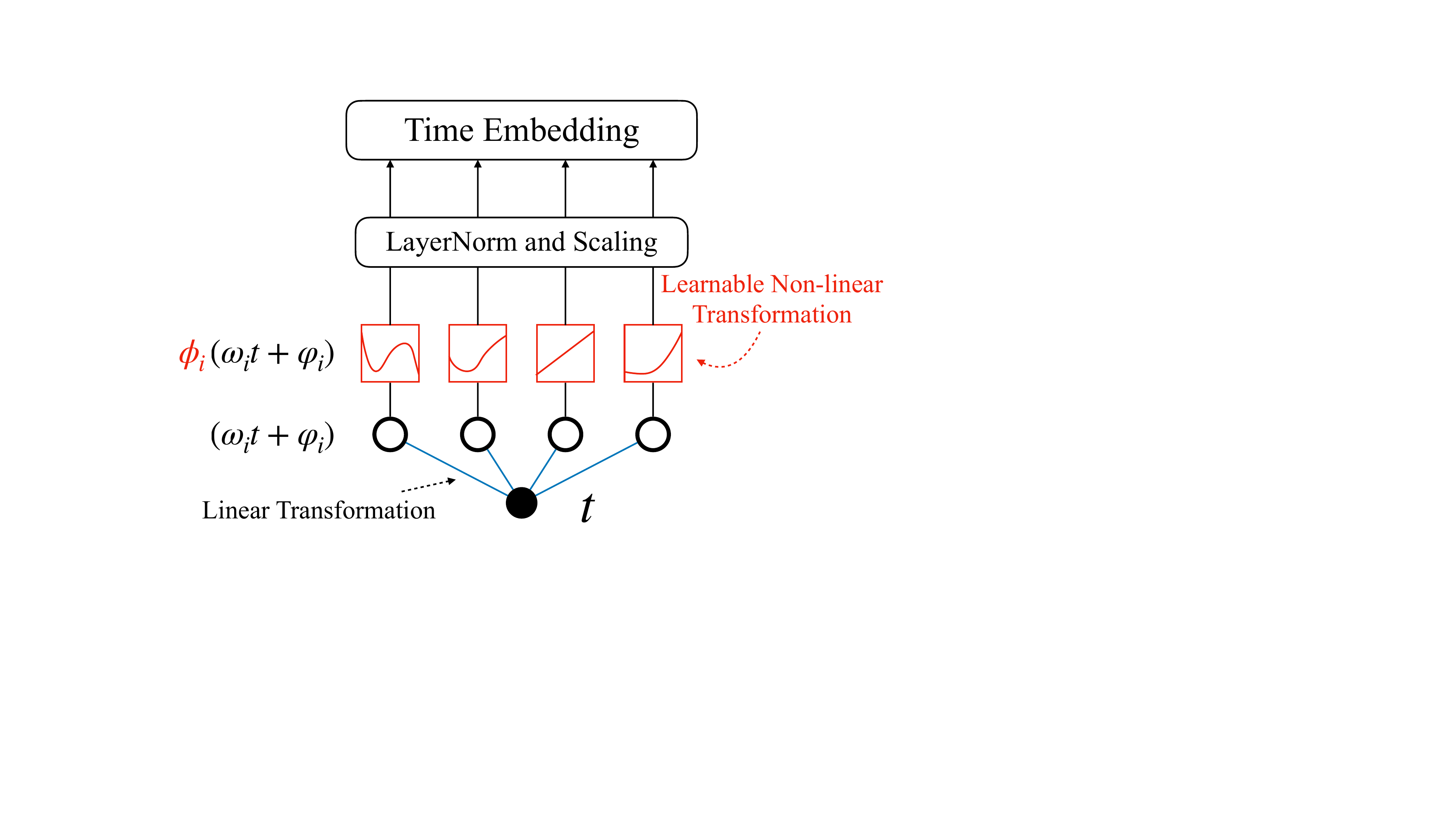}
    \caption{An illustration of Combined \ours: the first dimension is parameterized by Fourier series expansion and the last dimension is parameterized by B-Splines.}
    \label{fig:parameterlize}
\vspace{-10pt}
\end{figure*}

\subsection{\ours}
To address the limitations of previous time encodings—specifically, their restricted capacity to model fixed or pre-defined time patterns—we propose \textbf{\oursfull} (referred to as \textbf{\ours}). 
To capture diverse and complex patterns in time-related data, we propose techniques that make non-linear transformations learnable. This approach allows the model to dynamically adapt its transformations, enabling more precise representations of time patterns.
To achieve this, we employ two distinct approaches for constructing learnable transformation functions: Fourier series expansion and Spline functions. Both methods share the ability to effectively capture and model complex, non-linear temporal patterns while maintaining flexibility in handling various time dynamics.
Based on these methods, we propose three variations of \ours. First, we construct the learnable transformation functions using these two approaches, categorizing them as Fourier-based \ours~and Spline-based \ours~according to their respective construction methods. We then integrate these two variations to develop a more generalized version, referred to as Combined \ours, which leverages the strengths of both approaches.


For a scalar timestamp input $t$, \ours~for $t$, denoted as $\textbf{\ours}(t)$, is a $d$-dimensional time embedding vector:
\begin{equation}
\label{eq_t2vp}
    \textbf{\ours}(t)[i] = \phi_i(\omega_i t + \varphi_i),
\end{equation}
where $\omega_i$ and $\varphi_i$ are learnable parameters, and $\phi_i$ are learnable functions that can be parameterized by either Fourier series expansion or B-splines.

\textbf{Fourier-based \ours}: This method assumes that $\phi_i$ are parameterized by Fourier series expansion:
\begin{equation}
\label{eq_fourier}
    \phi_i(x) = a_0 + \sum_{k=1}^{K} \left( a_k \cos(k x) + b_k \sin(k x) \right),
\end{equation}
where $a_0$, $a_k$, and $b_k$ are the parameters to be learned, and $K$ represents the number of terms in the expansion. \ours~can then be expressed as:
\begin{equation}
\label{eq_fourier_t2vp}
    \begin{aligned}
        \textbf{\ours}(t)[i] = & \;  a_{i,0} + \sum_{k=1}^{K} \Big( a_{i,k} \cos(k (\omega_i t + \varphi_i)) \\
                              & \; + b_{i,k} \sin(k (\omega_i t + \varphi_i)) \Big) \Bigg.
    \end{aligned}
\end{equation}

\textbf{Spline-based \ours}: This method assumes that $\phi_i$ are parameterized by B-splines:
\begin{equation} 
\label{eq_bspline}
    \phi_i(x) = \sum_{j=1}^{M} c_j B_j(x),
\end{equation}
where $M$ is the number of B-spline basis functions, and $B_j(x)$ is the $j$-th B-spline basis function. \ours~can then be expressed as:
\begin{equation}
\label{eq_bspline_t2vp}
    \textbf{\ours}(t)[i] =  \sum_{j=1}^{M} c_{i,j} B_j(\omega_i t + \varphi_i) .
\end{equation}
Comparing these two approaches for constructing $\phi_i$, the Fourier series expansion in \ours~enforces periodicity in $\phi_i$, while the B-spline approach provides flexibility to model more complex $\phi_i$ functions, including both periodic and non-periodic patterns.

Recall that we neglect the first dimension of the linear transformation of time in Equations \eqref{equ_pro_unify_te_1} and \eqref{equ_pro_unify_te_2}, which models the non-periodic patterns for Time2Vec. However, in \ours~, by allowing $\phi_i$ to be learnable, different $\phi_i$ at different dimensions can capture more complex non-periodic patterns based on the supervision signals from downstream tasks.

\textbf{Combined \ours}: To enhance the capability of the time encoding to capture diverse and complex time patterns and to build a more generalized version of \ours, we further propose a straightforward extension: applying Fourier-based \ours~to a portion of the time encoding dimensions and Spline-based \ours~to the remaining dimensions. The proportion of Fourier-based \ours~and Spline-based \ours~used can be controlled by a hyperparameter $p$. To address potential differences in the output scales of Fourier-based and Spline-based \ours, we introduce a Layer Normalization layer followed by a learnable scaling weight for the time encoding. For a $d$-dimensional \ours, the time embedding is formulated as:
\begin{equation}
        \textbf{\ours}(t)[i] = s_i \cdot \text{LayerNorm} \left( \phi_i(\omega_i t + \varphi_i) \right),
\end{equation}
where $[s_i]$ is a $d$-length learnable scaling weight vector, and 
\begin{equation}
    \phi_i(x) =
    \begin{cases}
        \text{Equation} \eqref{eq_fourier}, & \text{if~~$i \le \lfloor p \cdot d \rfloor $}, \\
        \text{Equation} \eqref{eq_bspline}, & \text{if~~$i > \lfloor p\cdot d \rfloor$}.
    \end{cases}
\end{equation}
When $p = 1$, the method corresponds to Fourier-based \ours, and when $p = 0$, it corresponds to Spline-based \ours.
For the remainder of this paper, unless otherwise specified, \textbf{\ours}~will refer to the Combined \ours, where $p$ is set to 0.5. An illustration of \ours{} is shown in Figure \ref{fig:parameterlize}, and the implementation details are provided in Appendix \ref{app_imp}.

\subsection{Properties of \ours}
\label{sec:properties}
In this subsection, we present the properties of our method from the perspective of theoretical analysis. Specifically, we discuss the strengths of Fourier-based \ours~and Spline-based \ours~individually. Since Combined \ours~integrates these two variations, it naturally inherits their respective properties.

\textbf{Generalizability}: Compared with previous methods, which can only capture pre-defined time patterns—usually periodic ones—our method offers greater generalizability, enabling it to capture a wider range of diverse and complex patterns, including periodic, non-periodic and mixed ones.
Naturally, Fourier-based \ours~(as formulated in Equation \eqref{eq_fourier_t2vp}, the $\phi_i$ functions are parameterized by Fourier series expansion) can model periodicity, as it resembles a Fourier series expansion with weighted sums of sine and cosine terms at different frequencies and phases. By learning appropriate values for $\omega_i$ and incorporating different harmonics $k$, the function can approximate complex periodic patterns and capture repeating structures in time. The learnable parameters $\omega_i$ and $\varphi_i$ enable the model to adapt to various periodic characteristics in the data. Although the Fourier series is inherently periodic, non-periodic patterns can also be modeled: the learnable parameters $a_{i,0}$, $a_{i,k}$, $b_{i,k}$, $\omega_i$, and $\varphi_i$ provide flexibility to approximate non-periodic behaviors. By using very small or large values for $\omega_i$, the model can fit signals with long or slow-varying cycles, effectively creating non-repeating patterns over a finite interval. 
Additionally, the combination of learned frequencies, phases, and amplitudes can produce complex patterns that do not repeat over the observed range, thereby approximating non-periodic signals. 
For similar reasons, and given the generality of functions formed by splines, Spline-based \ours~(Equation \eqref{eq_bspline_t2vp}, the $\phi_i$ are parameterized by B-splines) can also model both periodic and non-periodic patterns. 
Naturally, through multi-dimensional encoding, both Fourier-based and Spline-based \ours~are capable of capturing mixed time patterns.
Although Fourier-based \ours~can capture non-periodic patterns, its inherent periodicity makes it particularly effective at modeling the periodicity of time. Conversely, while Spline-based \ours~is also capable of capturing periodic patterns, it exhibits a stronger ability to model non-periodic patterns. 
Therefore, by combining Fourier-based \ours~and Spline-based \ours, the resulting Combined \ours~achieves enhanced capability to capture diverse patterns.
Intuitively, the previous FTEs are special cases of \ours; we present the following proposition with its corresponding proof to demonstrate this.
\begin{proposition}
\label{pro_special_case}
For an arbitrary input $t$, the network can learn a set of parameters such that \ours~can replicate the effects of previous time encodings, making previous time encodings specific cases of \ours.
\end{proposition}
\proof
For Equations \eqref{eq_fourier_t2vp} and \eqref{eq_bspline_t2vp}, we only need to find a set of coefficients for Equations \eqref{eq_fourier} and \eqref{eq_bspline} to approximate the sine function, respectively.
By selecting $K = $1, $a_{i,0} = 0$, $a_{i,1} = 0$, and $b_{i,1} = 1$, Equation \eqref{eq_fourier_t2vp} becomes:
\begin{equation}
\begin{aligned}
    \textbf{\ours}(t)[i] &= 0 + \Big( 0 \cdot \cos\big(1 \cdot (\omega_i t + \varphi_i)\big) \\
    &\quad + 1 \cdot \sin\big(1 \cdot (\omega_i t + \varphi_i)\big) \Big) \\
    &= \sin(\omega_i t + \varphi_i).
\end{aligned}
\end{equation}
Thus, $\sin(\omega_i t + \varphi_i)$ is indeed a special case of the more general formula in Equation \eqref{eq_fourier_t2vp}.
The proof continues in Appendix \ref{proof_pro_special_case}, which demonstrates that $\sin(\omega_i t + \varphi_i)$ is also a special case of Equation \eqref{eq_bspline_t2vp}.
\endproof

Moreover, since Xu et al. claim that absolute position encoding is a special case of functional time representation \cite{xu2019time, tgat}, it is straightforward to see that absolute position encoding is also a special case of our \ours.

\textbf{Invariance to Time Rescaling}: Since time can be represented on various scales (such as days, hours, or seconds), a key characteristic of a time representation is its invariance to rescaling \cite{time2vec, tallec2018can}. Similar to FTE, our proposed time encoding is also invariant to time rescaling, as shown in the following proposition with proof provided in Appendix \ref{proof_pro_invariant}.
\begin{proposition}
\label{pro_invariant}
\ours~is invariant to time rescaling.
\end{proposition}

\textbf{Plug-and-Play}: \ours~is designed in a plug-and-play manner, ensuring seamless compatibility with various models and architectures. By producing a $d$-dimensional time embedding vector similar to previous time encodings, it can be easily integrated without requiring significant modifications to existing frameworks. Unlike prior methods that rely on fixed non-linear transformation functions, \ours~employs parameterized and learnable transformations, enabling it to capture additional information and complexity. This design allows \ours~to achieve superior performance even with lower-dimensional time encodings compared to traditional methods (see Section \ref{sec-exp_dim} for experimental results).

\textbf{Interpretability}: Previous time encodings exhibit natural interpretability because they use a fixed non-linear activation function, i.e., the sine function, which has obvious periodicity. Our proposed time encoding uses a learnable non-linear transformation function. However, by examining the learned parameters, we can reconstruct these transformation functions, enabling our method to also achieve strong interpretability. A visualization of our proposed time encoding is provided in Appendix \ref{app_visual}.

\begingroup

\begin{table*}[t]
\centering
\caption{Time series prediction: multivariate long-term forecasting task. The past sequence length is set to 96, while the prediction lengths are \{96, 192, 336, 720\}. The results are reported in terms of MAE, where lower values indicate better performance. HCTE (Hand-Crafted Time Encoding) is a method widely adopted in time series research. FTE stands for Functional Time Encoding. The win rate represents the percentage of cases where \ours~outperforms the HCTE. The best results for each baseline, dataset and prediction length combinations are in \textbf{bold}. 
ETT consists of 4 subsets. Here, we present the average results across these subsets, with the full results provided in Table \ref{tab_ett_mae}.}

\renewcommand{\arraystretch}{0.95}
\setlength{\tabcolsep}{2mm}{
\scriptsize
\begin{tabular}{cc|ccc|ccc|ccc|ccc|ccc|c}
\toprule
\multicolumn{2}{c|}{MAE}                                                           & \multicolumn{3}{c|}{Transformer}        & \multicolumn{3}{c|}{Pyraformer}         & \multicolumn{3}{c|}{NS Trans.}                        & \multicolumn{3}{c|}{MINC}                             & \multicolumn{3}{c|}{TimesNet}                                                         & Win                     \\ \cmidrule{1-17}
\multicolumn{2}{c|}{TE}                                                            & HCTE & FTE            & \ours           & HCTE          & FTE   & \ours           & HCTE & FTE   & \ours                                  & HCTE & FTE   & \ours                                  & HCTE                                 & FTE   & \ours                                  & Rate                    \\ \midrule
\multicolumn{1}{c|}{}                                                        & 96  & 0.797 & 0.803          & \textbf{0.550} & 0.642          & 0.720 & \textbf{0.583} & 0.405 & 0.435 & \textbf{0.377}                        & 0.405 & 0.367 & \textbf{0.350} & 0.355                                 & 0.362 & \textbf{0.352}                        &                         \\
\multicolumn{1}{c|}{}                                                        & 192 & 1.139 & 0.916          & \textbf{0.712} & 0.899          & 0.924 & \textbf{0.738} & 0.445 & 0.478 & \textbf{0.414}                        & 0.445 & 0.423 & \textbf{0.395}                        &  \textbf{0.385} & 0.400 & 0.388                                 &                         \\
\multicolumn{1}{c|}{}                                                        & 336 & 1.119 & 0.938          & \textbf{0.821} & 1.043          & 1.038 & \textbf{0.863} & 0.478 & 0.539 & \textbf{0.449}                        & 0.478 & 0.486 & \textbf{0.448}                        & 0.421                                 & 0.424 & \textbf{0.413} &                         \\
\multicolumn{1}{c|}{\multirow{-4}{*}{\rotatebox[origin=c]{90}{ETT}}}         & 720 & 1.070 & 1.146          & \textbf{0.878} & 1.196          & 1.189 & \textbf{0.959} & 0.526 & 0.557 & \textbf{0.490}                        & 0.526 & 0.561 & \textbf{0.505}                        & 0.455                                 & 0.455 &  \textbf{0.429} & \multirow{-4}{*}{95\%}  \\ \midrule
\multicolumn{1}{c|}{}                                                        & 96  & 0.357 & 0.375          & \textbf{0.347} & 0.376          & 0.375 & \textbf{0.365} & 0.273 & 0.275 & \textbf{0.265}                        & 0.269 & 0.263 &  \textbf{0.254} & 0.272                                 & 0.272 & \textbf{0.267}                        &                         \\
\multicolumn{1}{c|}{}                                                        & 192 & 0.367 & 0.402          & \textbf{0.353} & 0.391          & 0.385 & \textbf{0.372} & 0.286 & 0.292 & \textbf{0.278}                        & 0.285 & 0.278 &  \textbf{0.271} & 0.289                                 & 0.281 & \textbf{0.277}                        &                         \\
\multicolumn{1}{c|}{}                                                        & 336 & 0.370 & 0.425          & \textbf{0.357} & 0.399          & 0.401 & \textbf{0.369} & 0.304 & 0.300 & \textbf{0.293}                        & 0.304 & 0.298 & \textbf{0.294}                        & 0.300                                 & 0.308 &  \textbf{0.291} &                         \\
\multicolumn{1}{c|}{\multirow{-4}{*}{\rotatebox[origin=c]{90}{Electricity}}} & 720 & 0.374 & 0.453          & \textbf{0.363} & 0.390          & 0.394 & \textbf{0.380} & 0.321 & 0.330 & \textbf{0.317}                        & 0.321 & 0.330 & \textbf{0.317}                        & 0.320                                 & 0.363 &  \textbf{0.316} & \multirow{-4}{*}{100\%} \\ \midrule
\multicolumn{1}{c|}{}                                                        & 96  & 0.575 & 0.705          & \textbf{0.547} & \textbf{0.570} & 0.641 & 0.624          & 0.237 & 0.261 & \textbf{0.237}                        & 0.235 & 0.233 & \textbf{0.203} & 0.234                                 & 0.237 & \textbf{0.230}                        &                         \\
\multicolumn{1}{c|}{}                                                        & 192 & 0.747 & 0.791          & \textbf{0.744} & 0.803          & 0.815 & \textbf{0.786} & 0.335 & 0.369 & \textbf{0.319}                        & 0.316 & 0.332 &  \textbf{0.289} & 0.344                                 & 0.339 & \textbf{0.332}                        &                         \\
\multicolumn{1}{c|}{}                                                        & 336 & 0.945 & 1.123          & \textbf{0.879} & 0.903          & 0.991 & \textbf{0.859} & 0.476 & 0.501 & \textbf{0.439}                        & 0.407 & 0.472 & \textbf{0.402} & 0.448                                 & 0.472 & \textbf{0.446}                        &                         \\
\multicolumn{1}{c|}{\multirow{-4}{*}{\rotatebox[origin=c]{90}{Exchange}}}    & 720 & 1.329 & 1.147          & \textbf{1.066} & 1.075          & 1.046 & \textbf{0.938} & 0.769 & 0.901 &  \textbf{0.612} & 0.658 & 0.710 & \textbf{0.622}                        & 0.746                                 & 0.756 & \textbf{0.751}                        & \multirow{-4}{*}{95\%}  \\ \midrule
\multicolumn{1}{c|}{}                                                        & 96  & 0.422 & 0.257          & \textbf{0.245} & 0.303          & 0.296 & \textbf{0.267} & 0.223 & 0.222 & \textbf{0.221}                        & 0.229 & 0.258 & \textbf{0.225}                        & 0.220                                 & 0.221 &  \textbf{0.215} &                         \\
\multicolumn{1}{c|}{}                                                        & 192 & 0.523 & 0.308          & \textbf{0.295} & 0.336          & 0.317 & \textbf{0.311} & 0.285 & 0.271 & \textbf{0.260}                        & 0.281 & 0.306 & \textbf{0.261}                        & 0.261                                 & 0.263 &  \textbf{0.253} &                         \\
\multicolumn{1}{c|}{}                                                        & 336 & 0.607 & \textbf{0.355} & 0.365          & 0.403          & 0.377 & \textbf{0.349} & 0.338 & 0.321 & \textbf{0.308}                        & 0.331 & 0.335 &  \textbf{0.295} & 0.306                                 & 0.302 & \textbf{0.299}                        &                         \\
\multicolumn{1}{c|}{\multirow{-4}{*}{\rotatebox[origin=c]{90}{Weather}}}     & 720 & 0.690 & 0.459          & \textbf{0.429} & 0.434          & 0.417 & \textbf{0.415} & 0.410 & 0.357 & \textbf{0.349}                        & 0.356 & 0.387 & \textbf{0.339} & 0.359                                 & 0.350 & \textbf{0.348}                        & \multirow{-4}{*}{100\%} \\ \midrule
\multicolumn{2}{c|}{Win Rate}                                                      & \multicolumn{3}{c|}{100\%}              & \multicolumn{3}{c|}{94\%}               & \multicolumn{3}{c|}{100\%}                            & \multicolumn{3}{c|}{100\%}                            & \multicolumn{3}{c|}{94\%}                                                             & 98\%                    \\ \bottomrule
\end{tabular}

}
\label{tab_ts_mae}
\end{table*}

\endgroup



\subsection{Use of Time Encoding}
In time series forecasting research, time embeddings calculated by time encoding modules are usually directly added to feature embeddings and fed into the attention mechanism or Transformer \cite{attention}. As a result, they typically share the same dimensions as the feature embeddings:
\begin{equation}
\mathbf{x} = \textbf{TokenEncode}(\mathbf{x}) + \textbf{TE}(t) \in \mathbb{R}^d.
\end{equation}
Here, $\mathbf{x}$ represents the input, $\textbf{TokenEncode}$ denotes a token encoding function, $\textbf{TE}$ denotes the time encoding, and $d$ represents the dimension of both feature embeddings and time embeddings.

In dynamic graph representation learning research, time embeddings are usually concatenated with node features and edge features as part of the input. This allows for more flexibility in the choice of time embedding dimensions:
\begin{equation}
\mathbf{x} = \textbf{Node Features} \| \textbf{Edge Features} \| \textbf{TE}(t) \in \mathbb{R}^{d_n+d_e+d}.
\end{equation}
Here, $\|$ denotes the concatenation operation, while $d_n$, $d_e$, and $d$ represent the dimensions of node features, edge features, and time embeddings, respectively.

\section{Experiments}
\label{sec-experiments}


\subsection{Time as the Only Input}

To evaluate the performance of the time encoding method in scenarios where the only input is time, and to compare different time representations while minimizing the influence of extraneous variables, we follow \cite{time2vec} and create a sequential (event-based) MNIST dataset \cite{fatahi2016evt_mnist, campos2017skip, bellec2018long} and conduct image classification task (more details are shown in Appendix \ref{app_time_only_exps}). 

\begin{figure}[h]
    \centering
    \begin{subfigure}[b]{0.46\linewidth}
        \includegraphics[width=1\linewidth]{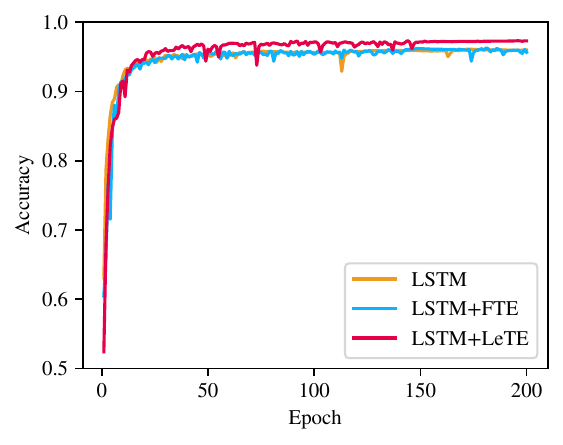}
        \caption{Testing Accuracy}
    \end{subfigure}
    \hspace{1pt}
    \begin{subfigure}[b]{0.46\linewidth}
        \includegraphics[width=1\linewidth]{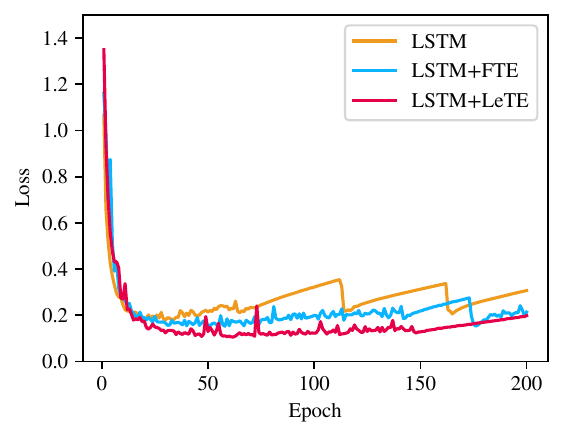}
        \caption{Testing Loss}
    \end{subfigure}

\caption{Time as the only input: Comparison of time encodings on sequential MNIST.}
\label{fig:mnist}
\end{figure}

We then apply an LSTM with a 32-dimensional learnable embedding for the time input and compare it to models where the FTE or our \ours~is used for encoding time. The results, shown in Figure \ref{fig:mnist}, indicate that the FTE achieves testing accuracy comparable to that of the LSTM without any time encoding method applied. However, our \ours~achieves significantly higher image classification accuracy. This simple experiment demonstrates that \ours~can efficiently encode time information for models. Next, we present experiments applying \ours~to time series tasks, dynamic graph tasks, and real-world applications.

\begin{table*}[h]

\caption{Dynamic graph link prediction task: The results are reported in AP, where higher values indicate better performance. The better results are in \textbf{bold}. Here, we present the top-performing results across variations of \ours, with the full results provided in Table \ref{tab_tg_types_ap}. FTE represents Functional Time Encoding which is commonly used in dynamic graph research.}

\renewcommand{\arraystretch}{1.0}
\setlength{\tabcolsep}{2.8mm}{
\scriptsize
\begin{tabular}{c|c|cc|cc|cc|cc}
\toprule
                                                                       & AP         & \multicolumn{2}{c|}{Wikipedia}            & \multicolumn{2}{c|}{Reddit}               & \multicolumn{2}{c|}{MOOC}                 & \multicolumn{2}{c}{LastFM}                \\ \cmidrule{2-10} 
                                                                       & TE         & Transductive        & Inductive           & Transductive        & Inductive           & Transductive        & Inductive           & Transductive        & Inductive           \\ \midrule
\multirow{2}{*}{TGAT}                                                  & FTE & 96.95 ± 0.24          & 96.33 ± 0.26          & 98.53 ± 0.04          & 97.01 ± 0.05          & 85.34 ± 0.19          & 84.94 ± 0.04          & 72.73 ± 0.11          & 77.78 ± 0.13          \\
                                                                       & \ours & \textbf{97.82 ± 0.09} & \textbf{97.34 ± 0.08} & \textbf{98.56 ± 0.01} & \textbf{97.05 ± 0.06} & \textbf{88.31 ± 0.10} & \textbf{88.37 ± 0.12} & \textbf{76.22 ± 0.25} & \textbf{81.32 ± 0.14} \\ \midrule
\multirow{2}{*}{TGN}                                                   & FTE & 98.45 ± 0.06          & 97.83 ± 0.04          & 98.63 ± 0.06          & 97.50 ± 0.07          & 89.15 ± 1.60          & 89.04 ± 1.17          & 77.07 ± 3.97          & 81.45 ± 4.29          \\
                                                                       & \ours & \textbf{98.78 ± 0.07} & \textbf{98.19 ± 0.09} & \textbf{98.74 ± 0.00} & \textbf{97.65 ± 0.04} & \textbf{91.41 ± 0.55} & \textbf{90.87 ± 0.83} & \textbf{83.64 ± 2.00} & \textbf{87.55 ± 1.88} \\ \midrule
\multirow{2}{*}{TCL}                                                   & FTE & 96.47 ± 0.16          & 96.22 ± 0.17          & 97.53 ± 0.02          & 94.09 ± 0.07          & 82.38 ± 0.24          & 80.60 ± 0.22          & 67.27 ± 2.16          & 73.53 ± 1.66          \\
                                                                       & \ours & \textbf{98.19 ± 0.04} & \textbf{97.89 ± 0.03} & \textbf{97.78 ± 0.03} & \textbf{94.99 ± 0.07} & \textbf{84.24 ± 0.10} & \textbf{82.72 ± 0.12} & \textbf{76.08 ± 0.79} & \textbf{80.68 ± 0.70} \\ \midrule
\multirow{2}{*}{\begin{tabular}[c]{@{}c@{}}DyG-\\ Former\end{tabular}} & FTE & 99.03 ± 0.02          & 98.59 ± 0.03          & 99.22 ± 0.01          & 98.84 ± 0.02          & 87.52 ± 0.49          & 86.96 ± 0.43          & 93.00 ± 0.12          & 94.23 ± 0.09          \\
                                                                       & \ours & \textbf{99.13 ± 0.02} & \textbf{98.73 ± 0.00} & \textbf{99.24 ± 0.01} & \textbf{98.86 ± 0.01} & \textbf{88.70 ± 0.21} & \textbf{88.39 ± 0.15} & \textbf{93.64 ± 0.10} & \textbf{94.69 ± 0.12} \\ \bottomrule
\end{tabular}
}

\label{tab_tg_ap}
\end{table*}

\subsection{Experiments on Time Series Tasks}
For time series forecasting tasks, we select 5 baseline models where we can directly replace the time encoding methods with \ours: vanilla Transformer \cite{attention}, Pyraformer \cite{pyraformer}, Non-stationary Transformer \cite{nonstransformer}, MICN \cite{MICN}, and TimesNet \cite{timesnet}. 
We conduct long-term forecasting tasks on these baseline models using 4 datasets: 
ETT, Weather, Exchange \cite{exchange}, and Electricity, 
covering various real-world scenarios. Implementation details and introductions to baselines and datasets are provided in Appendix \ref{app_ts_exps}. We apply \ours~and adjust the hyperparameter $p$ in all experiments to capture more comprehensive time information. 
We report the results in the multivariate setting, as shown in Table \ref{tab_ts_mae} and \ref{tab_ts_mse}. 
Because the time embeddings need to be added to the feature embeddings, they must have the same dimensions as the feature embeddings. 
Baseline models commonly apply hand-crafted time encoding (HCTE) with Date-Time Format inputs (e.g., ISO 8601 format, YYYY-MM-DD HH:mm:ss). Since our method, like FTE, takes UNIX timestamps as input, we include FTE in our experiments for comparison. In our approach, we transform the Date-Time Format timestamps into UNIX timestamps, encode them with our proposed \ours~, and feed the resulting time embeddings into the models in the same manner as the baselines.
In this context, the input consists of absolute timestamps, and the time encoding can therefore be regarded as an \textit{absolute time encoding}.

From the experimental results, we observe that: (1) When applying different time encoding methods to baseline models for time series forecasting, \ours~outperforms the benchmark in most cases, achieving an average win rate of 98\% on MAE (Mean Absolute Error) and 95\% on MSE (Mean Squared Error) across all baseline, dataset, and prediction length combinations, highlighting the effectiveness of the proposed time encoding. This demonstrates that our method can be seamlessly transferred to time series models, reliably achieving strong performance. (2) The improvements on baselines are considerable. For instance, applying \ours~to the Transformer model reduces the average MAE and MSE across all datasets by 25.1\% and 46.5\%, respectively. This illustrates that our method can be applied to various time series forecasting models, consistently achieving strong performance. (3) FTE can occasionally outperform benchmarks; however, it also fails in many cases, whereas \ours~steadily outperforms benchmarks in such situations. 
This demonstrates our method’s capability to model diverse time patterns, including periodic, non-periodic, and mixed, highlighting its generalizability across different models and data.



\subsection{Experiments on Dynamic Graph Tasks}
FTEs are widely used in dynamic graph representation learning models. Representative works include TGAT \cite{tgat}, TGN \cite{tgn}, TCL \cite{tcl}, and DyGFormer \cite{dygformer}. 
Thus, we apply these models as baselines and replace their time encodings with our \ours. 
We conduct link prediction experiments on 4 real-world datasets: Wikipedia, Reddit, MOOC, and LastFM \cite{jodie}. The details of the implementation, baseline methods and datasets are in Appendix \ref{app_dyg_exps}. The results are reported in both transductive and inductive settings, as shown in Tables \ref{tab_tg_ap} and \ref{tab_tg_auc}. In this context, the time encoding module takes the relative time difference between the current edge and the most recent previous edge, and can therefore be regarded as a \textit{relative time encoding}.

As shown in the experimental results, our proposed \ours~surpasses the benchmark results on all combinations of baselines and datasets, regardless of transductive or inductive settings, achieving state-of-the-art (SOTA) performance. 
This strongly demonstrates the effectiveness of our proposed time encoding and highlights its potential for improving the representation learning of dynamic graphs.
The dimensions for the main experiments are set to 100, following the original settings in previous work \cite{tgn, dygformer}. However, since time embeddings are concatenated with node and edge features in dynamic graph models, this provides significant flexibility in setting their dimensions. We compare the effects of time embeddings with different dimensions in Section \ref{sec-exp_dim}. A comparison and analysis of the variations of \ours~are also provided in Appendix \ref{app_exp_t2v_types}.

\begin{figure}[!t]
    \centering
    \includegraphics[width=1\linewidth]{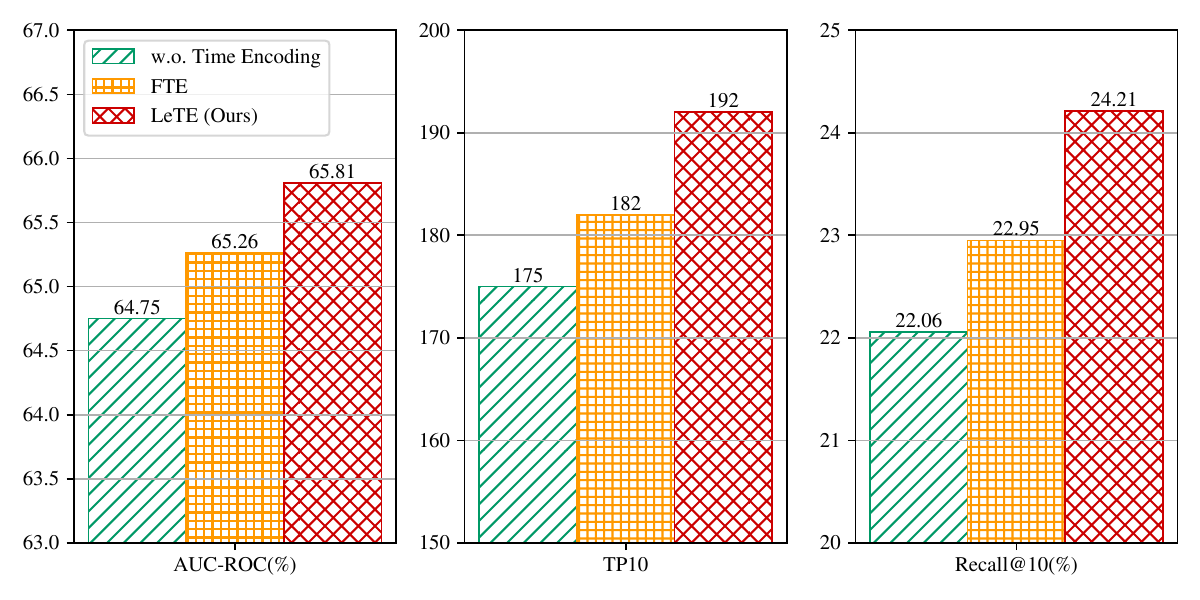}

\caption{Results evaluated by AUC-ROC, TP10 and Recall@10 on real business datasets.}
\label{fig:business}


\end{figure}

\subsection{Experiments on Real-World Application}
Time information plays a crucial role in many real-world fields. We apply our proposed \ours~in a real-world financial risk control scenario to demonstrate its effectiveness in practical applications. In financial risk control, a user's historical transaction data is typically used to predict their credit risk, which can be framed as a classification problem based on historical transaction information. However, in this scenario, users' transaction behaviors often exhibit a combination of complex periodic and non-periodic patterns. For instance, users may regularly receive salary deposits and purchase daily necessities, whereas peer-to-peer transfers may lack strong periodicity. Using financial risk control data from Tencent Mobile Payment \footnote{The data used in these experiments are properly sampled only for testing purposes and does not imply any commercial information. All users' private information is removed from the dataset. Moreover, the experiments were conducted locally on Tencent's server by formal employees who strictly followed data protection regulations.}, we conduct comparative experiments without time information, with the FTE, and with \ours~to encode time information. The backbone model treats the time embedding as a feature, concatenates it with the user's raw features, and takes the concatenated features as input. The objective is to use users' historical transaction data to predict whether they have default risk. Details of the dataset are provided in Appendix \ref{app_application_dataset}. The results are presented in Figure \ref{fig:business}. The results indicate that the model without time encoding performs the worst, as it completely ignores time information. With FTE, the periodicity of user behavior at different frequencies is captured, resulting in improved performance compared to the case without time information. Using \ours~yields the best performance, as our time encoding effectively models periodic, non-periodic and mixed patterns in a more general manner.


\subsection{Dimensions of Time Encoding}
\label{sec-exp_dim}

\begin{figure}[t]
    \centering
    \begin{subfigure}[b]{0.46\linewidth}
        \includegraphics[width=1\linewidth]{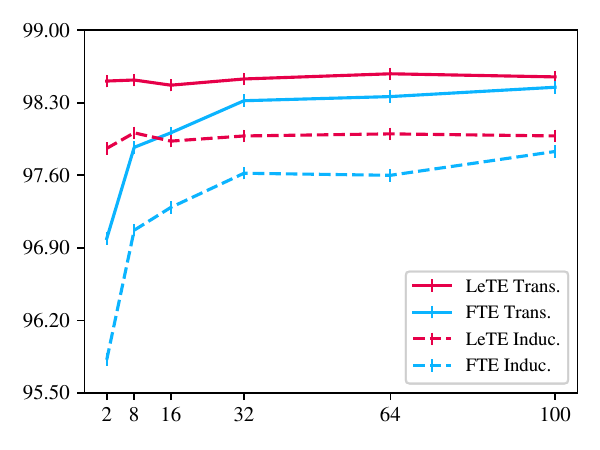}
        \caption{Wikipedia/TGN}
    \end{subfigure}
    \hspace{1pt}
    \begin{subfigure}[b]{0.46\linewidth}
        \includegraphics[width=1\linewidth]{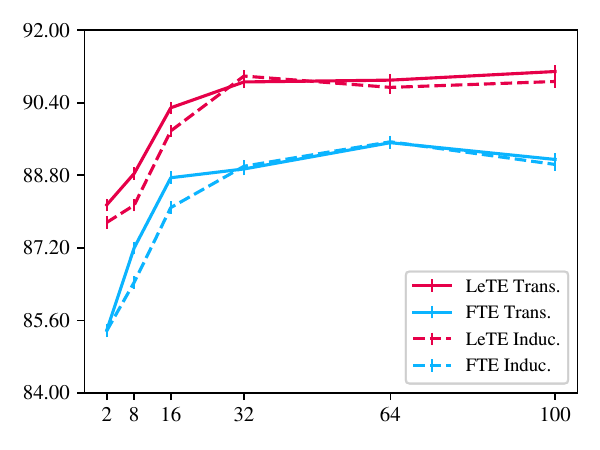}
        \caption{MOOC/TGN}
    \end{subfigure}

\caption{Average Precision results comparing different dimensions of the FTE and Spline-based \ours~on Wikipedia/TGN and MOOC/TGN.}
\label{fig:dim_ap_tgn}


\end{figure}

Compared to the previous FTE methods, our \ours~takes a step forward by making the non-linear transformation learnable, thereby generalizing the time encoding. Since part of the information from the data is captured by the learnable non-linear transformation, we hypothesize that using lower-dimensional \ours~may still outperform the FTE (which relies on a fixed non-linear transformation). Therefore, we conduct experiments with lower-dimensional time encodings. Since the dimensionality of time encoding in dynamic graph tasks is more flexible, we conduct experiments on dynamic graph link prediction tasks, with the results presented in Figures \ref{fig:dim_ap_tgn}, \ref{fig:dim_auc_tgn}, and \ref{fig:dim_dyg}. As illustrated in the results, models using the FTE suffer from severe performance degradation as the dimension decreases, whereas models with \ours~demonstrate more stable performance and consistently outperform those using the FTE, even at lower dimensions. Notably, models using \ours~with significantly lower dimensions (e.g., 2, 8 or 16) outperform models with the 100-dimensional FTE. This demonstrates the effectiveness and generalizability of our method.

\subsection{Additional Experiments}



We provide the complete experimental results for the experiments mentioned in the main text in Appendix \ref{app_additional_exp_results}. Additionally, we conduct further experiments to analyze the complex time patterns in real-world data (cf. Appendix \ref{app_periodic_and_non}); compare different variants of \ours~(cf. Appendix \ref{app_exp_t2v_types}); illustrate the interpretability of \ours~(cf. Appendix \ref{app_visual}); demonstrate \ours’s ability to simultaneously capture diverse time patterns, including periodic, non-periodic, and mixed ones (cf. Appendix \ref{app_p_and_nonp}); and assess \ours’s capability to fit various functions (cf. Appendix \ref{app_fitting_ability}).

\section{Conclusion}
\label{sec-conclusion}

In this paper, we propose a effective time encoding method—\oursfull~(\ours)—designed to accept both absolute timestamps and relative time differences as inputs, depending on the specific requirements of different models, enabling it to function as either an absolute or a relative time encoding method. 
Through comprehensive analysis, we demonstrate that our proposed \ours~is capable of modeling diverse and complex time patterns, including periodic, non-periodic, and mixed patterns. It is invariant to time rescaling, sufficiently simple for integration with various backbone models, and exhibits good interpretability and dimensional efficiency. 
Extensive experiments on event-based image classification, time-series forecasting tasks, dynamic graph link prediction tasks, and real-world financial risk control applications demonstrate the superior performance and generalizability of our method across various application scenarios.

\section*{Acknowledgements}
This work is partially supported by the Noncommunicable Chronic Diseases-National Science and Technology Major Project (NO. 2024ZD0532400 and NO. 2024ZD0532403). This work is sponsored by the Tencent Rhino-Bird Focused Research Program. This work is partially supported by NSF through grant IIS-2106972.


\section*{Impact Statement}
This paper presents work whose goal is to advance the field of Machine Learning. There are many potential societal consequences of our work, none which we feel must be specifically highlighted here.

\bibliography{references}
\bibliographystyle{icml2025}

\newpage
\appendix
\onecolumn

\section{Related Work}
\label{sec-related_work}
Currently, commonly used time encoding methods or strategies for modeling temporal information can be broadly categorized into two types: Hand-Crafted Time Encodings (HCTE) and Functional Time Encodings (FTE).

HCTE involves manually designed temporal encodings tailored to specific downstream tasks. These methods rely on specific design choices and incorporate various inductive biases to capture fixed periodic patterns, constructing hand-crafted temporal features. These features are typically fed into models such as RNNs \cite{rnn} or sequential architectures \cite{pyraformer, nonstransformer, MICN, timesnet} to meet specific modeling requirements \cite{choi2016doctor,baytas2017patient, kwon2018retainvis}, as illustrated in Figure \ref{fig_t2v}(a). Such approaches are often employed to address particular challenges in time series tasks.

Additionally, some methods in this category integrate time encoding directly with attention mechanisms \cite{attention}, simplifying temporal modeling by adopting position encoding strategies \cite{gehring2017convolutional}. Others embed discrete events into a continuous vector space to better capture event contexts in attention-based models \cite{bengio2013representation, li2017time}. These methods, while effective, are often limited to representing fixed or narrowly defined temporal patterns.

FTE represents an advanced and generalized version of time encoding, designed to overcome part of the limitations of Hand-Crafted Time Encodings. Two representative works in this category are functional time representation, proposed by \cite{xu2019time}, and Time2Vec, proposed by \cite{time2vec}, as shown in Figure \ref{fig_t2v}(b). Importantly, the previously mentioned position encoding methods integrated with attention mechanisms can be considered a simplified version of FTE.

Both functional time representation and Time2Vec adopt similar implementation methods, which resemble a one-dimensional-to-d-dimensional MLP with a specially designed trigonometric non-linear activation function. In Time2Vec, experiments comparing different non-linear activation functions demonstrate that the sine function performs best across various downstream tasks. Despite limitations in modeling restricted aspects of time, FTE is widely adopted in dynamic graph representation learning due to its ease of application and effectiveness \cite{tgn, tiger, rdgsl, speed, dygformer, chen2024prompt, zhang2024towards, chen2024dtformer, zhang2025unifying}.

Time encodings can be directly applied to sequential models such as RNNs and LSTMs \cite{graves2012long}, or easily integrated into attention-based architectures. In time series forecasting, for instance, many models now employ transformer-based structures \cite{attention}, where time encoding is often treated similarly to position encoding. 
This is usually achieved by adding it to the input of the attention mechanism
\cite{pyraformer, nonstransformer, timesnet}.

Dynamic graph representation learning models also require precise temporal modeling. For example, TGAT \cite{tgat} directly replaces position encoding with functional time representation within its attention mechanism. Subsequent methods, such as TGN and TIGER \cite{tgn, tiger}, have adopted similar approaches. DyGFormer \cite{dygformer}, which applies a Transformer to dynamic graph representation learning, uses the same encoding method by concatenating it with node and edge features before processing them with a Transformer-based model.

\section{Methods of Parameterize Continuous Functions}
\label{app_fourier_spline}
\subsection{Fourier Series Expansion}
\label{app_fourier}
A function $f(x)$ that is periodic with period $T$ and satisfies certain conditions (Dirichlet conditions) can be represented as a Fourier Series. This series represents $f(x)$ as an infinite sum of sines and cosines (or, equivalently, complex exponentials) with specific coefficients. The series takes the form:
\begin{equation}
    f(x) = a_0 + \sum_{n=1}^{\infty} \left( a_n \cos \frac{2\pi n x}{T} + b_n \sin \frac{2\pi n x}{T} \right).
\end{equation}
Here, $a_0$ is the average value of the function over one period, and $a_n$ and $b_n$ are Fourier coefficients that can be calculated by integrating $f(x)$ over the interval $[0, T]$.

These coefficients are given by:
\begin{equation}
    a_n = \frac{2}{T} \int_0^T f(x) \cos \frac{2\pi n x}{T} \, dx,
\end{equation}
\begin{equation}
    b_n = \frac{2}{T} \int_0^T f(x) \sin \frac{2\pi n x}{T} \, dx.
\end{equation}
Under these conditions, the Fourier Series converges to $f(x)$ at all points where $f$ is continuous and converges to the average of the left-hand and right-hand limits at points of discontinuity.

\subsection{KAN and Spline Functions}
\label{app_spline}
The Kolmogorov–Arnold Theorem \cite{kolmogorov1961representation, kolmogorov1957representation, braun2009constructive} states that for any continuous multivariate function $f(x_1, x_2, \ldots, x_n)$ on the unit cube $[0, 1]^n$, there exist continuous functions $\phi_{i}$ and $\psi_{ij}$ such that:
\begin{equation}
    f(x_1, x_2, \ldots, x_n) = \sum_{i=1}^{2n+1} \phi_{i} \left( \sum_{j=1}^n \psi_{ij}(x_j) \right),
\end{equation}
where $\phi_i$ are continuous functions of a single variable, enabling dimensionality reduction, and $\psi_{ij}$ are continuous functions mapping each input variable $x_j$ to a single output, contributing to the superposition structure. This theorem implies that every continuous function of multiple variables can be represented as a sum of compositions of univariate functions.

Building on the Kolmogorov–Arnold Theorem and the advantages of splines for function fitting, Liu et al. propose using splines to construct learnable non-linear activation functions for neural networks \cite{kan}. We briefly introduce B-splines here. Given a knot vector $T = { t_0, t_1, \dots, t_m }$ with non-decreasing values, the basis functions $N_{i,p}(x)$ for a B-spline of degree $p$ are defined recursively as follows:
For degree $p=0$:
\begin{equation}
    N_{i,0}(x) =
    \begin{cases}
        1 & \text{if } t_i \leq x < t_{i+1} \\
        0 & \text{otherwise}.
    \end{cases}
\end{equation}
For higher degrees $p > 0$:
\begin{equation}
    N_{i,p}(x) = \frac{x - t_i}{t_{i+p} - t_i} N_{i,p-1}(x) + \frac{t_{i+p+1} - x}{t_{i+p+1} - t_{i+1}} N_{i+1,p-1}(x).
\end{equation}
The B-spline curve $C(x)$ of degree $p$ with control points $\{ P_0, P_1, \dots, P_n \}$ is given by:
\begin{equation}
    C(x) = \sum_{i=0}^n N_{i,p}(x) P_i.
\end{equation}
Here, $N_{i,p}(x)$ are the B-spline basis functions of degree $p$ , and $P_i$ are the control points that influence the shape of the curve.

\section{Proofs}
\label{app_proofs}

\subsection{Proof of Proposition \ref{pro_unify_te}}
\label{proof_pro_unify_te}
\subsubsection{Details of Functional Time Representation and Time2Vec}

FTR is designed to use the time difference $t = t_i - t_j$, where $0 \le t_j \le t_i \le t_{\text{max}}$, as input. For the input time difference, a learnable frequency parameter is first applied. Next, a non-linear transformation is applied, using the cosine function on the odd dimensions and the sine function on the even dimensions. This method is mathematically represented as follows:
\begin{equation}
\label{eq_functional_tr}
    \textbf{TE}(t)[i]=
    \begin{cases}
        \cos{(\omega_i t)}, & \text{if~~$i$ is odd}, \\
        \sin{(\omega_i t)}, & \text{if~~$i$ is even},
    \end{cases}
\end{equation}
where $d$ is the dimension of the time encoding, $1 \le i \le d$, and $\omega_i$ are learnable parameters representing the frequency of the trigonometric functions. 
Since this time encoding uses time differences as input, it can be considered as a \textit{relative time encoding}.

T2V is designed to use timestamps $t$ as input. A linear transformation is applied to the first dimension to capture non-periodic time patterns. For the remaining dimensions, a linear transformation is followed by a sine-based non-linear transformation to model periodic time patterns: Mathematically, this method is represented as follows:
\begin{equation}
    \textbf{TE}(t)[i]=
    \begin{cases}
        \omega_i t + \varphi_i, & \text{if~~$i=1$}, \\
        \sin{(\omega_i t + \varphi_i)}, & \text{if~~$2\leq i \leq d$},
    \end{cases}
\end{equation}
where $\textbf{TE}(t)[i]$ is the $i^{th}$ element of the time encoding, and $\omega_i$ and $\varphi_i$ are learnable parameters representing frequency and phase-shift of the sine function, respectively.
Since this time encoding takes timestamps as input, it can be considered as an \textit{absolute time encoding}.

\subsubsection{Proof of Proposition \ref{pro_unify_te}}

\proof 

Since Equation \eqref{eq_functional_tr} can be written as $\textbf{TE}(t) = \left[ \cos(\omega_1 t), \sin(\omega_1 t), \ldots, \cos(\omega_d t), \sin(\omega_d t) \right]$, we show that the vector $\left[ \cos(\omega_1 t), \sin(\omega_1 t), \ldots, \cos(\omega_d t), \sin(\omega_d t) \right]$ can be expressed in the form $\sin(\omega_i t + \varphi_i)$ with suitable phase shifts $\varphi_i$. Recall the trigonometric identity:
\begin{equation}
    \cos(\theta) = \sin\left( \theta + \frac{\pi}{2} \right).
\end{equation}
Applying this identity, each cosine term in the vector can be rewritten as:
\begin{equation}
    \cos(\omega_i t) = \sin\left( \omega_i t + \frac{\pi}{2} \right), \qquad \text{for} \quad i = 1, 2, \ldots, d.
\end{equation}
The sine terms are already in the desired form with a zero phase shift:
\begin{equation}
    \sin(\omega_i t) = \sin\left( \omega_i t + 0 \right), \qquad \text{for} \quad i = 1, 2, \ldots, d.
\end{equation}
With these transformations, the original vector becomes:
\begin{equation}
    \textbf{TE}(t) = \left[ \sin\left( \omega_1 t + \frac{\pi}{2} \right), \sin\left( \omega_1 t + 0 \right), \ldots, \sin\left( \omega_d t + \frac{\pi}{2} \right), \sin\left( \omega_d t + 0 \right) \right], 
\end{equation}
or equivalently,
\begin{equation}
    \textbf{TE}(t) = \left[ \sin\left( \omega_i t + \varphi_i \right) \right]_{i=1}^{2d},
\end{equation}
where the phase shifts $\varphi_i$ are defined as follows:
\begin{align*}
\varphi_i = \begin{cases}
\frac{\pi}{2}, & \text{if } i \text{ is odd} \\
0, & \text{if } i \text{ is even.}
\end{cases}
\end{align*}
\endproof

\subsection{Continued Proof of Proposition \ref{pro_special_case}}
\label{proof_pro_special_case}
\proof
The function $\sin(\theta)$ is continuous and infinitely differentiable (i.e., $C^\infty$) on $\mathbb{R}$. Thus, it is continuous on any closed interval $[a, b]$.

B-spline basis functions of degree $k$ form a basis for the space of piecewise polynomial functions of degree $k$ with continuity $C^{k-1}$ at the knots. By the Weierstrass Approximation Theorem, any continuous function on a closed interval can be uniformly approximated by polynomials to any desired degree of accuracy.

Since B-splines are piecewise polynomials, they can uniformly approximate any continuous function on $[a, b]$. Specifically, for any $\epsilon > 0$, there exists a linear combination of B-spline basis functions that approximates $\sin(\theta)$ within $\epsilon$ over $[a, b]$.

To build the approximation of $\sin(\omega_i t + \varphi_i)$ using B-spline basis functions, we first select a knot vector $\mathbf{T} = \{ t_0, t_1, \ldots, t_{n+k+1} \}$ that partitions the interval $[a, b]$ appropriately. The choice of the knot vector determines the placement and spacing of the knots, which in turn affect the flexibility and local support of the B-spline basis functions.

Next, we choose the degree $k$ of the B-spline basis functions based on the desired smoothness and approximation quality. A higher degree allows for smoother basis functions, potentially improving the approximation at the cost of increased computational complexity.

With the knot vector and degree specified, we generate the B-spline basis functions $\{ B_j(\theta) \}_{j=1}^{M}$ of degree $k$ using standard recursive definitions. These basis functions possess local support and satisfy the partition of unity property, making them suitable for approximating functions over $[a, b]$.

To determine the coefficients $c_{ij}$ that yield the best approximation of $\sin(\omega_i t + \varphi_i)$, we formulate an optimization problem. Specifically, we set up a minimization problem that seeks to minimize the squared difference between the sine function and the weighted sum of B-spline basis functions over the interval $[a, b]$:
\begin{equation}
    \min_{c_{i,1}, c_{i,2}, \ldots, c_{i,M}} \int_{a}^{b} \left[ \sin(\omega_i t + \varphi_i) - \sum_{j=1}^{M} c_{i,j} \, B_j(\omega_i t + \varphi_i) \right]^2 dt.
\end{equation}
This minimization problem is a standard least squares problem, where the objective is to find the coefficients $c_{i,j}$ that minimize the integral of the squared error. Solving this problem can be accomplished using numerical methods such as the normal equations or singular value decomposition, leading to the optimal coefficients for the approximation.

By leveraging the properties of B-spline basis functions and the Weierstrass Approximation Theorem, we can assert that, for any $\epsilon > 0$, there exists a sufficiently large $M$ and appropriate coefficients $c_{i,j}$ such that:
\begin{equation}
    \sup_{t \in [a, b]} \left| \sin(\omega_i t + \varphi_i) - \sum_{j=1}^{M} c_{i,j} \, B_j(\omega_i t + \varphi_i) \right| < \epsilon.
\end{equation}

This inequality indicates that the maximum deviation between the sine function and its B-spline approximation over $[a, b]$ is less than $\epsilon$, satisfying the condition of uniform approximation.

Since $\epsilon > 0$ is arbitrary, we can make the approximation as accurate as desired by increasing $M$ and choosing appropriate coefficients $c_{i,j}$. Therefore, the sine function $\sin(\omega_i t + \varphi_i)$ can be represented as a sum of B-spline basis functions, making it a special case of Equation \eqref{eq_bspline_t2vp} in the limit as $M \to \infty$.
\endproof

\subsection{Proof of Proposition \ref{pro_invariant}}
\label{proof_pro_invariant}
A class $\mathcal{C}$ of models is considered invariant to time rescaling if, for any model $\mathcal{M}_1 \in \mathcal{C}$ and any scalar $\alpha > 0$, there exists a model $\mathcal{M}_2 \in \mathcal{C}$ that responds to $\alpha t$ (where $t$ is scaled by $\alpha$) in the same way that $\mathcal{M}_1$ responds to the original $t$ values. We provide the following proof to show that \ours~is invariant to time rescaling.

\proof
Consider time encoding $\mathcal{M}_1$, mapped by \ours:
\begin{equation}
    \textbf{\ours}(t)[i] = \phi_i(\omega_i t + \varphi_i).
\end{equation}
If we replace $t$ with $\alpha \cdot t$ (where $\alpha > 0$), the time encoding updates as follows:
\begin{equation}
    \textbf{\ours}(\alpha \cdot t)[i] = \phi_i(\omega_i (\alpha \cdot t) + \varphi_i).
\end{equation}
To preserve the behavior of the original model $\mathcal{M}_1$ under time rescaling, consider a new time encoding $\mathcal{M}_2$ with adjusted frequencies $\omega'_i = \frac{\omega_i}{\alpha}$. With this frequency adjustment, $\mathcal{M}_2$ behaves identically to $\mathcal{M}_1$ on $\alpha \cdot t$, demonstrating that \ours~is invariant to time rescaling.
\endproof

\section{Implementation Details of \ours}
\label{app_imp}
Previous implementations of FTEs can be summarized as inputting a timestamp or time difference between events into a single-layer MLP with a fixed trigonometric function as the non-linear activation function. Inspired by this, our method can be viewed as making the fixed activation function learnable by parameterizing it with a Fourier series expansion or B-spline functions. Following KAN \cite{kan}, which makes activation functions in deep learning models trainable, we use a similar implementation method.

\subsection{Fourier-based \ours}
\label{app_imp_fourier}
The implementation of Fourier-based \ours~is straightforward and is given by:
\begin{equation}
\phi_j(x) = \sum_{i=1}^{D} \sum_{m=1}^{K} \bigl( W^{(\cos)}_{j,i,m} \cos(m x_i) + W^{(\sin)}_{j,i,m} \sin(m x_i) \bigr) + b_j,
\end{equation}
where $i = 1, 2, \ldots, D$ indexes the input dimension, $j = 1, 2, \ldots, M$ indexes the output dimension (with $D = M$ in our method), and $m = 1, 2, \ldots, K$ indexes the Fourier frequencies. Here, $K$ is a hyper-parameter that determines the grid size. The parameters $\mathbf{W}^{(\cos)} \in \mathbb{R}^{M \times D \times K}$, $\mathbf{W}^{(\sin)} \in \mathbb{R}^{M \times D \times K}$, and $\mathbf{b} \in \mathbb{R}^{M}$ are learnable weights and biases.

\subsection{Spline-based \ours}
\label{app_imp_spline}
By using B-spline functions, we make the $\phi_i$ functions in Equation \eqref{eq_t2vp} learnable as follows:
\begin{equation}
    \phi_i(x) = b_i(x)+\text{spline}_i(x),
\end{equation}
\begin{equation}
    b(x)=\text{Tanh}(x) = \frac{\sinh(x)}{\cosh(x)} = \frac{e^x - e^{-x}}{e^x + e^{-x}},
\end{equation}
\begin{equation}
    \text{spline}_i(x) = \sum_j c_{ij} B_j(x),
\end{equation}
where $c_{ij}$ are learnable. Unlike the original KAN, we use Tanh as the basis function here, as we found it performs better in practice.



\section{More Details for Experimental Setting}

\subsection{Details for ``Time as the Only Input'' Experiments}
\label{app_time_only_exps}
Following \cite{time2vec}, We generate a sequential event-based version of MNIST by flattening the images and recording the positions of pixels with intensities greater than a threshold (0.9 in our experiment). After this transformation, each image is represented as an array of increasing numbers, such as $[t_1, t_2, t_3, \dots, t_m]$. These values are treated as event times and can be used for image classification task.
The backbone model we used is a 128-dimensional LSTM, with a batch size of 512, aligned with the settings in \cite{time2vec}.

\subsection{Time Series Baselines and Datasets}
\label{app_ts_exps}

\subsubsection{Experiment Implementation Details}
Our experiments setting on time series tasks aligns with the definition of long-term forecasting. The baseline results for Transformer and Pyraformer are based on our implementation, while the results for the other baselines are taken from their original papers. Baseline models typically use a hand-crafted time encoding method that applies date and timestamps to represent various time features—including minutes, hours, weekdays, days, and months. The mapped vectors are then added together and added to the feature embeddings and fed into the models. The results are evaluated using MAE (Mean Absolute Error) (Table \ref{tab_ts_mae}) and MSE (Mean Squared Error) (Table \ref{tab_ts_mse}), both of which are widely used metrics in time series forecasting research.

\subsubsection{Baselines}
\label{app_ts_baselines}
We select 5 commonly used time series prediction baselines—Transformer \cite{attention}, Pyraformer \cite{pyraformer}, Non-stationary Transformer \cite{nonstransformer}, MICN \cite{MICN}, and TimesNet \cite{timesnet}—and replace their original hand-crafted time encodings with our proposed \ours~to demonstrate that \ours~can be effectively applied to time series prediction models and improve the performance of downstream tasks. A brief introduction to these baseline models is provided below:

\begin{itemize}
    \item \textbf{Transformer} \cite{attention} leverages the self-attention mechanism to model long-range dependencies in sequences, making it a powerful tool for time series forecasting, especially in cases with complex time patterns. Its global context modeling capability enables it to capture intricate relationships between time steps effectively.

    \item \textbf{Pyraformer} \cite{pyraformer} introduces a pyramid attention mechanism that hierarchically reduces the computational burden while preserving the ability to model both local and global dependencies. This design makes it particularly well-suited for handling long time series with improved efficiency and scalability.

    \item \textbf{Non-stationary Transformer} \cite{nonstransformer}: addresses challenges in forecasting non-stationary time series by incorporating dynamic feature adjustments and context-aware attention mechanisms. This allows the model to adapt to evolving data distributions, ensuring robust and accurate predictions in dynamic environments.

    \item \textbf{MICN} \cite{MICN} integrates multi-scale architectures to capture both short-term patterns and long-term dependencies in time series data. By combining local convolutional operations and global attention, it provides a balanced approach to handling diverse temporal characteristics.

    \item \textbf{TimesNet} \cite{timesnet} innovatively models time series data in the frequency domain, leveraging discrete Fourier transformations to capture periodicity and trends. This approach enhances its ability to predict time series with prominent seasonal and cyclical behaviors efficiently. 
\end{itemize}

\subsubsection{Datasets}
\label{app_ts_datasets}
We utilize 4 real-world datasets to evaluate the effectiveness of our method on time series prediction tasks, encompassing various real-world scenarios. The dataset statistics are presented in Table \ref{tab_ts_data}, with detailed descriptions provided below.

\begin{table}[h]
\centering
\caption{Time Series Dataset Statistics: The dataset size is organized in (Train, Validation, Test). Please refer to \cite{timesnet} for the original table.}
\small
\begin{tabular}{c|c|c|c|c}
\toprule
Dataset& Dim & Series Length  & Dataset Size          & Information (Frequency) \\ \midrule
ETTm1, ETTm2           & 7            & \{96, 192, 336, 720\}    & (34465, 11521, 11521)          & Electricity (15 mins)            \\ \midrule
ETTh1, ETTh2           & 7            & \{96, 192, 336, 720\}    & (8545, 2881, 2881)             & Electricity (15 mins)            \\ \midrule
Electricity            & 321          & \{96, 192, 336, 720\}    & (18317, 2633, 5261)            & Electricity (Hourly)             \\ \midrule
Exchange               & 8            & \{96, 192, 336, 720\}    & (5120, 665, 1422)              & Exchange rate (Daily)            \\ \midrule
Weather                & 21           & \{96, 192, 336, 720\}    & (36792, 5271, 10540)           & Weather (10 mins)                \\ \bottomrule

\end{tabular}

\label{tab_ts_data}
\end{table}

\begin{itemize}
    \item \textbf{ETT}\footnote{\url{https://github.com/zhouhaoyi/ETDataset}} dataset includes time series data for oil temperature and power load measurements from electricity transformers, collected between July 2016 and July 2018. Specifically, the subsets ETTm1 and ETTm2 are sampled at 15-minute intervals, while ETTh1 and ETTh2 are recorded hourly.

    \item \textbf{Electricity}\footnote{\url{https://archive.ics.uci.edu/ml/datasets/ElectricityLoadDiagrams20112014}} dataset provides hourly electricity consumption data for 321 clients, spanning the period from 2012 to 2014.

    \item \textbf{Exchange} \cite{exchange} dataset offers daily panel data on exchange rates from eight countries, covering the years 1990 to 2016. 

    \item \textbf{Weather}\footnote{\url{https://www.bgc-jena.mpg.de/wetter/}} dataset contains meteorological time series data, comprising 21 weather indicators recorded at 10-minute intervals in 2020 by the Weather Station of the Max Planck Biogeochemistry Institute.
\end{itemize}

\subsection{Dynamic Graph Baselines and Datasets}
\label{app_dyg_exps}

\subsubsection{Experiment Implementation Details}
The hyper-parameters are based on the best configurations reported in the papers, and we keep them unchanged across different experiments for each baseline model to ensure a fair comparison. We rerun the baseline models TGAT with batch size 100 and reuse the baseline results reported in the DyGFormer paper for other baselines. The results are evaluated using Average Precision, i.e., AP (Table \ref{tab_tg_ap}) and Area Under the Receiver Operating Characteristic Curve, i.e., AUC-ROC (Table \ref{tab_tg_auc}), both of which are widely used metrics in dynamic graph representation learning research.

\subsubsection{Baselines}
\label{app_dyg_baselines}
We select 4 commonly used continuous dynamic graph representation learning baselines—TGAT \cite{tgat}, TGN \cite{tgn}, TCL \cite{tcl}, and DyGFormer \cite{dygformer}—and replace the Functional Time Encoding methods \cite{time2vec, xu2019time} with \ours~to demonstrate its optimal performance on the dynamic graph link prediction task. A brief introduction to these baseline models is provided below:

\begin{itemize}
    \item \textbf{TGAT} \cite{tgat} introduces a temporal attention mechanism to aggregate information from temporal-topological neighbors, thereby generating temporal node representations in temporal graphs. Additionally, it proposes a trainable time encoding function to capture distinguishable temporal information, which has been widely adopted in subsequent dynamic graph network architectures.

    \item \textbf{TGN} \cite{tgn} integrates key ideas from previous models and introduces a memory module that maintains a state vector for each node. The memory is updated dynamically whenever nodes participate in interactions. Additionally, TGN incorporates a message-passing module, a memory update module, and a temporal embedding module to generate effective temporal representations for nodes within temporal graphs.

    \item \textbf{TCL} \cite{tcl} utilizes a breadth-first search algorithm to construct a temporal dependency interaction sub-graph, extracting interaction sequences. It employs a Transformer encoder that integrates both topological and temporal information to learn representations of central nodes. Additionally, TCL introduces a cross-attention mechanism within the Transformer to model the inter-dependencies between interacting node pairs.

    \item \textbf{DyGFormer} \cite{dygformer} leverages 1-hop neighbor information for learning temporal graph representations. It employs a Transformer encoder enhanced with a patching technique to effectively capture long-term dependencies among nodes in temporal graphs. Furthermore, DyGFormer incorporates a Neighbor Co-occurrence Feature to preserve the correlation information between source and target nodes.

\end{itemize}

\subsubsection{Datasets}
\label{app_dyg_datasets}
We utilize 4 real-world datasets \cite{jodie} to evaluate the effectiveness of our method on dynamic graph representation learning tasks, encompassing various real-world scenarios. The dataset statistics are presented in Table \ref{tab_tg_data}, with detailed descriptions provided below.


\begin{table}[!h]
\centering
\caption{Dynamic Graph Dataset Statistics: $\text{Dim}{n}$ and $\text{Dim}{e}$ represent the dimensions of node features and edge features, respectively. For non-attributed graphs, we follow previous studies \cite{tgat, tgn} and use 172-dimensional zero vectors as padding.}

\small
\begin{tabular}{c|cc|cc|c|c|c}
\toprule
Dataset   & $\text{Dim}_{n}$ & $\text{Dim}_{e}$ & \# Nodes & \# Edges & Information & Duration  &  Time Granularity \\ \midrule
Wikipedia   & -    & 172  & 9,227    & 157,474  &  Social  & 1 month   &  Unix timestamps  \\ \midrule
Reddit       & -     & 172   & 10,984   & 672,447  &  Social  & 1 month   &  Unix timestamps   \\ \midrule
MOOC        & -     & 4  & 7,144    & 411,749  &  Interaction & 17 months &  Unix timestamps     \\ \midrule
LastFM     & -     & -  & 1,980    & 1,293,103 &  Interaction  & 1 month   &  Unix timestamps   \\ \bottomrule
\end{tabular}
\label{tab_tg_data}
\end{table}


\begin{itemize}
    \item \textbf{Wikipedia} records editing activities on Wikipedia pages over a one-month timeframe. Nodes in this graph represent users or pages, and temporal links with timestamps capture the edits. Each link is associated with a 172-dimensional feature vector based on LIWC (Linguistic Inquiry and Word Count) \cite{pennebaker2001linguistic}.

    \item \textbf{Reddit} captures user activity across subreddits over a one-month period. In this dataset, nodes represent users or subreddits, while timestamped links denote posting actions. Each link is further characterized by a 172-dimensional feature vector derived from LIWC.

    \item \textbf{MOOC} captures the interactions of users on a widely used MOOC platform, structured as a directed, temporal network. In this representation, nodes correspond to users and course activities (referred to as targets), while edges denote the actions performed by users on these targets

    \item \textbf{LastFM} records interaction data where users listen to songs over a month. In LastFM, nodes correspond to users and songs, and the links represent listening activities performed by users.

\end{itemize}

\subsection{Real-World Application Dataset}
\label{app_application_dataset}
The dataset used for real-world application experiments is a financial risk control dataset, containing records of 483,379 users’ transaction behavior at various merchants over a 60-day period. It includes a total of 26,850,000 transactions. Each user is represented by a 585-dimensional feature, each merchant by a 128-dimensional feature, and each transaction by a 202-dimensional feature, with all transactions labeled with UNIX timestamps. The ratio of positive users (with default risk) to negative users (without default risk) is 1:9.92 in the training dataset and 1:20.25 in the test dataset. The backbone model employs a specially designed Transformer-based architecture to aggregate users’ historical transaction features, merchant features, user features, and an optional time embedding into user embeddings, which are then used to predict whether the user has default risk.

\section{Additional Experimental Results}
\label{app_additional_exp_results}
Here, we provide the complete results of the experiments discussed in the main text.

\subsection{Results of multivariate time series long-term forecasting task evaluated using MSE}
The results of the multivariate time series long-term forecasting task, evaluated using MSE, are presented in Table \ref{tab_ts_mse}. These results are organized in the same manner as those in Table \ref{tab_ts_mae}.

\begin{table*}[h]

\caption{Time series prediction: multivariate long-term forecasting task. The past sequence length is set to 96, while the prediction lengths are \{96, 192, 336, 720\}. The results are reported in terms of MSE, where lower values indicate better performance. HCTE (Hand-Crafted Time Encoding) is a method widely adopted in time series research. FTE stands for Functional Time Encoding. The win rate represents the percentage of cases where \ours~outperforms the HCTE. The best results for each baseline, dataset and prediction length combinations are in \textbf{bold}. 
ETT consists of 4 subsets. Here, we present the average results across these subsets, with the full results provided in Table \ref{tab_ett_mse}.}

\setlength{\tabcolsep}{2mm}{
\scriptsize
\begin{tabular}{cc|ccc|ccc|ccc|ccc|ccc|c}
\toprule
\multicolumn{2}{c|}{MSE}                                                           & \multicolumn{3}{c|}{Transformer} & \multicolumn{3}{c|}{Pyraformer}         & \multicolumn{3}{c|}{NS Trans.}          & \multicolumn{3}{c|}{MICN}                                                             & \multicolumn{3}{c|}{TimesNet}                                  & Win                    \\ \cmidrule{1-17}
\multicolumn{2}{c|}{TE}                                                            & HCTE  & FTE    & \ours           & HCTE          & FTE   & \ours           & HCTE          & FTE   & \ours           & HCTE                                 & FTE   & \ours                                  & HCTE          & FTE   & \ours                                  & Rate                   \\ \midrule
\multicolumn{1}{c|}{}                                                        & 96  & 1.219  & 1.204  & \textbf{0.568} & 0.794          & 0.960 & \textbf{0.650} & 0.392          & 0.486 & \textbf{0.347} & 0.392                                 & 0.310 & \textbf{0.296} & 0.312          & 0.322 & \textbf{0.305}                        &                        \\
\multicolumn{1}{c|}{}                                                        & 192 & 2.601  & 1.643  & \textbf{0.959} & 1.667          & 1.812 & \textbf{1.058} & 0.446          & 0.548 & \textbf{0.410} & 0.446                                 & 0.396 &  \textbf{0.357} & \textbf{0.365} & 0.390 & 0.368                                 &                        \\
\multicolumn{1}{c|}{}                                                        & 336 & 2.438  & 1.557  & \textbf{1.192} & 1.981          & 1.962 & \textbf{1.297} & 0.492          & 0.651 & \textbf{0.467} & 0.492                                 & 0.487 & \textbf{0.430}                        & 0.419          & 0.422 & \textbf{0.404} &                        \\
\multicolumn{1}{c|}{\multirow{-4}{*}{\rotatebox[origin=c]{90}{ETT}}}         & 720 & 1.998  & 2.217  & \textbf{1.313} & 2.589          & 2.442 & \textbf{1.612} & 0.552          & 0.660 & \textbf{0.531} & 0.552                                 & 0.630 & \textbf{0.517}                        & 0.467          & 0.468 &\textbf{0.441} & \multirow{-4}{*}{95\%} \\ \midrule
\multicolumn{1}{c|}{}                                                        & 96  & 0.258  & 0.280  & \textbf{0.252} & 0.285          & 0.281 & \textbf{0.267} & 0.169          & 0.171 & \textbf{0.163} & 0.164                                 & 0.156 &  \textbf{0.150} & 0.168          & 0.168 & \textbf{0.164}                        &                        \\
\multicolumn{1}{c|}{}                                                        & 192 & 0.266  & 0.310  & \textbf{0.259} & 0.298          & 0.288 & \textbf{0.274} & 0.182          & 0.192 & \textbf{0.178} & 0.177                                 & 0.170 & \textbf{0.166} & 0.184          & 0.179 & \textbf{0.179}                        &                        \\
\multicolumn{1}{c|}{}                                                        & 336 & 0.275  & 0.339  & \textbf{0.265} & 0.307          & 0.306 & \textbf{0.274} & 0.200          & 0.201 & \textbf{0.190} & 0.193                                 & 0.189 & \textbf{0.184} & 0.198          & 0.210 & \textbf{0.193}                        &                        \\
\multicolumn{1}{c|}{\multirow{-4}{*}{\rotatebox[origin=c]{90}{Electricity}}} & 720 & 0.288  & 0.395  & \textbf{0.276} & 0.304          & 0.304 & \textbf{0.293} & \textbf{0.222} & 0.237 & 0.226          & 0.212                                 & 0.228 & \textbf{0.208} & 0.220          & 0.287 & \textbf{0.220}                        & \multirow{-4}{*}{95\%} \\ \midrule
\multicolumn{1}{c|}{}                                                        & 96  & 0.545  & 0.788  & \textbf{0.464} & \textbf{0.505} & 0.616 & 0.584          & 0.111          & 0.137 & \textbf{0.102} & 0.102                                 & 0.098 &  \textbf{0.080} & 0.107          & 0.108 & \textbf{0.103}                        &                        \\
\multicolumn{1}{c|}{}                                                        & 192 & 0.950  & 1.028  & \textbf{0.903} & 1.015          & 0.983 & \textbf{0.904} & 0.219          & 0.273 & \textbf{0.193} & 0.172                                 & 0.186 & \textbf{0.152} & 0.226          & 0.217 & \textbf{0.211}                        &                        \\
\multicolumn{1}{c|}{}                                                        & 336 & 1.462  & 1.866  & \textbf{1.114} & 1.263          & 1.420 & \textbf{1.056} & 0.421          & 0.463 & \textbf{0.342} & 0.272                                 & 0.356 & \textbf{0.265} & 0.367          & 0.413 & \textbf{0.367}                        &                        \\
\multicolumn{1}{c|}{\multirow{-4}{*}{\rotatebox[origin=c]{90}{Exchange}}}    & 720 & 2.569  & 2.002  & \textbf{1.736} & 1.762          & 1.714 & \textbf{1.316} & 1.092          & 1.546 & \textbf{0.682} & 0.714                                 & 0.833 &  \textbf{0.636} & 0.964          & 0.989 & \textbf{0.958}                        & \multirow{-4}{*}{95\%} \\ \midrule
\multicolumn{1}{c|}{}                                                        & 96  & 0.393  & 0.184  & \textbf{0.172} & 0.225          & 0.207 & \textbf{0.181} & 0.173          & 0.172 & \textbf{0.168} &  \textbf{0.161} & 0.198 & 0.166                                 & 0.172          & 0.172 & \textbf{0.166}                        &                        \\
\multicolumn{1}{c|}{}                                                        & 192 & 0.547  & 0.249  & \textbf{0.220} & 0.252          & 0.238 & \textbf{0.230} & 0.245          & 0.224 & \textbf{0.215} & 0.220                                 & 0.243 &  \textbf{0.209} & 0.219          & 0.221 & \textbf{0.212}                        &                        \\
\multicolumn{1}{c|}{}                                                        & 336 & 0.678  & 0.328  & \textbf{0.309} & 0.362          & 0.324 & \textbf{0.284} & 0.321          & 0.295 & \textbf{0.277} & 0.278                                 & 0.285 & \textbf{0.251} & 0.280          & 0.278 & \textbf{0.274}                        &                        \\
\multicolumn{1}{c|}{\multirow{-4}{*}{\rotatebox[origin=c]{90}{Weather}}}     & 720 & 0.844  & 0.480  & \textbf{0.425} & 0.411          & 0.394 & \textbf{0.383} & 0.414          & 0.350 & \textbf{0.338} & 0.311                                 & 0.350 &  \textbf{0.303} & 0.365          & 0.354 & \textbf{0.352}                        & \multirow{-4}{*}{95\%} \\ \midrule
\multicolumn{2}{c|}{Win Rate}                                                      & \multicolumn{3}{c|}{100\%}       & \multicolumn{3}{c|}{94\%}               & \multicolumn{3}{c|}{94\%}               & \multicolumn{3}{c|}{94\%}                                                             & \multicolumn{3}{c|}{94\%}                                      & 95\%                   \\ \bottomrule
\end{tabular}

}

\label{tab_ts_mse}

\end{table*}

\subsection{Results of dynamic graph link prediction task evaluated using AUC-ROC}
The results of the dynamic graph link prediction task, evaluated using AUC-ROC, are presented in Table \ref{tab_tg_auc}. 

\begin{table*}[h]

\caption{Dynamic graph link prediction task: The results are reported in AUC-ROC, where higher values indicate better performance. The better results are in \textbf{bold}. Here, we present the top-performing results across variations of \ours, with the full results provided in Table \ref{tab_tg_types_auc}. FTE represents Functional Time Encoding which is commonly used in dynamic graph research.}

\setlength{\tabcolsep}{2.8mm}{
\scriptsize
\begin{tabular}{c|c|cc|cc|cc|cc}
\toprule
                                                                       & AUC        & \multicolumn{2}{c|}{Wikipedia}            & \multicolumn{2}{c|}{Reddit}               & \multicolumn{2}{c|}{MOOC}                 & \multicolumn{2}{c}{LastFM}                \\ \cmidrule{2-10} 
                                                                       & TE         & Transductive        & Inductive           & Transductive        & Inductive           & Transductive        & Inductive           & Transductive        & Inductive           \\ \midrule
\multirow{2}{*}{TGAT}                                                  & FTE & 96.69 ± 0.26          & 95.95 ± 0.33          & 98.48 ± 0.04          & 96.90 ± 0.07          & 86.44 ± 0.24          & 86.04 ± 0.19          & 70.89 ± 0.10          & 76.11 ± 0.11          \\
                                                                       & \ours & \textbf{97.63 ± 0.11} & \textbf{97.07 ± 0.10} & \textbf{98.51 ± 0.01} & \textbf{96.96 ± 0.05} & \textbf{89.46 ± 0.08} & \textbf{89.50 ± 0.05} & \textbf{74.24 ± 0.28} & \textbf{79.63 ± 0.14} \\ \midrule
\multirow{2}{*}{TGN}                                                   & FTE & 98.37 ± 0.07          & 97.72 ± 0.03          & 98.60 ± 0.06          & 97.39 ± 0.07          & 91.21 ± 1.15          & 91.24 ± 0.99          & 78.47 ± 2.94          & 82.61 ± 3.15          \\
                                                                       & \ours & \textbf{98.73 ± 0.07} & \textbf{98.11 ± 0.10} & \textbf{98.72 ± 0.00} & \textbf{97.55 ± 0.05} & \textbf{92.68 ± 0.37} & \textbf{92.41 ± 0.73} & \textbf{83.94 ± 1.85} & \textbf{87.74 ± 1.75} \\ \midrule
\multirow{2}{*}{TCL}                                                   & FTE & 95.84 ± 0.18          & 95.57 ± 0.20          & 97.42 ± 0.02          & 93.80 ± 0.07          & 83.12 ± 0.18          & 81.43 ± 0.19          & 64.06 ± 1.16          & 70.84 ± 0.85          \\
                                                                       & \ours & \textbf{97.84 ± 0.06} & \textbf{97.56 ± 0.02} & \textbf{97.68 ± 0.03} & \textbf{94.63 ± 0.05} & \textbf{84.73 ± 0.13} & \textbf{83.25 ± 0.23} & \textbf{70.17 ± 0.47} & \textbf{75.86 ± 0.44} \\ \midrule
\multirow{2}{*}{\begin{tabular}[c]{@{}c@{}}DyG-\\ Former\end{tabular}} & FTE & 98.91 ± 0.02          & 98.48 ± 0.03          & 99.15 ± 0.01          & 98.71 ± 0.01          & 87.91 ± 0.58          & 87.62 ± 0.51          & 93.05 ± 0.10          & 94.08 ± 0.08          \\
                                                                       & \ours & \textbf{99.04 ± 0.01} & \textbf{98.67 ± 0.02} & \textbf{99.17 ± 0.00} & \textbf{98.74 ± 0.01} & \textbf{89.18 ± 0.21} & \textbf{89.16 ± 0.30} & \textbf{93.65 ± 0.07} & \textbf{94.52 ± 0.08} \\ \bottomrule
\end{tabular}
}

\label{tab_tg_auc}
\end{table*}

\subsection{Full results of the multivariate long-term forecasting task on 4 ETT subsets}
We present the full results of the multivariate long-term forecasting task on the 4 ETT subsets in Tables \ref{tab_ett_mae} and \ref{tab_ett_mse}, as Tables \ref{tab_ts_mae} and \ref{tab_ts_mse} report the average results.


\begin{table*}[h]

\caption{Time series prediction: multivariate long-term forecasting task on 4 subsets of ETT. The past sequence length is set to 96, while the prediction lengths are \{96, 192, 336, 720\}. The results are reported in terms of MAE. The best results for each baseline, dataset and prediction length combinations are in \textbf{bold}.}

\centering
\setlength{\tabcolsep}{2.3mm}{
\scriptsize
\begin{tabular}{cc|ccc|ccc|ccc|ccc|ccc}
\toprule
\multicolumn{2}{c|}{MAE}                                                    & \multicolumn{3}{c|}{Transformer} & \multicolumn{3}{c|}{Pyraformer}         & \multicolumn{3}{c|}{NS Trans.}          & \multicolumn{3}{c|}{MINC}                        & \multicolumn{3}{c}{TimesNet}            \\ \midrule
\multicolumn{2}{c|}{TE}                                                     & HCTE  & FTE    & \ours           & HCTE & FTE            & \ours           & HCTE          & FTE   & \ours           & HCTE          & FTE            & \ours           & HCTE          & FTE   & \ours           \\ \midrule
\multicolumn{1}{c|}{\multirow{4}{*}{\rotatebox[origin=c]{90}{ETTm1}}} & 96  & 0.621  & 0.601  & \textbf{0.507} & 0.581 & 0.522          & \textbf{0.521} & 0.398          & 0.409 & \textbf{0.389} & 0.398          & 0.376          & \textbf{0.372} & \textbf{0.375} & 0.389 & 0.378          \\
\multicolumn{1}{c|}{}                                                 & 192 & 0.703  & 0.550  & \textbf{0.526} & 0.577 & 0.547          & \textbf{0.531} & 0.444          & 0.427 & \textbf{0.417} & 0.444          & \textbf{0.393} & 0.399          & \textbf{0.387} & 0.418 & 0.404          \\
\multicolumn{1}{c|}{}                                                 & 336 & 0.795  & 0.755  & \textbf{0.596} & 0.675 & 0.637          & \textbf{0.610} & 0.464          & 0.496 & \textbf{0.439} & 0.464          & 0.425          & \textbf{0.426} & \textbf{0.411} & 0.418 & 0.418          \\
\multicolumn{1}{c|}{}                                                 & 720 & 0.798  & 0.782  & \textbf{0.663} & 0.760 & 0.700          & \textbf{0.645} & 0.516          & 0.498 & \textbf{0.470} & 0.516          & 0.467          & \textbf{0.465} & \textbf{0.450} & 0.456 & 0.454          \\ \midrule
\multicolumn{1}{c|}{\multirow{4}{*}{\rotatebox[origin=c]{90}{ETTm2}}} & 96  & 0.506  & 0.451  & \textbf{0.420} & 0.458 & 0.756          & \textbf{0.411} & \textbf{0.274} & 0.304 & 0.278          & 0.274          & 0.287          & \textbf{0.271} & 0.267          & 0.268 & \textbf{0.262} \\
\multicolumn{1}{c|}{}                                                 & 192 & 0.908  & 0.700  & \textbf{0.558} & 0.649 & \textbf{0.611} & 0.625          & 0.339          & 0.379 & \textbf{0.325} & 0.339          & 0.349          & \textbf{0.321} & 0.309          & 0.306 & \textbf{0.302} \\
\multicolumn{1}{c|}{}                                                 & 336 & 0.796  & 0.806  & \textbf{0.753} & 0.811 & 0.902          & \textbf{0.775} & 0.361          & 0.400 & \textbf{0.353} & \textbf{0.361} & 0.439          & 0.364          & 0.351          & 0.351 & \textbf{0.342} \\
\multicolumn{1}{c|}{}                                                 & 720 & 1.192  & 1.301  & \textbf{0.851} & 1.416 & 1.491          & \textbf{1.137} & \textbf{0.413} & 0.446 & 0.420          & \textbf{0.413} & 0.506          & 0.432          & 0.403          & 0.409 & \textbf{0.399} \\ \midrule
\multicolumn{1}{c|}{\multirow{4}{*}{\rotatebox[origin=c]{90}{ETTh1}}} & 96  & 0.739  & 0.814  & \textbf{0.524} & 0.637 & 0.613          & \textbf{0.593} & 0.491          & 0.616 & \textbf{0.452} & 0.491          & 0.413          & \textbf{0.409} & \textbf{0.402} & 0.425 & 0.408          \\
\multicolumn{1}{c|}{}                                                 & 192 & 0.762  & 0.815  & \textbf{0.632} & 0.738 & 0.778          & \textbf{0.647} & 0.504          & 0.631 & \textbf{0.483} & 0.504          & 0.465          & \textbf{0.452} & 0.429          & 0.457 & \textbf{0.439} \\
\multicolumn{1}{c|}{}                                                 & 336 & 0.772  & 0.786  & \textbf{0.679} & 0.794 & 0.758          & \textbf{0.736} & \textbf{0.535} & 0.730 & 0.541          & 0.535          & 0.511          & \textbf{0.502} & 0.469          & 0.469 & \textbf{0.453} \\
\multicolumn{1}{c|}{}                                                 & 720 & 0.800  & 0.878  & \textbf{0.724} & 0.776 & 0.804          & \textbf{0.782} & 0.616          & 0.792 & \textbf{0.610} & 0.616          & 0.598          & \textbf{0.565} & 0.500          & 0.490 & \textbf{0.421} \\ \midrule
\multicolumn{1}{c|}{\multirow{4}{*}{\rotatebox[origin=c]{90}{ETTh2}}} & 96  & 1.323  & 1.349  & \textbf{0.752} & 0.892 & 0.989          & \textbf{0.806} & 0.458          & 0.413 & \textbf{0.390} & 0.458          & 0.392          & \textbf{0.349} & 0.374          & 0.367 & \textbf{0.359} \\
\multicolumn{1}{c|}{}                                                 & 192 & 2.184  & 1.599  & \textbf{1.133} & 1.632 & 1.760          & \textbf{1.150} & 0.493          & 0.475 & \textbf{0.434} & 0.493          & 0.485          & \textbf{0.408} & 0.414          & 0.418 & \textbf{0.406} \\
\multicolumn{1}{c|}{}                                                 & 336 & 2.113  & 1.405  & \textbf{1.256} & 1.893 & 1.856          & \textbf{1.330} & 0.551          & 0.528 & \textbf{0.463} & 0.551          & 0.569          & \textbf{0.500} & 0.452          & 0.458 & \textbf{0.438} \\
\multicolumn{1}{c|}{}                                                 & 720 & 1.488  & 1.623  & \textbf{1.276} & 1.832 & 1.761          & \textbf{1.274} & 0.560          & 0.492 & \textbf{0.460} & 0.560          & 0.673          & \textbf{0.558} & 0.468          & 0.466 & \textbf{0.443} \\ \bottomrule
\end{tabular}

}

\label{tab_ett_mae}

\end{table*}


\begin{table*}[h]

\caption{Time series prediction: multivariate long-term forecasting task on 4 subsets of ETT. The past sequence length is set to 96, while the prediction lengths are \{96, 192, 336, 720\}. The results are reported in terms of MSE. The best results for each baseline, dataset and prediction length combinations are in \textbf{bold}.}

\centering
\setlength{\tabcolsep}{2.3mm}{
\scriptsize
\begin{tabular}{cc|ccc|ccc|ccc|ccc|ccc}
\toprule
\multicolumn{2}{c|}{MSE}                                                    & \multicolumn{3}{c|}{Transformer} & \multicolumn{3}{c|}{Pyraformer}         & \multicolumn{3}{c|}{NS Trans.}          & \multicolumn{3}{c|}{MINC}                        & \multicolumn{3}{c}{TimesNet}            \\ \midrule
\multicolumn{2}{c|}{TE}                                                     & HCTE  & FTE    & \ours           & HCTE & FTE            & \ours           & HCTE          & FTE   & \ours           & HCTE          & FTE            & \ours           & HCTE          & FTE   & \ours           \\ \midrule
\multicolumn{1}{c|}{\multirow{4}{*}{\rotatebox[origin=c]{90}{ETTm1}}} & 96  & 0.713  & 0.667  & \textbf{0.511} & 0.708 & \textbf{0.581} & 0.584          & 0.386          & 0.430 & \textbf{0.370} & 0.386          & 0.329          & \textbf{0.325} & \textbf{0.338} & 0.365 & 0.339          \\
\multicolumn{1}{c|}{}                                                 & 192 & 0.866  & 0.592  & \textbf{0.573} & 0.693 & 0.621          & \textbf{0.560} & 0.459          & 0.464 & \textbf{0.418} & 0.459          & \textbf{0.364} & 0.373          & \textbf{0.374} & 0.430 & 0.392          \\
\multicolumn{1}{c|}{}                                                 & 336 & 1.063  & 1.035  & \textbf{0.692} & 0.848 & 0.754          & \textbf{0.689} & 0.495          & 0.651 & \textbf{0.451} & 0.495          & 0.403          & \textbf{0.396} & \textbf{0.410} & 0.412 & 0.412          \\
\multicolumn{1}{c|}{}                                                 & 720 & 1.075  & 1.062  & \textbf{0.788} & 1.009 & 0.855          & \textbf{0.777} & 0.585          & 0.584 & \textbf{0.510} & 0.585          & 0.469          & \textbf{0.463} & 0.478          & 0.482 & \textbf{0.477} \\ \midrule
\multicolumn{1}{c|}{\multirow{4}{*}{\rotatebox[origin=c]{90}{ETTm2}}} & 96  & 0.486  & 0.354  & \textbf{0.327} & 0.384 & 1.047          & \textbf{0.310} & \textbf{0.192} & 0.246 & 0.195          & 0.192          & 0.190          & \textbf{0.187} & 0.187          & 0.191 & \textbf{0.182} \\
\multicolumn{1}{c|}{}                                                 & 192 & 1.499  & 0.834  & \textbf{0.522} & 0.730 & \textbf{0.653} & 0.672          & 0.280          & 0.378 & \textbf{0.266} & 0.280          & 0.271          & \textbf{0.241} & 0.249          & 0.253 & \textbf{0.248} \\
\multicolumn{1}{c|}{}                                                 & 336 & 1.107  & 1.113  & \textbf{0.949} & 1.144 & 1.377          & \textbf{1.023} & 0.334          & 0.387 & \textbf{0.315} & 0.334          & 0.398          & \textbf{0.307} & 0.321          & 0.327 & \textbf{0.310} \\
\multicolumn{1}{c|}{}                                                 & 720 & 2.609  & 2.931  & \textbf{1.104} & 3.624 & 3.708          & \textbf{2.195} & \textbf{0.417} & 0.491 & 0.431          & 0.417          & 0.525          & \textbf{0.414} & 0.408          & 0.427 & \textbf{0.408} \\ \midrule
\multicolumn{1}{c|}{\multirow{4}{*}{\rotatebox[origin=c]{90}{ETTh1}}} & 96  & 0.876  & 1.032  & \textbf{0.535} & 0.727 & 0.664          & \textbf{0.632} & 0.513          & 0.841 & \textbf{0.470} & 0.513          & 0.381          & \textbf{0.379} & \textbf{0.384} & 0.412 & 0.387          \\
\multicolumn{1}{c|}{}                                                 & 192 & 0.919  & 1.083  & \textbf{0.710} & 0.903 & 0.949          & \textbf{0.734} & 0.534          & 0.786 & \textbf{0.522} & 0.534          & 0.452          & \textbf{0.434} & 0.436          & 0.461 & \textbf{0.433} \\
\multicolumn{1}{c|}{}                                                 & 336 & 0.960  & 1.003  & \textbf{0.805} & 1.011 & 0.931          & \textbf{0.887} & \textbf{0.588} & 0.941 & 0.638          & 0.588          & 0.520          & \textbf{0.501} & 0.491          & 0.483 & \textbf{0.470} \\
\multicolumn{1}{c|}{}                                                 & 720 & 1.030  & 1.183  & \textbf{0.856} & 0.992 & 1.012          & \textbf{0.961} & \textbf{0.643} & 1.052 & 0.735          & 0.643          & 0.646          & \textbf{0.578} & 0.521          & 0.509 & \textbf{0.455} \\ \midrule
\multicolumn{1}{c|}{\multirow{4}{*}{\rotatebox[origin=c]{90}{ETTh2}}} & 96  & 2.802  & 2.762  & \textbf{0.897} & 1.357 & 1.548          & \textbf{1.074} & 0.476          & 0.427 & \textbf{0.355} & 0.476          & 0.339          & \textbf{0.294} & 0.340          & 0.319 & \textbf{0.313} \\
\multicolumn{1}{c|}{}                                                 & 192 & 7.123  & 4.066  & \textbf{2.030} & 4.342 & 5.025          & \textbf{2.268} & 0.512          & 0.563 & \textbf{0.436} & 0.512          & 0.495          & \textbf{0.380} & 0.402          & 0.416 & \textbf{0.398} \\
\multicolumn{1}{c|}{}                                                 & 336 & 6.621  & 3.078  & \textbf{2.322} & 4.922 & 4.786          & \textbf{2.589} & 0.552          & 0.625 & \textbf{0.466} & 0.552          & 0.625          & \textbf{0.516} & 0.452          & 0.464 & \textbf{0.422} \\
\multicolumn{1}{c|}{}                                                 & 720 & 3.279  & 3.695  & \textbf{2.502} & 4.733 & 4.191          & \textbf{2.516} & 0.562          & 0.513 & \textbf{0.448} & \textbf{0.562} & 0.880          & 0.615          & 0.462          & 0.456 & \textbf{0.426} \\ \bottomrule
\end{tabular}

}

\label{tab_ett_mse}
\end{table*}


\subsection{Dimensions of Time Embedding}
\begin{figure}[!h]
    \centering
    \begin{subfigure}[b]{0.33\linewidth}
        \includegraphics[width=1\linewidth]{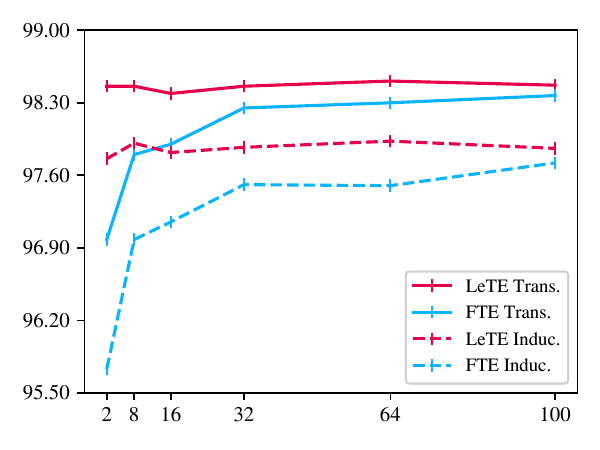}
        \caption{Wikipedia/TGN}
    \end{subfigure}
    \hspace{20pt}
    \begin{subfigure}[b]{0.33\linewidth}
        \includegraphics[width=1\linewidth]{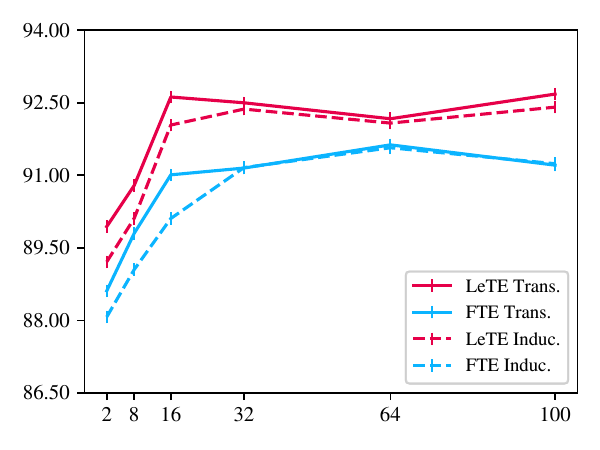}
        \caption{MOOC/TGN}
    \end{subfigure}

\caption{AUC-ROC results comparing different dimensions of the FTE and Spline-based \ours~on Wikipedia/TGN and MOOC/TGN.}
\label{fig:dim_auc_tgn}
\end{figure}

\begin{figure}[!h]
    \centering
    \begin{subfigure}[b]{0.33\linewidth}
        \includegraphics[width=1\linewidth]{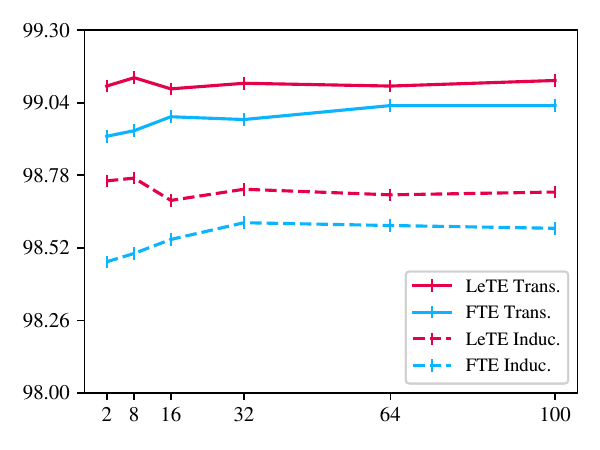}
        \caption{AP}
    \end{subfigure}
    \hspace{20pt}
    \begin{subfigure}[b]{0.33\linewidth}
        \includegraphics[width=1\linewidth]{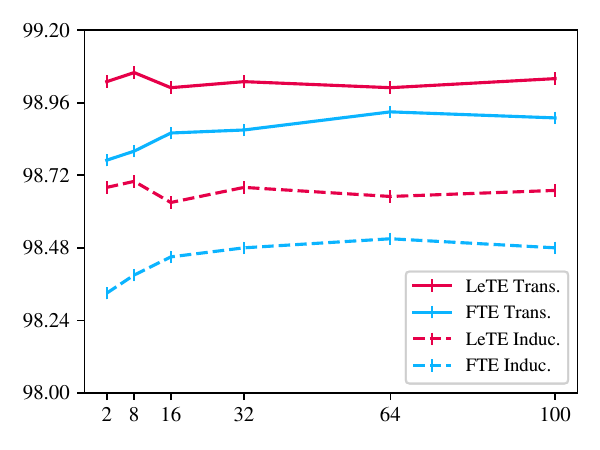}
        \caption{AUC-ROC}
    \end{subfigure}

\caption{AP and AUC-ROC results comparing different dimensions of the FTE and Spline-based \ours~on Wikipedia/DyGFormer.}
\label{fig:dim_dyg}
\end{figure}

We present the AUC-ROC results for Wikipedia/TGN and MOOC/TGN with different time embedding dimensions (for both FTE and \ours) in Figure \ref{fig:dim_auc_tgn}. To cover a broader range of scenarios, we also include the results for DyGFormer on the Wikipedia dataset in Figure \ref{fig:dim_dyg}.


\section{More Experiments}

\subsection{Statistic Analysis of the Complex Time Patterns in Data}
\label{app_periodic_and_non}
Time-related data often contains mixed and complex patterns, which can primarily be categorized as periodic and non-periodic.
To investigate the periodic and non-periodic patterns in the data, we analyze four dynamic graph datasets using spectral entropy \cite{entropy}. First, we normalize the time or time differences (since previous dynamic graph representation learning methods typically use time differences as inputs to the time encoding, we include this analysis here as well) for each node with more than five interactions, mapping the values to the range $[0,1]$. We then treat each node’s interaction times as a signal sequence. The spectral entropy for each node is computed as follows:
We begin by applying the Fast Fourier Transform (FFT) to the normalized signal sequences: $X(f) = \text{FFT}(
t_{\text{norm}})$, where $X(f)$ is the frequency-domain representation of the signal. Next, we calculate the magnitude of the frequency components $M(f) = |X(f)|$. The magnitudes are then normalized to form a probability distribution: $P(f) = \frac{M(f)}{\sum_{f} M(f)}$. Finally, the spectral entropy is computed as: $H(P) = - \sum_{f} P(f) \log P(f)$, which measures the uniformity of the frequency components. A lower entropy value indicates periodicity, while a higher entropy value suggests randomness.

\begin{figure}[!h]
    \centering
    \begin{subfigure}[b]{0.45\linewidth}
        \includegraphics[width=1\linewidth]{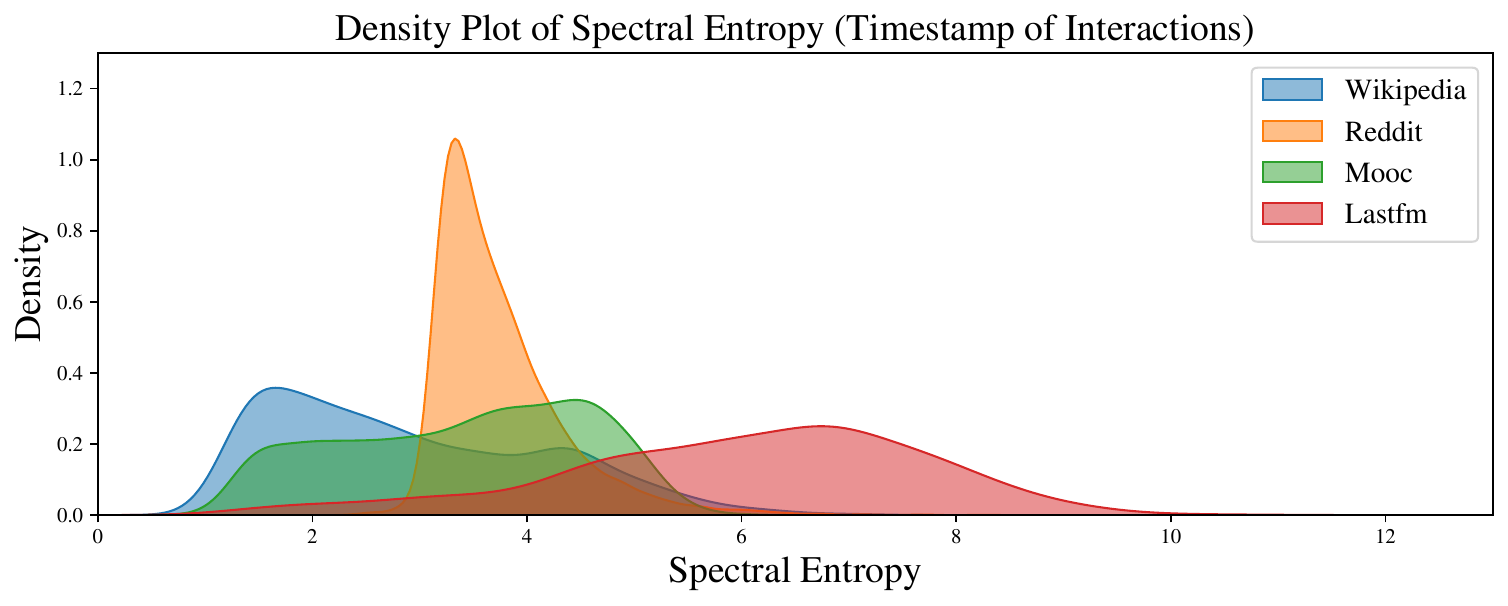}
    \end{subfigure}
    \hspace{20pt}
    \begin{subfigure}[b]{0.45\linewidth}
        \includegraphics[width=1\linewidth]{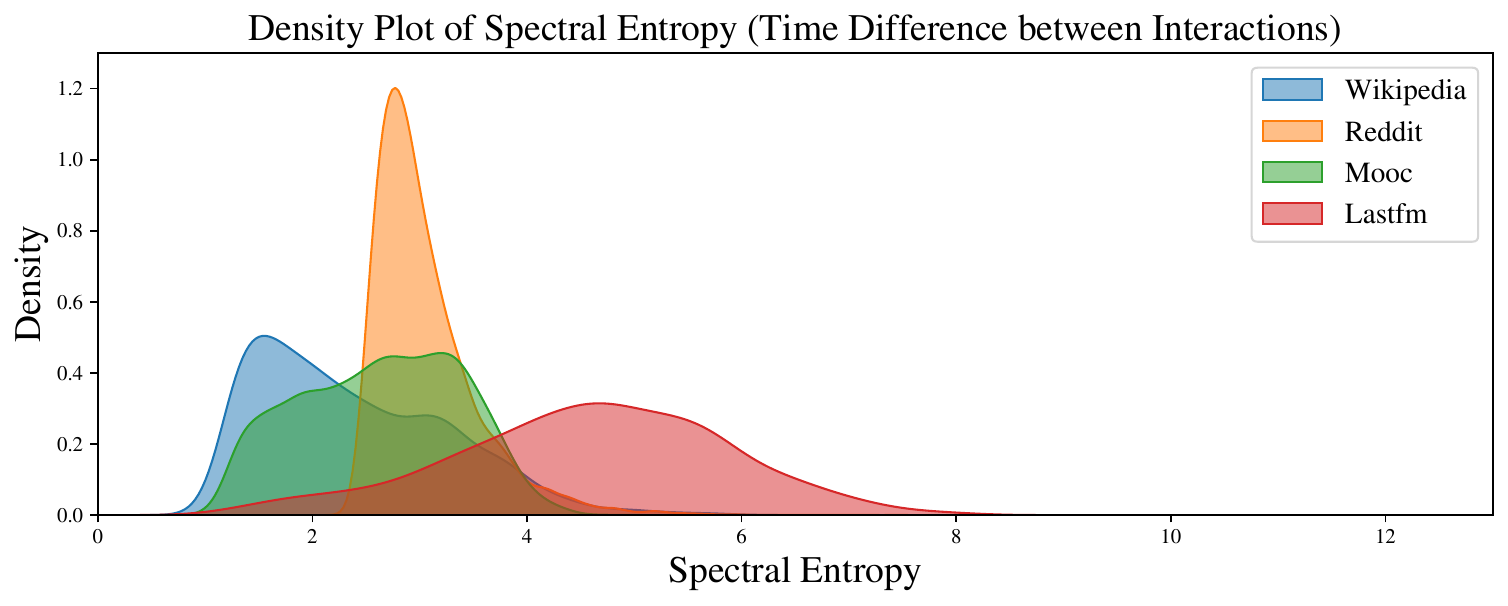}
    \end{subfigure}

\caption{Density plots of spectral entropy for dynamic graph datasets.}
\label{fig:entropy}
\end{figure}

We present the density plots of the spectral entropy in Figure \ref{fig:entropy}. As shown in the figures, only a small portion of the nodes exhibit strong periodicity in their interaction times or time differences, while most nodes show high entropy, indicating non-periodic behavior. This suggests that capturing periodic patterns alone is insufficient; it is also important to model non-periodic patterns to enhance the efficiency and expressiveness of the time encoding.

\subsection{Comparative Analysis of Different Variants of \ours}
\label{app_exp_t2v_types}

We conducted a set of additional experiments on dynamic graph link prediction tasks, applying different variants of \ours~and comparing their downstream task performance, evaluated by AP and AUC-ROC. The results, presented in Tables \ref{tab_tg_types_ap} and \ref{tab_tg_types_auc}, indicate that, in most cases, Combined \ours~achieves the best performance among the three variants of \ours. This outcome is intuitive, as Combined \ours~leverages the strengths of both Fourier-based \ours~and Spline-based \ours, enabling it to effectively model diverse time patterns.

Due to differences in the periodicity and non-periodicity of node interactions across datasets, the effectiveness of Fourier-based \ours~and Spline-based \ours~varies. Nonetheless, in most cases, both methods outperform the benchmark. This demonstrates that even when using only Fourier-based \ours~or Spline-based \ours, they can effectively model different patterns, including periodic, non-periodic and mixed patterns, in the data.

\begin{table*}[h]

\caption{Comparing Functional Time Encoding (FTE), Fourier-based \ours~(F-\ours), Spline-based \ours~(S-\ours) and Combined \ours~(C-\ours): Dynamic graph link prediction results in AP. The best results are in \textbf{bold}.}

\centering
\setlength{\tabcolsep}{2.7mm}{
\scriptsize
\begin{tabular}{c|c|cc|cc|cc|cc}
\toprule
                                                                       & AP         & \multicolumn{2}{c|}{Wikipedia}            & \multicolumn{2}{c|}{Reddit}               & \multicolumn{2}{c|}{MOOC}                 & \multicolumn{2}{c}{LastFM}                \\ \cmidrule{2-10} 
                                                                       & TE         & Transductive        & Inductive           & Transductive        & Inductive           & Transductive        & Inductive           & Transductive        & Inductive           \\ \midrule
\multirow{4}{*}{TGAT}                                                  & FTE & 96.95 ± 0.24          & 96.33 ± 0.26          & 98.53 ± 0.04          & 97.01 ± 0.05          & 85.34 ± 0.19          & 84.94 ± 0.04          & 72.73 ± 0.11          & 77.78 ± 0.13          \\
                                                                       & F-\ours     & 96.82 ± 0.16          & 96.31 ± 0.13          & 98.54 ± 0.03          & 97.03 ± 0.02          & 85.25 ± 0.29          & 85.08 ± 0.29          & 72.31 ± 0.30          & 77.19 ± 0.38          \\
                                                                       & S-\ours     & 97.54 ± 0.06          & 97.06 ± 0.05          & 98.56 ± 0.01          & \textbf{97.05 ± 0.06} & \textbf{88.31 ± 0.10} & 88.13 ± 0.28          & 75.68 ± 0.55          & 80.61 ± 0.42          \\
                                                                       & C-\ours     & \textbf{97.82 ± 0.09} & \textbf{97.34 ± 0.08} & \textbf{98.56 ± 0.01} & 96.99 ± 0.06          & 88.30 ± 0.05          & \textbf{88.37 ± 0.12} & \textbf{76.22 ± 0.25} & \textbf{81.32 ± 0.14} \\ \midrule
\multirow{4}{*}{TGN}                                                   & FTE & 98.45 ± 0.06          & 97.83 ± 0.04          & 98.63 ± 0.06          & 97.50 ± 0.07          & 89.15 ± 1.60          & 89.04 ± 1.17          & 77.07 ± 3.97          & 81.45 ± 4.29          \\
                                                                       & F-\ours     & 98.57 ± 0.09          & 97.94 ± 0.08          & 98.66 ± 0.01          & 97.39 ± 0.07          & 90.04 ± 0.67          & 89.94 ± 0.49          & 77.58 ± 5.22          & 82.82 ± 6.53          \\
                                                                       & S-\ours     & 98.55 ± 0.06          & 97.98 ± 0.08          & \textbf{98.74 ± 0.00} & \textbf{97.65 ± 0.04} & 91.09 ± 0.20          & \textbf{90.87 ± 0.83} & 82.26 ± 2.27          & 86.46 ± 0.77          \\
                                                                       & C-\ours     & \textbf{98.78 ± 0.07} & \textbf{98.19 ± 0.09} & 98.74 ± 0.01          & 97.52 ± 0.12          & \textbf{91.41 ± 0.55} & 90.17 ± 0.69          & \textbf{83.64 ± 2.00} & \textbf{87.55 ± 1.88} \\ \midrule
\multirow{4}{*}{TCL}                                                   & FTE & 96.47 ± 0.16          & 96.22 ± 0.17          & 97.53 ± 0.02          & 94.09 ± 0.07          & 82.38 ± 0.24          & 80.60 ± 0.22          & 67.27 ± 2.16          & 73.53 ± 1.66          \\
                                                                       & F-\ours     & 97.83 ± 0.05          & 97.58 ± 0.10          & 97.74 ± 0.03          & 94.75 ± 0.20          & 83.40 ± 1.32          & 81.75 ± 1.43          & \textbf{76.08 ± 0.79} & \textbf{80.68 ± 0.70} \\
                                                                       & S-\ours     & 97.33 ± 0.06          & 97.05 ± 0.13          & \textbf{97.78 ± 0.03} & \textbf{94.99 ± 0.07} & 83.87 ± 0.30          & 82.34 ± 0.31          & 69.92 ± 0.46          & 76.44 ± 0.42          \\
                                                                       & C-\ours     & \textbf{98.19 ± 0.04} & \textbf{97.89 ± 0.03} & 97.75 ± 0.09          & 94.83 ± 0.20          & \textbf{84.24 ± 0.10} & \textbf{82.72 ± 0.12} & 72.76 ± 4.64          & 78.70 ± 3.67          \\ \midrule
\multirow{4}{*}{\begin{tabular}[c]{@{}c@{}}DyG-\\ Former\end{tabular}} & FTE & 99.03 ± 0.02          & 98.59 ± 0.03          & 99.22 ± 0.01          & 98.84 ± 0.02          & 87.52 ± 0.49          & 86.96 ± 0.43          & 93.00 ± 0.12          & 94.23 ± 0.09          \\
                                                                       & F-\ours     & 99.04 ± 0.01          & 98.66 ± 0.05          & 99.22 ± 0.01          & 98.85 ± 0.02          & 87.60 ± 0.26          & 87.15 ± 0.22          & 93.06 ± 0.05          & 94.11 ± 0.09          \\
                                                                       & S-\ours     & 99.12 ± 0.01          & 98.72 ± 0.03          & 99.17 ± 0.10          & 98.78 ± 0.13          & 88.66 ± 0.20          & 88.37 ± 0.25          & 93.50 ± 0.12          & 94.57 ± 0.15          \\
                                                                       & C-\ours     & \textbf{99.13 ± 0.02} & \textbf{98.73 ± 0.00} & \textbf{99.24 ± 0.01} & \textbf{98.86 ± 0.01} & \textbf{88.70 ± 0.21} & \textbf{88.39 ± 0.15} & \textbf{93.64 ± 0.10} & \textbf{94.69 ± 0.12} \\ \bottomrule
\end{tabular}

}

\label{tab_tg_types_ap}
\end{table*}
\begin{table*}[h]

\caption{Comparing Functional Time Encoding (FTE), Fourier-based \ours~(F-\ours), Spline-based \ours~(S-\ours) and Combined \ours~(C-\ours): Dynamic graph link prediction results in AUC-ROC. The best results are in \textbf{bold}.}

\centering
\setlength{\tabcolsep}{2.7mm}{
\scriptsize
\begin{tabular}{c|c|cc|cc|cc|cc}
\toprule
                                                                       & AUC        & \multicolumn{2}{c|}{Wikipedia}            & \multicolumn{2}{c|}{Reddit}               & \multicolumn{2}{c|}{MOOC}                 & \multicolumn{2}{c}{LastFM}                \\ \cmidrule{2-10} 
                                                                       & TE         & Transductive        & Inductive           & Transductive        & Inductive           & Transductive        & Inductive           & Transductive        & Inductive           \\ \midrule
\multirow{4}{*}{TGAT}                                                  & FTE & 96.69 ± 0.26          & 95.95 ± 0.33          & 98.48 ± 0.04          & 96.90 ± 0.07          & 86.44 ± 0.24          & 86.04 ± 0.19          & 70.89 ± 0.10          & 76.11 ± 0.11          \\
                                                                       & F-\ours     & 96.53 ± 0.17          & 95.94 ± 0.20          & 98.48 ± 0.03          & 96.91 ± 0.03          & 86.36 ± 0.29          & 86.18 ± 0.33          & 70.53 ± 0.18          & 75.61 ± 0.22          \\
                                                                       & S-\ours     & 97.33 ± 0.07          & 96.76 ± 0.06          & 98.51 ± 0.02          & \textbf{96.96 ± 0.05} & 89.38 ± 0.14          & 89.14 ± 0.24          & 73.89 ± 0.51          & 79.17 ± 0.44          \\
                                                                       & C-\ours     & \textbf{97.63 ± 0.11} & \textbf{97.07 ± 0.10} & \textbf{98.51 ± 0.01} & 96.88 ± 0.05          & \textbf{89.46 ± 0.08} & \textbf{89.50 ± 0.05} & \textbf{74.24 ± 0.28} & \textbf{79.63 ± 0.14} \\ \midrule
\multirow{4}{*}{TGN}                                                   & FTE & 98.37 ± 0.07          & 97.72 ± 0.03          & 98.60 ± 0.06          & 97.39 ± 0.07          & 91.21 ± 1.15          & 91.24 ± 0.99          & 78.47 ± 2.94          & 82.61 ± 3.15          \\
                                                                       & F-\ours     & 98.50 ± 0.10          & 97.86 ± 0.09          & 98.63 ± 0.02          & 97.27 ± 0.10          & 91.71 ± 0.65          & 91.53 ± 0.33          & 78.24 ± 4.86          & 83.16 ± 6.28          \\
                                                                       & S-\ours     & 98.47 ± 0.06          & 97.86 ± 0.06          & \textbf{98.72 ± 0.00} & \textbf{97.55 ± 0.05} & \textbf{92.68 ± 0.37} & \textbf{92.41 ± 0.73} & 82.48 ± 2.17          & 86.49 ± 0.62          \\
                                                                       & C-\ours     & \textbf{98.73 ± 0.07} & \textbf{98.11 ± 0.10} & 98.72 ± 0.02          & 97.43 ± 0.11          & 92.65 ± 0.48          & 91.27 ± 0.85          & \textbf{83.94 ± 1.85} & \textbf{87.74 ± 1.75} \\ \midrule
\multirow{4}{*}{TCL}                                                   & FTE & 95.84 ± 0.18          & 95.57 ± 0.20          & 97.42 ± 0.02          & 93.80 ± 0.07          & 83.12 ± 0.18          & 81.43 ± 0.19          & 64.06 ± 1.16          & 70.84 ± 0.85          \\
                                                                       & F-\ours     & 97.35 ± 0.07          & 97.14 ± 0.13          & 97.62 ± 0.03          & 94.42 ± 0.22          & 83.73 ± 0.92          & 82.11 ± 0.98          & \textbf{70.17 ± 0.47} & \textbf{75.86 ± 0.44} \\
                                                                       & S-\ours     & 96.90 ± 0.08          & 96.62 ± 0.15          & \textbf{97.68 ± 0.03} & \textbf{94.63 ± 0.05} & 84.50 ± 0.20          & 83.02 ± 0.23          & 67.32 ± 0.50          & 74.37 ± 0.51          \\
                                                                       & C-\ours     & \textbf{97.84 ± 0.06} & \textbf{97.56 ± 0.02} & 97.64 ± 0.08          & 94.49 ± 0.17          & \textbf{84.73 ± 0.13} & \textbf{83.25 ± 0.23} & 69.16 ± 3.21          & 75.85 ±  2.64         \\ \midrule
\multirow{4}{*}{\begin{tabular}[c]{@{}c@{}}DyG-\\ Former\end{tabular}} & FTE & 98.91 ± 0.02          & 98.48 ± 0.03          & 99.15 ± 0.01          & 98.71 ± 0.01          & 87.91 ± 0.58          & 87.62 ± 0.51          & 93.05 ± 0.10          & 94.08 ± 0.08          \\
                                                                       & F-\ours     & 98.94 ± 0.02          & 98.55 ± 0.03          & 99.16 ± 0.01          & 98.72 ± 0.03          & 88.09 ± 0.16          & 87.89 ± 0.14          & 93.09 ± 0.03          & 94.00 ± 0.04          \\
                                                                       & S-\ours     & \textbf{99.04 ± 0.01} & \textbf{98.67 ± 0.02} & 99.08 ± 0.13          & 98.61 ± 0.19          & \textbf{89.18 ± 0.21} & \textbf{89.16 ± 0.30} & 93.56 ± 0.06          & 94.46 ± 0.06          \\
                                                                       & C-\ours     & 99.04 ± 0.02          & 98.65 ± 0.01          & \textbf{99.17 ± 0.00} & \textbf{98.74 ± 0.01} & 89.17 ± 0.20          & 89.14 ± 0.09          & \textbf{93.65 ± 0.07} & \textbf{94.52 ± 0.08} \\ \bottomrule
\end{tabular}

}

\label{tab_tg_types_auc}
\end{table*}

\begin{figure*}[!tbp]
    \centering
    \includegraphics[width=0.8\linewidth]{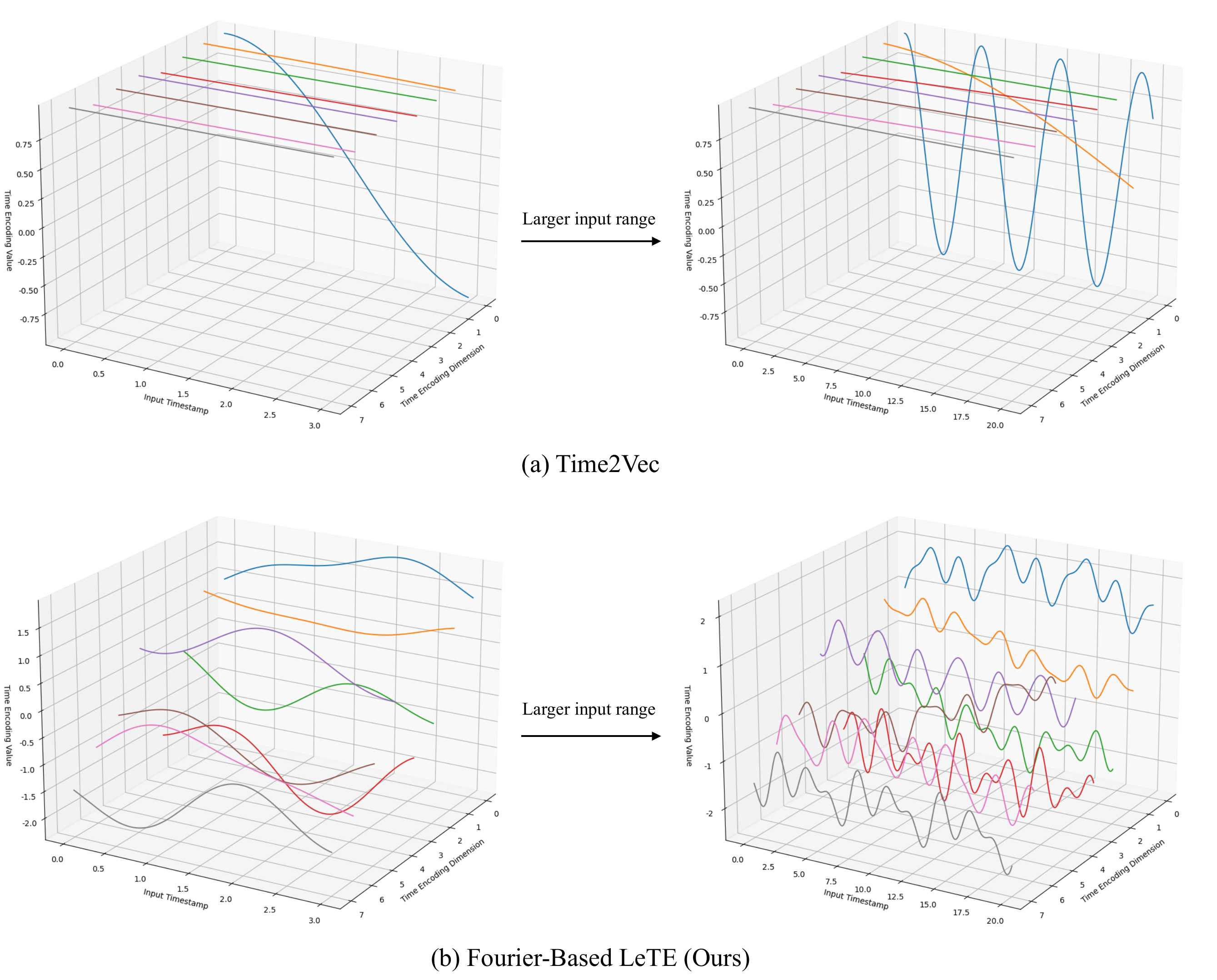}
\caption{Example of feature map for the FTE and Fourier-based \ours~at different dimensions (the total dimension is 8, the parameters weight are based on a learned TGN model on Wikipedia dataset).}
\label{fig:te_visual}
\end{figure*}

\begin{figure}[!tbp]
    \centering
    \includegraphics[width=0.65\linewidth]{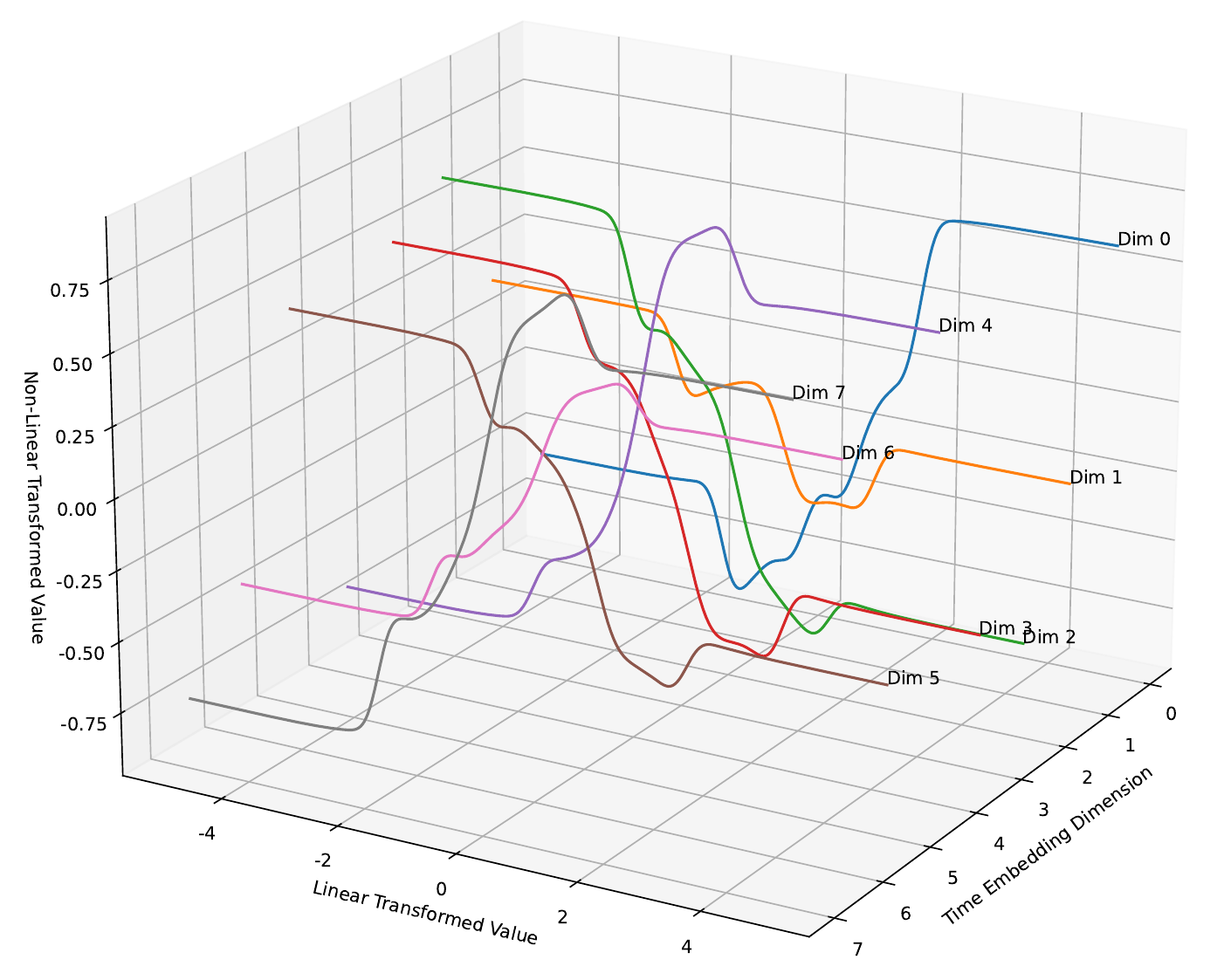}
\caption{Example of non-linear transformation for Spline-based \ours~at different dimensions.}
\label{fig:spline_trans}
\end{figure}

\subsection{Visualization and Interpretability}
\label{app_visual}

As mentioned in Section \ref{sec:properties}, the learned parameters of our proposed method can be used to reconstruct the time embedding feature map or the non-linear transformations. Since the previous time encoding method (FTE) uses fixed non-linear transformation functions, we present an example of its feature map and compare it with the Fourier-based \ours~in Figure \ref{fig:te_visual}. We compare the local and global mappings of the two time encoding methods. As shown in the figure, our method captures richer and more detailed time patterns in different dimensions of the time encoding for local mappings. By contrast, the FTE method exhibits periodicity in only one dimension. This occurs because the learned frequency parameters in other dimensions are too small to capture sufficient periodicity locally. Similarly, for global mappings, strong periodicity is still observed in the feature map of our method, alongside varying degrees of non-periodicity. The FTE continues to exhibit periodicity in only one dimension. In other dimensions, the similar frequency parameters result in insufficient periodicity modeling and a lack of non-periodic pattern representation.

We also present the non-linear transformation sketches of Spline-based \ours~in Figure \ref{fig:spline_trans}. The figure shows that the learned non-linear activation functions vary across dimensions, significantly enhancing the model’s expressiveness. Additionally, it demonstrates the modeling of both periodic and non-periodic patterns at local and global scales. Since Combined \ours~is a combination of Fourier-based \ours~and Spline-based \ours, it is intuitive that Combined \ours~inherits the same interpretability as its components. 
Furthermore, as our method supports the reconstruction of non-linear activation functions, it retains strong interpretability.

\begin{figure}[!t]
\includegraphics[width=1\linewidth]{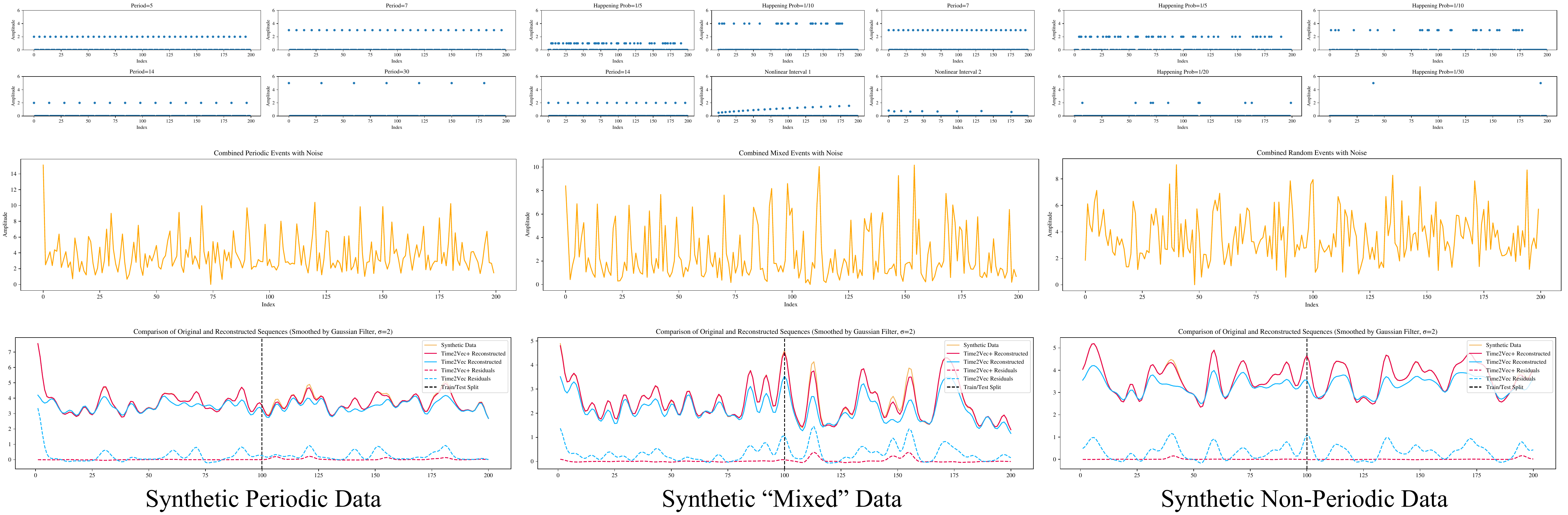}
\caption{Capturing periodic, non-periodic and mixed patterns in synthetic data.}
\label{fig:case_signal}
\end{figure}

\begin{figure}[!t]
\includegraphics[width=1\linewidth]{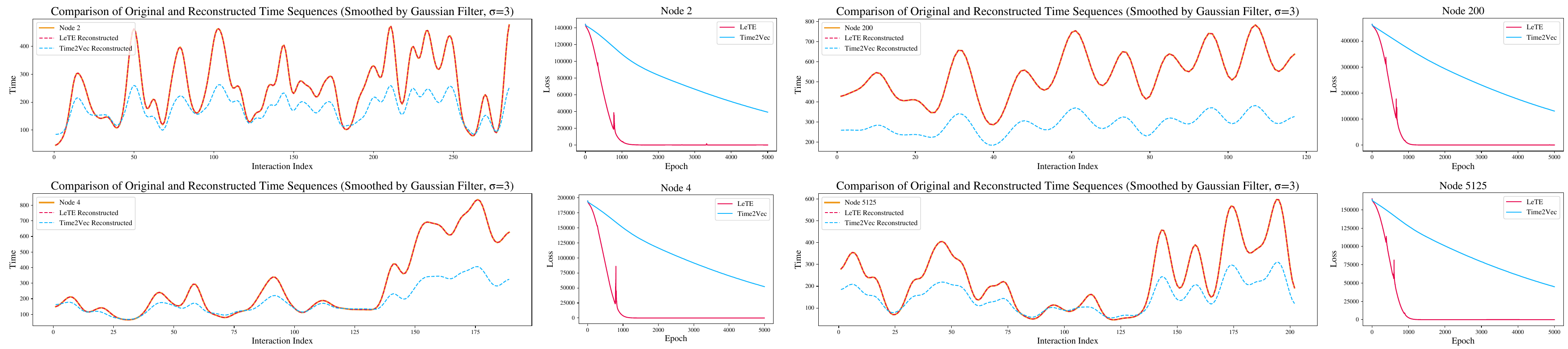}
\caption{Capturing periodic, non-periodic and mixed patterns in real data.}
\label{fig:case_wiki}
\end{figure}

\subsection{Capturing Periodic, Non-Periodic and Mixed Patterns in Data}
\label{app_p_and_nonp}

Complex periodic and non-periodic patterns often coexist in real-world data, forming mixed time patterns. To demonstrate that our method surpasses previous methods in modeling such patterns, we design a mini reconstruction task using both synthetic data and real data from the Wikipedia dataset. Specifically, we construct an encoder-decoder model to reconstruct the data. The encoder is either ($d$-dimensional) our \ours~or the FTE, while the decoder is a simple linear layer mapping a $d$-dimensional vector to a 1-dimensional output. The reconstruction objective minimizes the MSE loss, which also quantifies the modeling capability of the time encodings. Additionally, the reconstructed time sequence plots visually indicate the models’ ability to fit the data.

To isolate periodic and non-periodic patterns, we first generate synthetic data containing purely periodic signals, purely non-periodic signals, and mixed signals. These data is used to evaluate the performance of different encoders (\ours~or FTE). The ground truth and reconstructed sequences are shown in Figure \ref{fig:case_signal}. As illustrated, the FTE method performs reasonably well on periodic data but struggles with non-periodic and mixed data. In contrast, our method consistently outperforms FTE, demonstrating its capability to model both periodic and non-periodic patterns effectively.

Real-world data often exhibit complex combinations of periodic and non-periodic patterns, i.e., mixed patterns. To further evaluate our method, we randomly select 4 nodes’ interaction sequences from the Wikipedia dataset and perform the same reconstruction experiments. The results are presented in Figure \ref{fig:case_wiki}, where the time sequences are smoothed using a 1D Gaussian filter for clarity. As shown, the time sequences reconstructed using our \ours~align more closely with the original data compared to those reconstructed using FTE. Additionally, the loss of our \ours~is significantly lower than that of FTE, further validating our method’s ability to capture complex periodic and non-periodic patterns in real-world data.

The experimental results show that, regardless of whether the sequence is periodic or non-periodic, our method consistently outperforms better. This is primarily due to the incorporation of learnable non-linear transformations into our time encoding approach.

\begin{figure}[!tbp]
    \centering
    \includegraphics[width=1\linewidth]{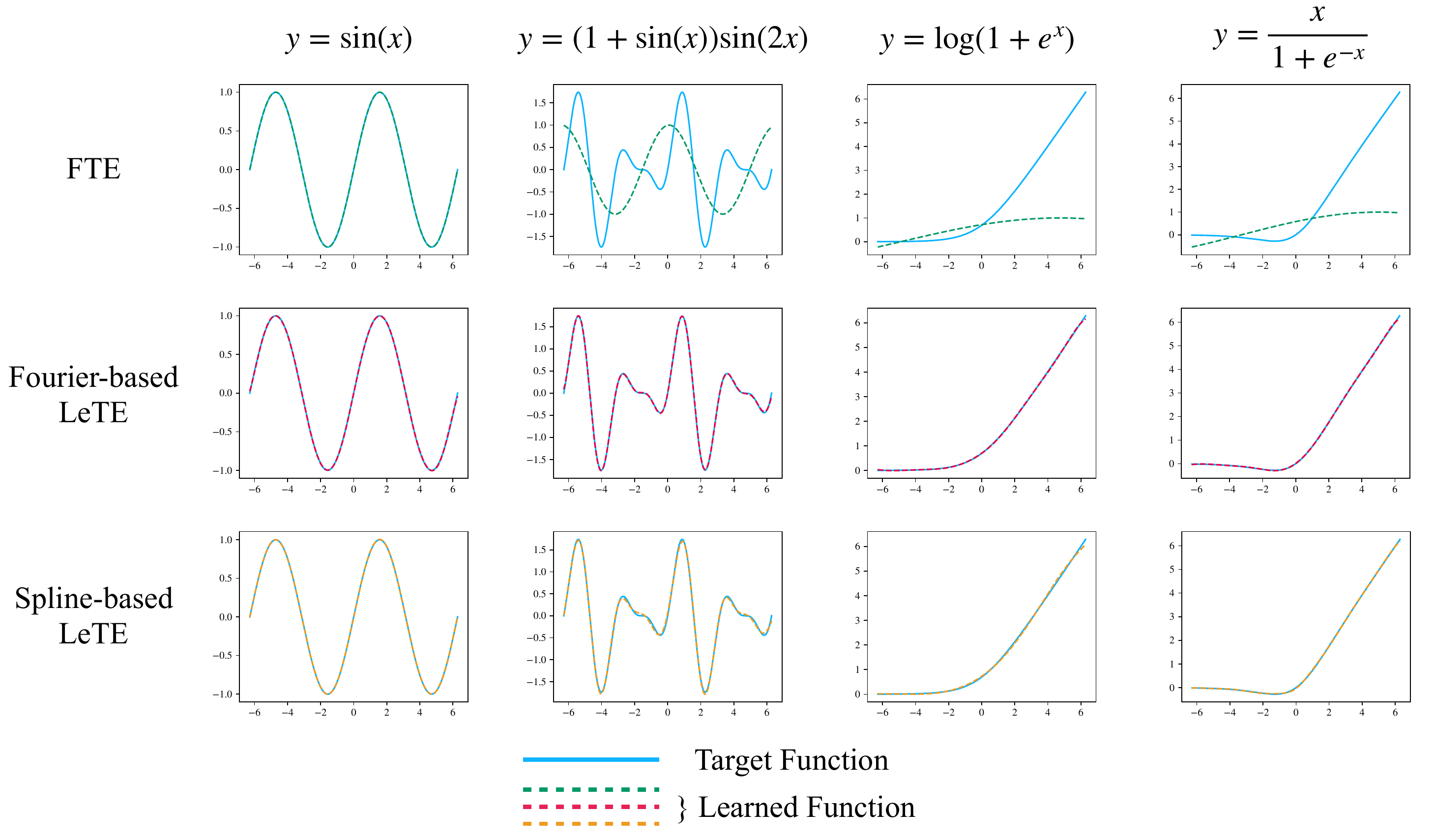}
\caption{FTE, Fourier-based \ours~and Spline-based \ours~fitting different functions.}
\label{fig:fitting}
\end{figure}

\subsection{Fitting Ability}
\label{app_fitting_ability}
We conduct a simple toy experiment to further demonstrate that both Fourier-based \ours~and Spline-based \ours~are capable of capturing different patterns. Consequently, Combined \ours~inherits this ability as well.
To illustrate this, we generate a set of training data using 4 different non-linear transformation functions. Two of these functions are periodic: the sine function $y = \sin(x)$, and a more complex periodic function $y = (1+\sin(x))\sin(2x)$. The other two functions are non-periodic: the Softplus activation function $y = \log(1+e^x)$ \cite{softplus}, and the Swish activation function $y = \frac{x}{1+e^{-x}}$ \cite{swish}.

We fit the data using simple 1-dimensional FTE, Fourier-based \ours~and Spline-based \ours, evaluating their ability to capture complex patterns, including both periodic and non-periodic. The learned non-linear transformation functions are plotted in Figure \ref{fig:fitting}. As shown in the figure, both Fourier-based \ours~and Spline-based \ours~successfully capture diverse patterns. 
We also compare our method with FTE. Due to the fixed non-linear transformation functions used in FTE, it fails to capture the complex periodic and non-periodic patterns present in the data.
These results demonstrate that our proposed \ours~has the capability to model complex patterns in data effectively and is more general than previous time encoding methods.

\section{More Explanations and Examples about Interpretability}
We choose to use a 4-dimensional Combined LeTE to present our analysis related to interpretability of LeTE. The experiments are conducted on the Wikipedia and MOOC datasets, with TGN and DyGFormer as backbone models. The training process and settings are consistent with those used in the main experiments. We will demonstrate the interpretability of our model from the following perspectives:
\begin{enumerate}
\item Reconstructing the learned non-linear transformation functions and plotting them to provide a clear and intuitive analysis.
\item Analyzing each dimension to understand what information it represents. Specifically, the first two dimensions of the time encoding are Fourier-based, while the last two dimensions are Spline-based.
\item Comparing different datasets under the same backbone model.
\item Comparing different backbone models' LeTE under the same dataset.
\item Comparing the plots of low- vs. high-dimensional LeTE to assess the impact of dimensionality on interpretability.
\end{enumerate}

\subsection{Reconstructing}
As previously discussed, the previous time encoding methods exhibit a degree of interpretability by using fixed sinusoidal functions, which inherently reflect periodic patterns. However, this strong inductive bias also limits their expressiveness and generalization to complex or non-periodicity.

In contrast, our proposed LeTE is a fully learnable time encoding, and the learnable non-linear functions can still be reconstructed and visualized from learned parameters, allowing for interpretability analysis through function inspection.

\begin{figure*}
    \centering
    \includegraphics[width=1\linewidth]{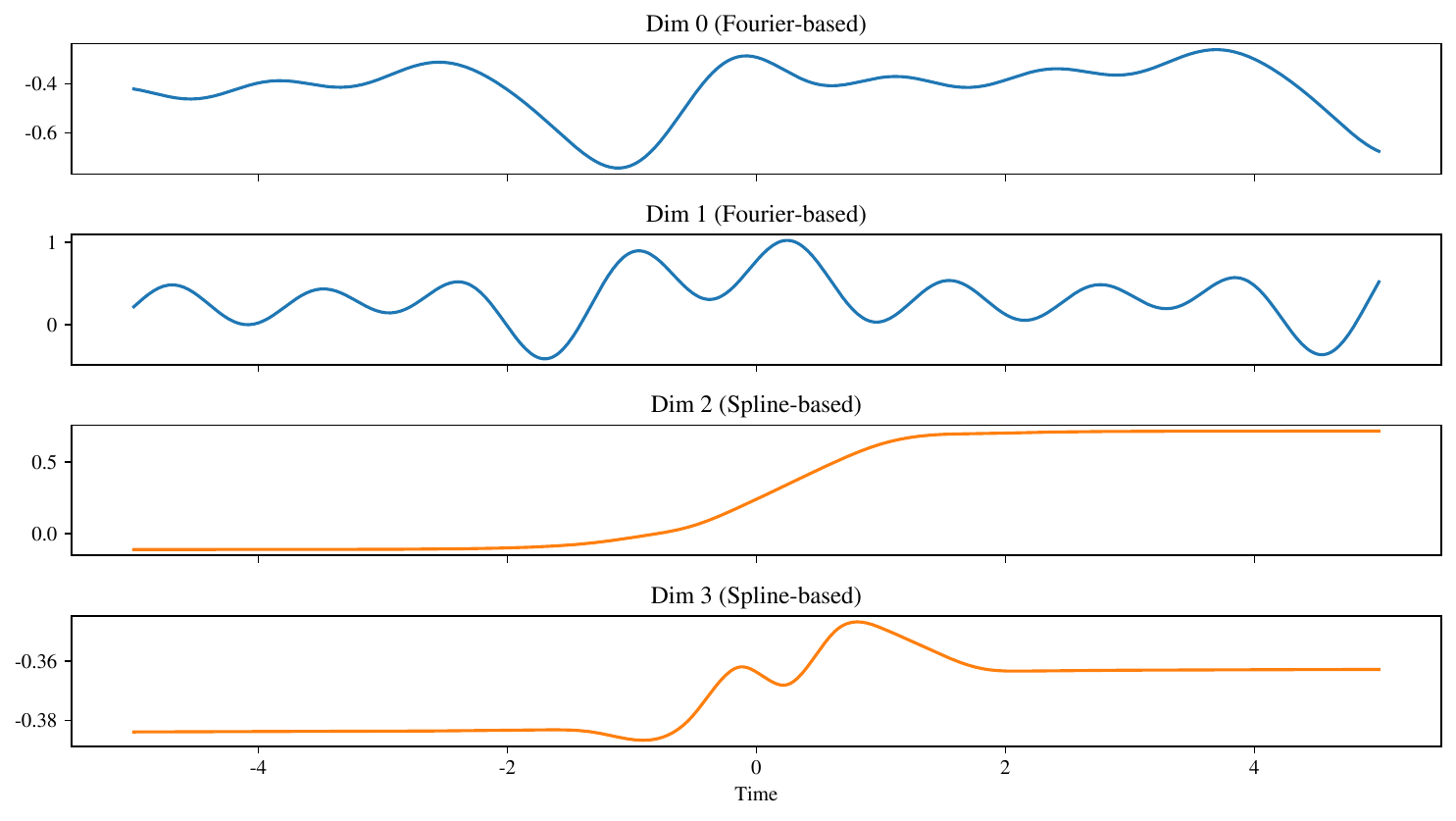}
    \caption{Plots of the four dimensions of the non-linear transformation functions of a 4-dimensional LeTE trained on Wikipedia/TGN. We further present the four functions here, the parameters are learned and read from the trained model:\\ \\
    $f_{0}(x) =-0.0444 \cdot \cos(1\cdot x'_0 )+0.0758 \cdot \sin(1\cdot x'_0 )+0.0875 \cdot \cos(2\cdot x'_0 )+0.0704 \cdot \sin(2\cdot x'_0 )+0.0712 \cdot \cos(3\cdot x'_0 )-0.0327 \cdot \sin(3\cdot x'_0 )+0.0040 \cdot \cos(4\cdot x'_0 )-0.0340 \cdot \sin(4\cdot x'_0 )+0.0150 \cdot \cos(5\cdot x'_0 )-0.0220 \cdot \sin(5\cdot x'_0 )+0.0710 \cdot \cos(1\cdot x'_1 )+0.1506 \cdot \sin(1\cdot x'_1 )-0.1483 \cdot \cos(2\cdot x'_1 )+0.2502 \cdot \sin(2\cdot x'_1 )+0.2938 \cdot \cos(3\cdot x'_1 )+0.0878 \cdot \sin(3\cdot x'_1 )-0.1641 \cdot \cos(4\cdot x'_1 )+0.0640 \cdot \sin(4\cdot x'_1 )-0.4155 \cdot \cos(5\cdot x'_1 )+0.0395 \cdot \sin(5\cdot x'_1 )-0.0762$,\\ \\
    $f_{1}(x) =+0.1860 \cdot\cos(1\cdot x'_0 )+0.0267 \cdot\sin(1\cdot x'_0 )+0.1971 \cdot\cos(2\cdot x'_0 )-0.0510 \cdot\sin(2\cdot x'_0 )-0.0225 \cdot\cos(3\cdot x'_0 )-0.0909 \cdot\sin(3\cdot x'_0 )-0.0501 \cdot\cos(4\cdot x'_0 )+0.1460 \cdot\sin(4\cdot x'_0 )+0.0952 \cdot\cos(5\cdot x'_0 )+0.2974 \cdot\sin(5\cdot x'_0 )-0.1604 \cdot\cos(1\cdot x'_1 )+0.9609 \cdot\sin(1\cdot x'_1 )+0.2323 \cdot\cos(2\cdot x'_1 )-0.3430 \cdot\sin(2\cdot x'_1 )-0.0441 \cdot\cos(3\cdot x'_1 )-0.1428 \cdot\sin(3\cdot x'_1 )+0.2930 \cdot\cos(4\cdot x'_1 )+0.6073 \cdot\sin(4\cdot x'_1 )+0.0330 \cdot\cos(5\cdot x'_1 )-0.1345 \cdot\sin(5\cdot x'_1 )-0.0130$,\\ \\
    here, for both $f_{0}(x)$ and $f_{1}(x)$, $x'_0 = 1.0069\cdot x+0.0069$ and $x'_1 = 0.0054\cdot x+0.0108.$\\ \\
    $f_{2}(x) = +0.0013 \cdot B_{0}(x) \text{ (support: [-2.20,-1.80])}+0.0047 \cdot B_{1}(x) \text{ (support: [-1.80,-1.40])}-0.0353 \cdot B_{2}(x) \text{ (support: [-1.40,-1.00])}-0.0321 \cdot B_{3}(x) \text{ (support: [-1.00,-0.60])}-0.0455 \cdot B_{4}(x) \text{ (support: [-0.60,-0.20])}-0.0273 \cdot B_{5}(x) \text{ (support: [-0.20, 0.20])}+0.0211 \cdot B_{6}(x) \text{ (support: [0.20, 0.60])}+0.0248 \cdot B_{7}(x) \text{ (support: [0.60, 1.00])}+0.4133 \cdot \mathrm{Tanh}(x)$, \\ \\
    $f_{3}(x) =-0.0008 \cdot B_{0}(x) \text{ (support: [-2.20,-1.80])}-0.0072 \cdot B_{1}(x) \text{ (support: [-1.80,-1.40])}-0.0040 \cdot B_{2}(x) \text{ (support: [-1.40,-1.00])}+0.0227 \cdot B_{3}(x) \text{ (support: [-1.00,-0.60])}-0.0067 \cdot B_{4}(x) \text{ (support: [-0.60,-0.20])}+0.0248 \cdot B_{5}(x) \text{ (support: [-0.20, 0.20])}+0.0165 \cdot B_{6}(x) \text{ (support: [0.20, 0.60])}+0.0078 \cdot B_{7}(x) \text{ (support: [0.60, 1.00])}+0.0100 \cdot \mathrm{Tanh}(x)$.
    }
    \label{fig:wiki_tgn_4}
\end{figure*}

We demonstrate this interpretability using a 4-dimensional Combined LeTE, trained on the Wikipedia/TGN. Figure \ref{fig:wiki_tgn_4} shows the learned transformation functions for each dimension. The first two dimensions are Fourier-based, and the last two are Spline-based.

\subsection{Analyzing Each Dimension}
\textbf{Fourier-based:}
The Fourier coefficients explicitly encode frequency components, offering a clear, intuitive view of the captured periodicity. Compared to fixed sinusoidal functions, our learnable Fourier-based time encoding captures periodic patterns with finer granularity and greater flexibility, enabling the representation of both periodic signals and subtle non-periodicities within specific ranges.

\begin{figure*}
    \centering
    \begin{subfigure}[b]{0.45\linewidth}
        \includegraphics[width=1\linewidth]{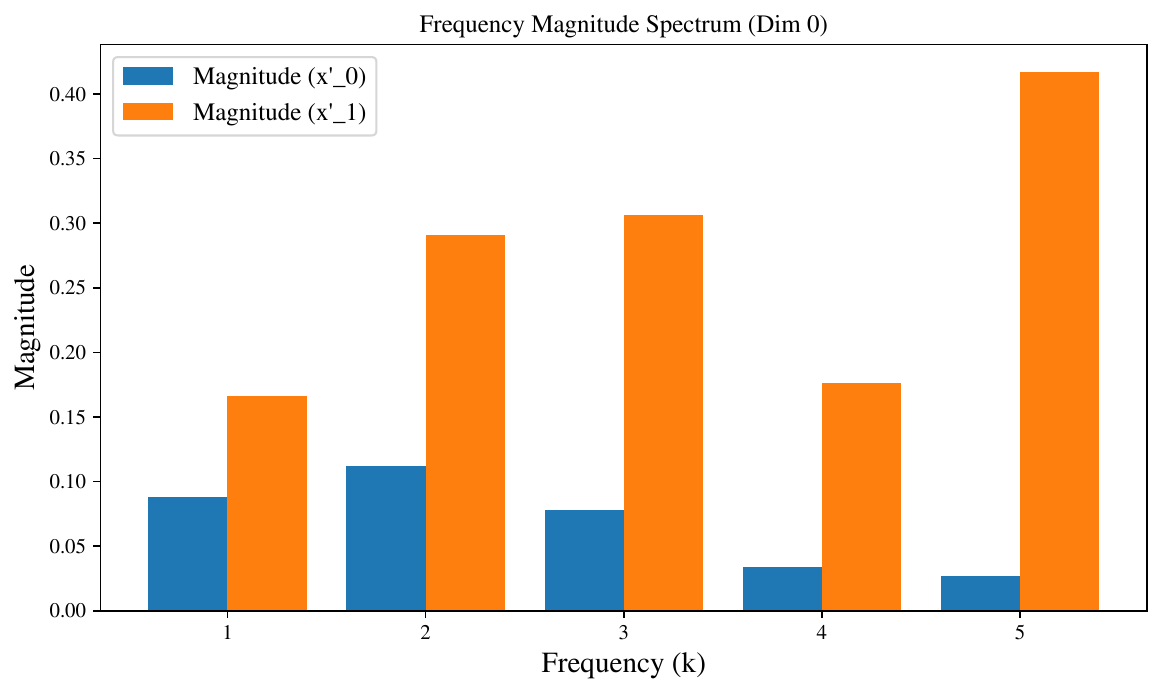}
        \caption{Dim 0 (Fourier-based)}
    \end{subfigure}
    \hspace{20pt}
    \begin{subfigure}[b]{0.45\linewidth}
        \includegraphics[width=1\linewidth]{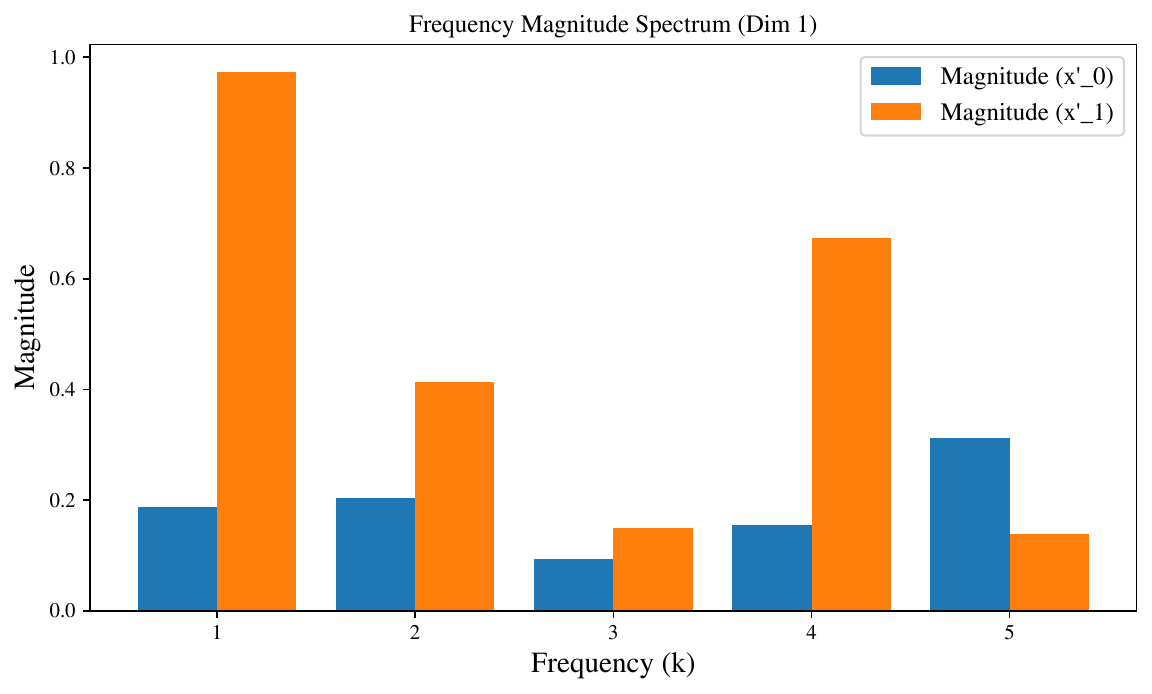}
        \caption{Dim 1 (Fourier-based)}
    \end{subfigure}

\caption{Frequency Magnitude Spectrum for the first two dimensions of the non-linear transformation functions of LeTE trained on Wikipedia/TGN. These two dimensions are Fourier-based.}
\label{fig:spectrums}
\end{figure*}

For a single dimension, low-frequency components capture long-term trends, while high-frequency components focus on short-term fluctuations. This allows the model to encode both long-term dynamics and short-term variations simultaneously. As an example, we apply this to the Wikipedia dataset, which records editing activities, where nodes represent users or pages, and edges with timestamps capture editing events (frequency magnitude spectrum is shown in Figure \ref{fig:spectrums}, note that the inputs are time differences in this case).

Specifically, Dim 0 shows a strong high-frequency response. The learned coefficients include
$\cos(3x{\prime}): +0.29$, $ \cos(4x{\prime}): -0.16$, $ \cos(5x{\prime}): -0.42$. This suggests that Dim 0 is sensitive to short-term repetitive edits, i.e., high-frequency editing behavior. 

Dim 1 captures low- to mid-frequency patterns, with large coefficients: $\sin(1x{\prime}): +0.96$, $\sin(4x{\prime}): +0.61$, $\cos(4x{\prime}): +0.29$. These reflect longer-term periodic behaviors. For example, frequency-1 may correspond to daily or weekly editing cycles, while frequency-4 may capture sub-daily repeated interactions. This dimension may reflect user habits or regular community editing patterns. Thus, LeTE's Fourier-based dimensions not only retain the periodic interpretability of sine functions but also exhibit richer frequency composition, allowing it to simultaneously capture both short-term bursts and long-term rhythms.

In addition, this approach could be extended to analyze more complex patterns. However, as our goal here is to present the underlying idea, we will not go deeper here.

\textbf{Spline-based:} The Spline-based functions offer complementary advantages, particularly for non-periodicity.

In our spline-based dimensions, where we applied a basis function (Tanh), if the weight of the basis function is higher, it may dominate a specific dimension—such as Dim 2 in Figure \ref{fig:wiki_tgn_4}. However, there are other dimensions where splines dominate, such as Dim 3. To further clarify this, we combine the specific Wikipedia dataset and explain:

Dim 2: The output increases monotonically with time difference, indicating a time-decay-like effect — the longer the time since last edit, the stronger the encoding response. This may suggest the time encoding has learned that re-activation after long inactivity is a significant event in this specific case.

Dim 3: The function exhibits sharp peaks and local bumps, indicating that the model assigns particular importance to certain time intervals. These may correspond to known active editing windows or reaction delays. The sharpness of some coefficients suggests the model has captured rare but important temporal phenomena, such as one-off campaigns or anomaly spikes.

The Spline coefficients inherently capture local temporal features, indicating specific time intervals that the model considers critical or active. Sharp peaks coefficients within these curves suggest the occurrence of sudden events or anomalies. This local characteristic is advantageous for identifying rare phenomena.

\subsection{Different Datasets}
\begin{figure*}
    \centering
    \includegraphics[width=1\linewidth]{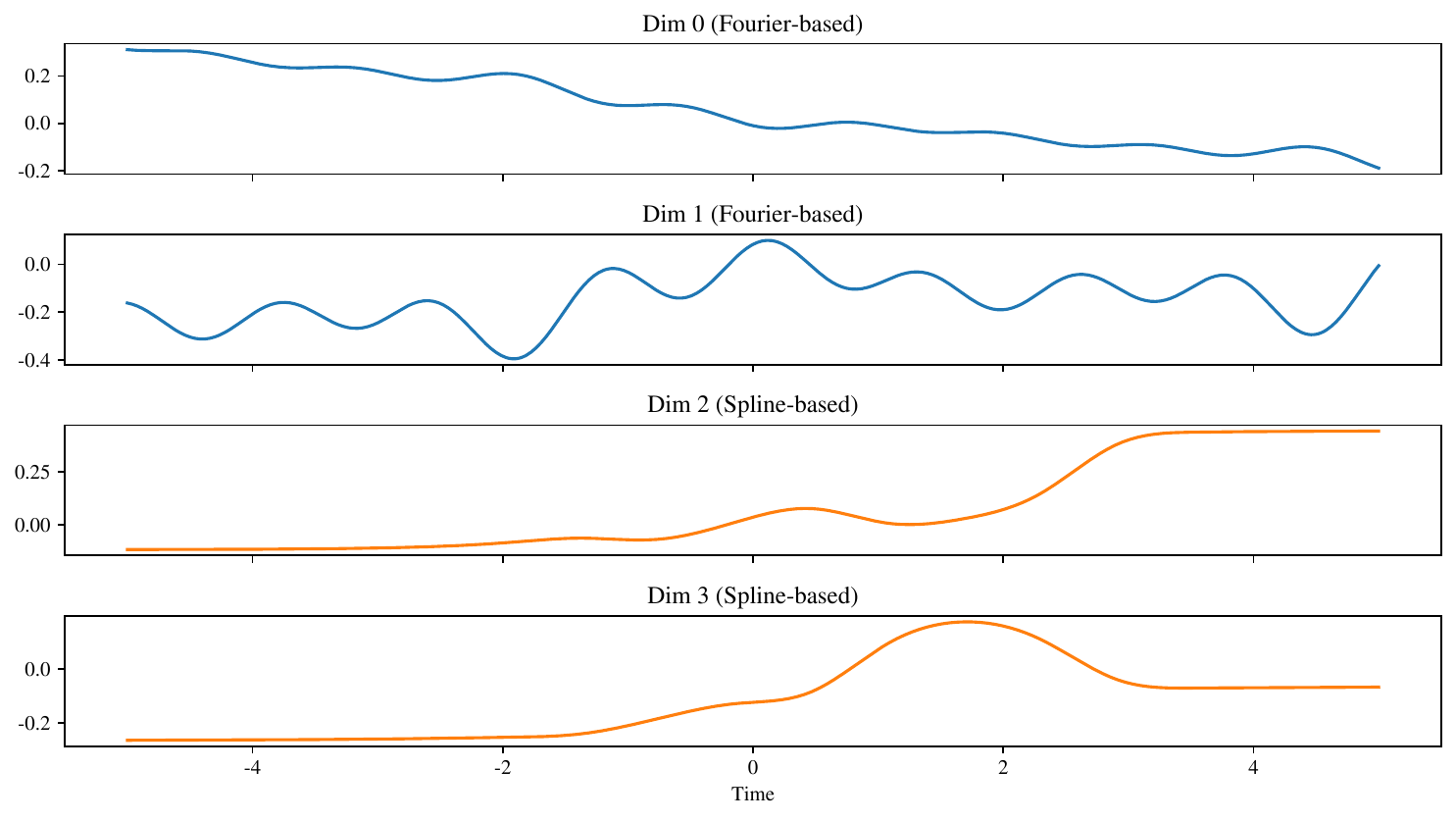}
    \caption{Plots of the four dimensions of the non-linear transformation functions of a 4-dimensional LeTE trained on MOOC/TGN. We further present the four functions here, the parameters are learned and read from the trained model:\\ \\
    $f_{0}(x) = -0.0186 \cdot\cos(1\cdot x'_0)  - 0.0131 \cdot\sin(1\cdot x'_0)  - 0.0173 \cdot\cos(2\cdot x'_0)  + 0.0039 \cdot\sin(2\cdot x'_0)  + 0.0019 \cdot\cos(3\cdot x'_0)  + 0.0013 \cdot\sin(3\cdot x'_0)  + 0.0032 \cdot\cos(4\cdot x'_0)  - 0.0096 \cdot\sin(4\cdot x'_0)  - 0.0134 \cdot\cos(5\cdot x'_0)  - 0.0136 \cdot\sin(5\cdot x'_0)  + 0.3140 \cdot\cos(1\cdot x'_1)  + 0.5998 \cdot\sin(1\cdot x'_1)  - 0.2781 \cdot\cos(2\cdot x'_1)  + 0.4591 \cdot\sin(2\cdot x'_1)  + 0.1376 \cdot\cos(3\cdot x'_1)  + 0.2296 \cdot\sin(3\cdot x'_1)  - 0.0880 \cdot\cos(4\cdot x'_1)  - 0.0198 \cdot\sin(4\cdot x'_1)  - 0.2621 \cdot\cos(5\cdot x'_1)  + 0.0030 \cdot\sin(5\cdot x'_1)  + 0.0016$,  \\ \\
    $f_{1}(x) = +0.0816 \cdot\cos(1\cdot x'_0)  + 0.0362 \cdot\sin(1\cdot x'_0)  + 0.0588 \cdot\cos(2\cdot x'_0)  - 0.0046 \cdot\sin(2\cdot x'_0)  - 0.0199 \cdot\cos(3\cdot x'_0)  - 0.0082 \cdot\sin(3\cdot x'_0)  - 0.0121 \cdot\cos(4\cdot x'_0)  + 0.0325 \cdot\sin(4\cdot x'_0)  + 0.0591 \cdot\cos(5\cdot x'_0)  + 0.0609 \cdot\sin(5\cdot x'_0)  - 0.4787 \cdot\cos(1\cdot x'_1)  + 1.1011 \cdot\sin(1\cdot x'_1)  + 0.0467 \cdot\cos(2\cdot x'_1)  - 0.4438 \cdot\sin(2\cdot x'_1)  - 0.1300 \cdot\cos(3\cdot x'_1)  - 0.2972 \cdot\sin(3\cdot x'_1)  + 0.3344 \cdot\cos(4\cdot x'_1)  + 0.1396 \cdot\sin(4\cdot x'_1)  + 0.0593 \cdot\cos(5\cdot x'_1)  - 0.0841 \cdot\sin(5\cdot x'_1)  + 0.1166$, \\ \\
    here, for both $f_{0}(x)$ and $f_{1}(x)$, $x'_0 = 0.9857\cdot x + 0.0971$ and $x'_1 = -0.0187\cdot x + 0.0885$.\\ \\
    $f_{2}(x) = +0.0020 \cdot B_{0}(x) \text{ (support: [-2.20, -1.80])}  + 0.0071 \cdot B_{1}(x) \text{ (support: [-1.80, -1.40])}  - 0.0979 \cdot B_{2}(x) \text{ (support: [-1.40, -1.00])}  - 0.0916 \cdot B_{3}(x) \text{ (support: [-1.00, -0.60])}  - 0.1039 \cdot B_{4}(x) \text{ (support: [-0.60, -0.20])}  - 0.3442 \cdot B_{5}(x) \text{ (support: [-0.20, 0.20])}  - 0.3597 \cdot B_{6}(x) \text{ (support: [0.20, 0.60])}  - 0.3093 \cdot B_{7}(x) \text{ (support: [0.60, 1.00])} + 0.2827 \cdot \mathrm{Tanh}(x) $, \\ \\
    $f_{3}(x) = - 0.0011 \cdot B_{0}(x) \text{ (support: [-2.20, -1.80])}  - 0.0099 \cdot B_{1}(x) \text{ (support: [-1.80, -1.40])}  + 0.0364 \cdot B_{2}(x) \text{ (support: [-1.40, -1.00])}  + 0.0712 \cdot B_{3}(x) \text{ (support: [-1.00, -0.60])}  + 0.0169 \cdot B_{4}(x) \text{ (support: [-0.60, -0.20])}  + 0.2372 \cdot B_{5}(x) \text{ (support: [-0.20, 0.20])}  + 0.2886 \cdot B_{6}(x) \text{ (support: [0.20, 0.60])}  + 0.2155 \cdot B_{7}(x) \text{ (support: [0.60, 1.00])} + 0.0947 \cdot \mathrm{Tanh}(x) $.
    }
    \label{fig:mooc_tgn_4}
\end{figure*}

We reconstructed and plotted the four non-linear functions for a 4-dimensional LeTE trained on MOOC/TGN (shown in Figure \ref{fig:mooc_tgn_4}). By comparing these results to those from the Wikipedia (Figure \ref{fig:wiki_tgn_4}), it can be seen that the Dim 0 exhibit a lack of periodicity. From the reconstructed equations of Dim 0, the higher-frequency terms do have coefficients with some magnitude, but they are generally small. For instance, the coefficients of $\cos(5x{\prime})$ and $\sin(5x{\prime})$ are relatively small (e.g., $-0.0134$ and $-0.0136$), suggesting that their contribution is minimal and insufficient to generate significant fluctuations. As a result, the overall function primarily exhibits slow oscillations, making the plot appear to be predominantly non-periodic within a certain input window.

This observation aligns with the findings in Appendix \ref{app_periodic_and_non} and Figure \ref{fig:entropy}, where the spectral entropy statistics also show that the Wikipedia exhibits stronger periodicity compared to the MOOC.

Thus, by comparing the non-linear functions of LeTE across different datasets, we can indirectly explore the periodic or non-periodic nature of the data present.

\subsection{Different Backbones}
\begin{figure*}
    \centering
    \includegraphics[width=1\linewidth]{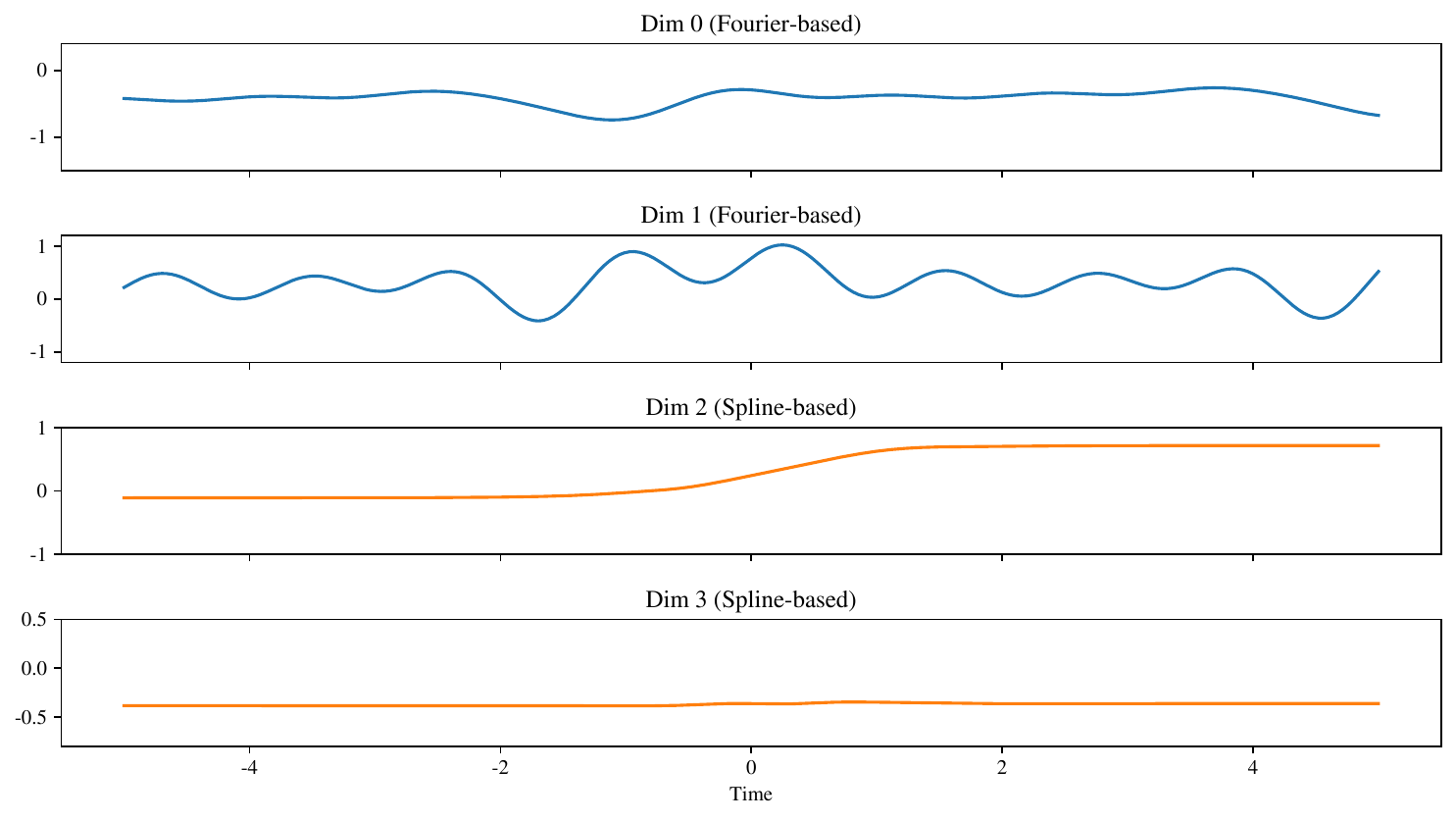}
    \caption{Plots of the four dimensions of the non-linear transformation functions of a 4-dimensional LeTE trained on Wikipedia/TGN. The y-axes have been scaled to the same level as \ref{fig:wiki_dyg_4_scaled} to facilitate a direct comparison. The reconstructed functions are the same as in the Figure \ref{fig:wiki_tgn_4}.}
    \label{fig:wiki_tgn_4_scaled}
\end{figure*}

\begin{figure*}
    \centering
    \includegraphics[width=1\linewidth]{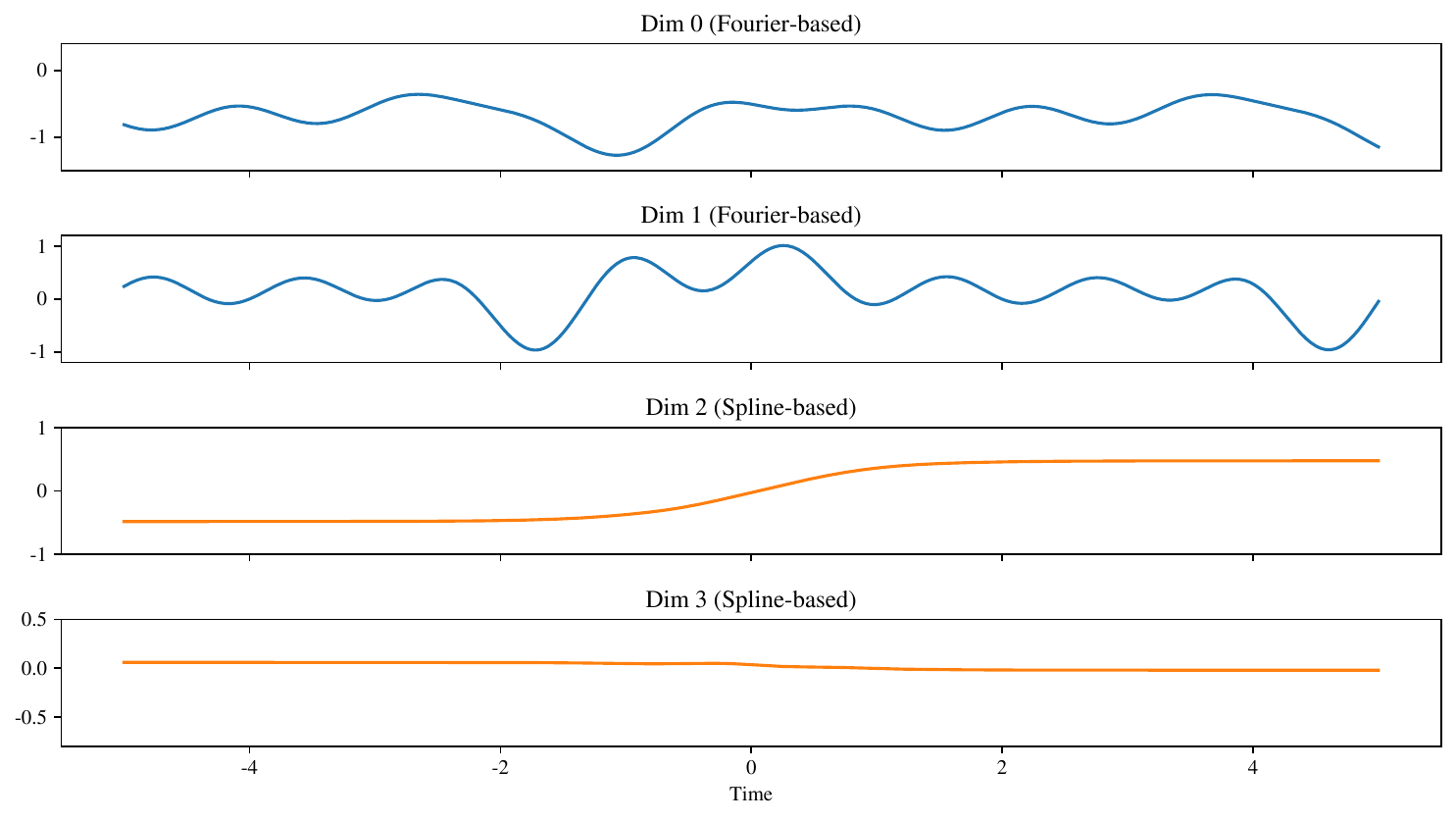}
    \caption{Plots of the four dimensions of the non-linear transformation functions of a 4-dimensional LeTE trained on Wikipedia/TGN. The y-axes have been scaled to the same level as \ref{fig:wiki_tgn_4_scaled} to facilitate a direct comparison. We further present the four functions here, the parameters are learned and read from the trained model:\\ \\
    $f_{0}(x) = -0.0727 \cdot\cos(1\cdot x'_0)  + 0.0704 \cdot\sin(1\cdot x'_0)  + 0.1529 \cdot\cos(2\cdot x'_0)  + 0.1720 \cdot\sin(2\cdot x'_0)  + 0.1554 \cdot\cos(3\cdot x'_0)  - 0.0320 \cdot\sin(3\cdot x'_0)  - 0.0385 \cdot\cos(4\cdot x'_0)  - 0.0220 \cdot\sin(4\cdot x'_0)  - 0.0086 \cdot\cos(5\cdot x'_0)  - 0.1107 \cdot\sin(5\cdot x'_0)  + 0.0758 \cdot\cos(1\cdot x'_1)  + 0.2909 \cdot\sin(1\cdot x'_1)  - 0.2430 \cdot\cos(2\cdot x'_1)  + 0.3166 \cdot\sin(2\cdot x'_1)  + 0.2097 \cdot\cos(3\cdot x'_1)  + 0.0334 \cdot\sin(3\cdot x'_1)  - 0.3308 \cdot\cos(4\cdot x'_1)  - 0.0664 \cdot\sin(4\cdot x'_1)  - 0.4329 \cdot\cos(5\cdot x'_1)  - 0.1351 \cdot\sin(5\cdot x'_1)  - 0.0009$,  \\ \\
    $f_{1}(x) = +0.2232 \cdot\cos(1\cdot x'_0)  + 0.1531 \cdot\sin(1\cdot x'_0)  + 0.2963 \cdot\cos(2\cdot x'_0)  - 0.0852 \cdot\sin(2\cdot x'_0)  - 0.0409 \cdot\cos(3\cdot x'_0)  - 0.1263 \cdot\sin(3\cdot x'_0)  - 0.0938 \cdot\cos(4\cdot x'_0)  + 0.1772 \cdot\sin(4\cdot x'_0)  + 0.1431 \cdot\cos(5\cdot x'_0)  + 0.3435 \cdot\sin(5\cdot x'_0)  - 0.2795 \cdot\cos(1\cdot x'_1)  + 1.0935 \cdot\sin(1\cdot x'_1)  + 0.1297 \cdot\cos(2\cdot x'_1)  - 0.4577 \cdot\sin(2\cdot x'_1)  - 0.0711 \cdot\cos(3\cdot x'_1)  - 0.3519 \cdot\sin(3\cdot x'_1)  + 0.4151 \cdot\cos(4\cdot x'_1)  + 0.5552 \cdot\sin(4\cdot x'_1)  + 0.0696 \cdot\cos(5\cdot x'_1)  - 0.1653 \cdot\sin(5\cdot x'_1)  - 0.0245$, \\ \\
    here, for both $f_{0}(x)$ and $f_{1}(x)$, $x'_0 = 0.9936\cdot x - 0.0016$ and $x'_1 = 0.0009\cdot x - 0.0678$. \\ \\
    $f_{2}(x) = + 0.0012 \cdot B_{0}(x) \text{ (support: [-2.20, -1.80])}  + 0.0043 \cdot B_{1}(x) \text{ (support: [-1.80, -1.40])}  - 0.0096 \cdot B_{2}(x) \text{ (support: [-1.40, -1.00])}  - 0.0071 \cdot B_{3}(x) \text{ (support: [-1.00, -0.60])}  - 0.0197 \cdot B_{4}(x) \text{ (support: [-0.60, -0.20])}  - 0.0075 \cdot B_{5}(x) \text{ (support: [-0.20, 0.20])}  + 0.0045 \cdot B_{6}(x) \text{ (support: [0.20, 0.60])}  + 0.0038 \cdot B_{7}(x) \text{ (support: [0.60, 1.00])} + 0.4794 \cdot \mathrm{Tanh}(x)  $, \\ \\
    $f_{3}(x) = - 0.0008 \cdot B_{0}(x) \text{ (support: [-2.20, -1.80])}  - 0.0074 \cdot B_{1}(x) \text{ (support: [-1.80, -1.40])}  + 0.0026 \cdot B_{2}(x) \text{ (support: [-1.40, -1.00])}  + 0.0294 \cdot B_{3}(x) \text{ (support: [-1.00, -0.60])}  - 0.0009 \cdot B_{4}(x) \text{ (support: [-0.60, -0.20])}  + 0.0179 \cdot B_{5}(x) \text{ (support: [-0.20, 0.20])}  + 0.0037 \cdot B_{6}(x) \text{ (support: [0.20, 0.60])}  + 0.0031 \cdot B_{7}(x) \text{ (support: [0.60, 1.00])} - 0.0395 \cdot \mathrm{Tanh}(x)$.
    }
    \label{fig:wiki_dyg_4_scaled}
\end{figure*}

We provide plots of the same dataset trained with TGN and DyGFormer, shown in Figure \ref{fig:wiki_tgn_4_scaled} and Figure \ref{fig:wiki_dyg_4_scaled}, with the y-axes set to the same level for each backbone to facilitate a direct comparison). As the figures demonstrate, despite using different backbone models, the learned functions exhibit similar trends and shapes for each dimension. This illustrates the stability of our method and makes the interpretability process more reliable.

Of course, there may be some detailed differences between LeTEs trained on different backbones. This is intuitively due to the presence of various influencing factors, such as the model architecture, the interaction of LeTE with other modules, the optimization process and etc. However, we can validate the idea by inspecting the plot in a simplified manner.

\subsection{Comparing Lower- and Higher-dimensional LeTE}
\begin{figure*}
    \centering
    \includegraphics[width=1\linewidth]{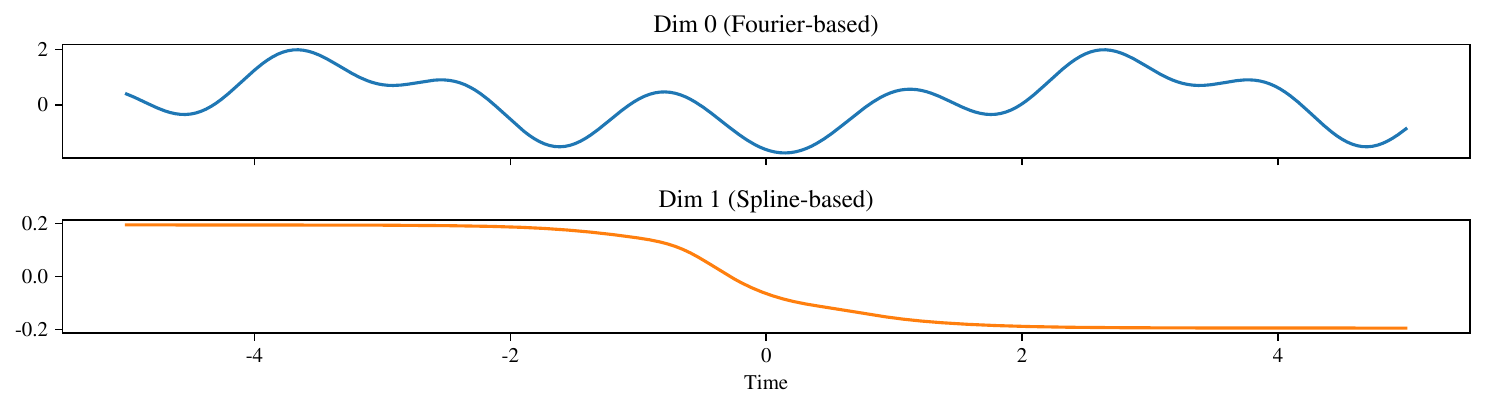}
    \caption{Plots of the two dimensions of the non-linear transformation functions of a 2-dimensional LeTE trained on Wikipedia/TGN. We further present the four functions here, the parameters are learned and read from the trained model: \\ \\
    $f_{0}(x) =-0.9560 \cdot\cos(1\cdot(x'))+0.3821 \cdot\sin(1\cdot(x'))+0.2430 \cdot\cos(2\cdot(x'))-0.3138 \cdot\sin(2\cdot(x'))-0.4807 \cdot\cos(3\cdot(x'))-0.1918 \cdot\sin(3\cdot(x'))-0.6372 \cdot\cos(4\cdot(x'))-0.2811 \cdot\sin(4\cdot(x'))+0.1555 \cdot\cos(5\cdot(x'))+0.0840 \cdot\sin(5\cdot(x'))+0.0334$, \\ \\
    here, $x' = 0.9963\cdot x-0.0092$. \\ \\
    $f_{1}(x) =-0.0020 \cdot B_{0}(x) \text{ (support: [-2.20,-1.80])}-0.0022 \cdot B_{1}(x) \text{ (support: [-1.80,-1.40])}+0.0097 \cdot B_{2}(x) \text{ (support: [-1.40,-1.00])}-0.0636 \cdot B_{3}(x) \text{ (support: [-1.00,-0.60])}-0.0556 \cdot B_{4}(x) \text{ (support: [-0.60,-0.20])}-0.0123 \cdot B_{5}(x) \text{ (support: [-0.20, 0.20])}-0.0054 \cdot B_{6}(x) \text{ (support: [0.20, 0.60])}-0.0011 \cdot B_{7}(x) \text{ (support: [0.60, 1.00])}-0.1945 \cdot \mathrm{Tanh}(x)$.}
    \label{fig:wiki_tgn_2}
\end{figure*}

We further compare the lower- and higher-dimensional LeTE by reconstructing the non-linear functions and plotting them (please refer to and compare Figure \ref{fig:wiki_tgn_4} and \ref{fig:wiki_tgn_2}). Intuitively, the higher-dimensional representation will provide more information. As seen from the plots, Dim 2 in Figure \ref{fig:wiki_tgn_2} is dominated by the basis function, partially losing the information captured by Dim 3 in Figure \ref{fig:wiki_tgn_4}.

From the perspective of the reconstructed functions, for the Fourier-based dimensions, the LeTE with only one Fourier-based dimension has a single input transformation, $x{\prime}_0$, and all frequency components are computed based on this transformation. This means the LeTE encodes on a broader time scale (reminder: we used the Wikipedia dataset) and models the time difference variations of editing activities without distinguishing patterns at different scales. Since there is only one Fourier-based dimension, all frequency components are controlled by the same input transformation, making it harder for the model to interpret editing patterns at different time scales. In contrast, for the LeTE with two Fourier-based dimensions, each dimension has different input transformations ($x{\prime}_0$ and $x{\prime}_1$), enabling the model to capture more detailed editing behaviors at different scales. For example, Dim 0 might rely more on $x{\prime}_0$ (with a larger scaling factor), focusing on short-term fluctuations (high-frequency components), while Dim 1 might rely more on $x{\prime}_1$ (with a smaller scaling factor), focusing more on long-term trends (low-frequency components). Thus, higher dimensions allow the model to handle editing behaviors at different time scales, providing higher interpretability.

Similarly, for the LeTE with only one Spline-based dimension, it primarily focuses on adjusting a single level, potentially describing how time affects editing behaviors. However, relying on just one Spline-based dimension may make it difficult to capture relatively complex time dynamics. For the LeTE with two Spline-based dimensions, the weights of the coefficients are more distributed, granting the overall LeTE stronger local adjustment capabilities. Moreover, since a dimension may be dominated by basis function or Spline functions, higher dimensions naturally have stronger expressive power.

Although higher-dimensional LeTEs offer stronger performance and better explain the information captured by the model, the interpretability analysis of such higher-dimensional LeTEs becomes more complex and may require a dimension-by-dimension analysis.

\section{Limitation and Future Work}

In this paper, we introduce \ours, a general time encoding method. Generally, Combined \ours~offers better performance as it leverages the strengths of both Fourier-based \ours~and Spline-based \ours, enabling it to effectively capture both the periodicity and non-periodicity of time. However, in practical scenarios, the choice of the hyperparameter $p$ or among the three variants may depend on the characteristics of the data and the specific task requirements.

We also explore the impact of the time encoding dimension on downstream task performance. Similarly, selecting an appropriate time encoding dimension may vary depending on the data and tasks. Notably, we observe that even with a small dimension, \ours~can achieve acceptable results in downstream tasks.

Additionally, we mention that certain position encoding methods can be considered special cases of our approach. However, as position encoding is not the primary focus of this paper, we did not provide formal proofs. We believe that extending the ideas proposed in this paper to models that use position encoding could yield improved results, making this a promising direction for future research.

\section{Code Implementation}

The codes are available at a \href{https://github.com/chenxi1228/LeTE}{GitHub Repository}.


\end{document}